%% file: main.tex
\documentclass[sigconf]{acmart}
\acmSubmissionID{304}

\usepackage{booktabs} %

\usepackage[ruled]{algorithm2e} %

\SetAlFnt{\small}
\SetAlCapFnt{\small}
\SetAlCapNameFnt{\small}
\SetAlCapHSkip{0pt}

\acmJournal{TOG}

\copyrightyear{2023}
\acmYear{2023}
\setcopyright{rightsretained}
\acmConference[SA Conference Papers '23]{SIGGRAPH Asia 2023 Conference Papers}{December 12--15, 2023}{Sydney, NSW, Australia}
\acmBooktitle{SIGGRAPH Asia 2023 Conference Papers (SA Conference Papers '23), December 12--15, 2023, Sydney, NSW, Australia}
\acmDOI{10.1145/3610548.3618154}
\acmISBN{979-8-4007-0315-7/23/12}

\usepackage{algorithmic}
\usepackage{cleveref}
\usepackage[export]{adjustbox}
\usepackage{rotating}
\usepackage{wrapfig}
\usepackage{float}
\usepackage{array}
\usepackage{subfig}
\usepackage{tabularx}
\usepackage{tikz}
\usetikzlibrary{spy}
\usepackage{enumitem}
\usepackage{colortbl}
\usepackage[utf8]{inputenc}
\usepackage{pgfplots}
\usepackage{pifont}

\newcommand\blfootnote[1]{%
  \begingroup
  \renewcommand\thefootnote{}\footnote{#1}%
  \addtocounter{footnote}{-1}%
  \endgroup
}

\begin{document}
\title{Break-A-Scene: Extracting Multiple Concepts from a Single Image}

\author{Omri Avrahami}
\orcid{0000-0002-7628-7525}
\affiliation{
 \institution{The Hebrew University of Jerusalem \\ Google Research}
 \city{Jerusalem}
 \country{Israel}}
\email{omri.avrahami@mail.huji.ac.il}
\authornote{Performed this work while working at Google}

\author{Kfir Aberman}
\orcid{0000-0002-4958-601X}
\affiliation{
 \institution{Google Research}
 \city{San Fransisco}
 \country{USA}}
\email{kfiraberman@gmail.com}

\author{Ohad Fried}
\orcid{0000-0001-7109-4006}
\affiliation{
 \institution{Reichman University}
 \city{Herzliya}
 \country{Israel}}
\email{ofried@runi.ac.il}

\author{Daniel Cohen-Or}
\orcid{0000-0001-6777-7445}
\affiliation{
 \institution{Tel Aviv University \\ Google Research}
 \city{Tel Aviv}
 \country{Israel}}
\email{cohenor@gmail.com}
\authornotemark[1]

\author{Dani Lischinski}
\orcid{0000-0002-6191-0361}
\affiliation{
 \institution{The Hebrew University of Jerusalem \\ Google Research}
 \city{Jerusalem}
 \country{Israel}}
\email{danix@mail.huji.ac.il}
\authornotemark[1]

\renewcommand\shortauthors{Avrahami et al.}

\def\ShowNotes{}
\input{macros.tex}

\input{sections/abstract.tex}



\input{figures/teaser/fig.tex}

\maketitle

\blfootnote{Project page is available at: \textcolor{red}{\href{https://omriavrahami.com/break-a-scene/}{https://omriavrahami.com/break-a-scene/}}}

\input{sections/introduction.tex}
\input{sections/related_work.tex}
\input{sections/method.tex}
\input{sections/experiments.tex}
\input{sections/limitations.tex}

\input{figures/qualitative_comparison/fig.tex}
\input{figures/applications/fig.tex}

\clearpage
\bibliographystyle{ACM-Reference-Format}
\bibliography{egbib}

\clearpage
\appendix
\begin{acks}
  We thank Nataniel Ruiz, Chu Qinghao, and Yael Pitch for their inspiring inputs that influenced this work. Additionally, we thank Jason Baldrige for providing valuable inputs that enhanced the quality of this project.
\end{acks}
\input{sections/appendix/additional_experiments.tex}
\input{sections/appendix/implementation_details.tex}
\input{sections/appendix/societal_impact.tex}

\end{document}

%% file: macros.tex
\newcommand{\ignorethis}[1]{}
\newcommand{\redund}[1]{#1}

\newcommand{\etal       }     {{et~al.}}
\newcommand{\apriori    }     {\textit{a~priori}}
\newcommand{\aposteriori}     {\textit{a~posteriori}}
\newcommand{\perse      }     {\textit{per~se}}
\newcommand{\eg         }     {{e.g.}}
\newcommand{\Eg         }     {{E.g.}}
\newcommand{\ie         }     {{i.e.}}
\newcommand{\naive      }     {{na\"{\i}ve}}
\newcommand{\Naive      }     {{Na\"{\i}ve}}

\newcommand{\Identity   }     {\mat{I}}
\newcommand{\Zero       }     {\mathbf{0}}
\newcommand{\Reals      }     {{\textrm{I\kern-0.18em R}}}
\newcommand{\isdefined  }     {\mbox{\hspace{0.5ex}:=\hspace{0.5ex}}}
\newcommand{\texthalf   }     {\ensuremath{\textstyle\frac{1}{2}}}
\newcommand{\half       }     {\ensuremath{\frac{1}{2}}}
\newcommand{\third      }     {\ensuremath{\frac{1}{3}}}
\newcommand{\fourth     }     {\ensuremath{\frac{1}{4}}}

\newcommand{\Lone} {\ensuremath{L_1}}
\newcommand{\Ltwo} {\ensuremath{L_2}}

\newcommand{\mat        } [1] {{\text{\boldmath $\mathbit{#1}$}}}
\newcommand{\Approx     } [1] {\widetilde{#1}}
\newcommand{\change     } [1] {\mbox{{\footnotesize $\Delta$} \kern-3pt}#1}

\newcommand{\Order      } [1] {O(#1)}
\newcommand{\set        } [1] {{\lbrace #1 \rbrace}}
\newcommand{\floor      } [1] {{\lfloor #1 \rfloor}}
\newcommand{\ceil       } [1] {{\lceil  #1 \rceil }}
\newcommand{\inverse    } [1] {{#1}^{-1}}
\newcommand{\transpose  } [1] {{#1}^\mathrm{T}}
\newcommand{\invtransp  } [1] {{#1}^{-\mathrm{T}}}
\newcommand{\relu       } [1] {{\lbrack #1 \rbrack_+}}

\newcommand{\abs        } [1] {{| #1 |}}
\newcommand{\Abs        } [1] {{\left| #1 \right|}}
\newcommand{\norm       } [1] {{\| #1 \|}}
\newcommand{\Norm       } [1] {{\left\| #1 \right\|}}
\newcommand{\pnorm      } [2] {\norm{#1}_{#2}}
\newcommand{\Pnorm      } [2] {\Norm{#1}_{#2}}
\newcommand{\inner      } [2] {{\langle {#1} \, | \, {#2} \rangle}}
\newcommand{\Inner      } [2] {{\left\langle \begin{array}{@{}c|c@{}}
                               \displaystyle {#1} & \displaystyle {#2}
                               \end{array} \right\rangle}}

\newcommand{\twopartdef}[4]
{
  \left\{
  \begin{array}{ll}
    #1 & \mbox{if } #2 \\
    #3 & \mbox{if } #4
  \end{array}
  \right.
}

\newcommand{\fourpartdef}[8]
{
  \left\{
  \begin{array}{ll}
    #1 & \mbox{if } #2 \\
    #3 & \mbox{if } #4 \\
    #5 & \mbox{if } #6 \\
    #7 & \mbox{if } #8
  \end{array}
  \right.
}

\newcommand{\len}[1]{\text{len}(#1)}

\newlength{\w}
\newlength{\h}
\newlength{\x}

\definecolor{darkred}{rgb}{0.7,0.1,0.1}
\definecolor{darkgreen}{rgb}{0.1,0.6,0.1}
\definecolor{cyan}{rgb}{0.7,0.0,0.7}
\definecolor{otherblue}{rgb}{0.1,0.4,0.8}
\definecolor{maroon}{rgb}{0.76,.13,.28}
\definecolor{burntorange}{rgb}{0.81,.33,0}

\newif\ifdraft
\drafttrue

\ifdraft
\newcommand{\omri}[1]{{\color{burntorange}[\textbf{Omri:} #1]}}
\newcommand{\kfir}[1]{{\color{darkred}[\textbf{Kfir:} #1]}}
\newcommand{\ohad}[1]{{\color{magenta}[\textbf{Ohad:} #1]}}
\newcommand{\dcor}[1]{{\color{red}[\textbf{Danny:} #1]}}
\newcommand{\danix}[1]{{\color{blue}[\textbf{Dani:} #1]}}

\newcommand{\oa}[1]{{\color{burntorange}#1}}
\newcommand{\ka}[1]{{\color{darkred}#1}}
\newcommand{\of}[1]{{\color{magenta}#1}}
\newcommand{\dc}[1]{{\color{red}#1}}
\newcommand{\dl}[1]{{\color{blue}#1}}

\newcommand{\todo}[1]{{\color{cyan}[\textbf{TODO:} #1]}}

\else
\newcommand{\omri}[1]{}
\newcommand{\kfir}[1]{}
\newcommand{\ohad}[1]{}
\newcommand{\dcor}[1]{}
\newcommand{\danix}[1]{}
\newcommand {\note}[1]{}
\newcommand {\todo}[1]{}
\newcommand{\oa}[1]{{\color{black}#1}}
\newcommand{\ka}[1]{{\color{black}#1}}
\newcommand{\of}[1]{{\color{black}#1}}
\newcommand{\dc}[1]{{\color{black}#1}}
\newcommand{\dl}[1]{{\color{black}#1}}
\fi

\newcommand {\reqs}[1]{\colornote{red}{\tiny #1}}

\newcommand {\new}[1]{{\color{red}{#1}}}

\newcommand*\rot[1]{\rotatebox{90}{#1}}

\newcommand {\newstuff}[1]{#1}

\newcommand\todosilent[1]{}

\definecolor{tokenacolor}{RGB}{200,0,0}
\newcommand {\tokena}{{\color{tokenacolor}{$[v_1]$ }}}
\definecolor{tokenbcolor}{RGB}{0,0,200}
\newcommand {\tokenb}{{\color{tokenbcolor}{$[v_2]$ }}}
\definecolor{tokenccolor}{RGB}{0,200,0}
\newcommand {\tokenc}{{\color{tokenccolor}{$[v_3]$ }}}
\definecolor{tokendcolor}{RGB}{0,180,180}
\newcommand {\tokend}{{\color{tokendcolor}{$[v_4]$ }}}
\definecolor{tokenecolor}{RGB}{180,180,180}
\newcommand {\tokene}{{\color{tokenecolor}{$[v_5]$ }}}
\definecolor{tokenfcolor}{RGB}{180,0,180}
\newcommand {\tokenf}{{\color{tokenfcolor}{$[v_6]$ }}}
\definecolor{tokenbgcolor}{RGB}{180,180,0}
\newcommand {\tokenbg}{{\color{tokenbgcolor}{$[v_{bg}]$}}}

\newcommand{\woBGmask}{{w/o~bg~\&~mask}}
\newcommand{\woMask}{{w/o~mask}}

\providecommand{\keywords}[1]
{
  \textbf{\textit{Keywords---}} #1
}

\newcommand{\DALLE}{{DALL$\cdot$E}}

\newcommand{\cmark}{\color{darkgreen}{\ding{51}}}
\newcommand{\xmark}{\color{darkred}{\ding{55}}}

\newlength{\ww}

\newcommand{\maskedDB}{DB-m\xspace}
\newcommand{\maskedTI}{TI-m\xspace}
\newcommand{\maskedCD}{CD-m\xspace}

%% file: sections/abstract.tex
\begin{abstract}
    Text-to-image model personalization aims to introduce a user-provided concept to the model, allowing its synthesis in diverse contexts. However, current methods primarily focus on the case of learning a \emph{single} concept from \emph{multiple} images with variations in backgrounds and poses, and struggle when adapted to a different scenario. In this work, we introduce the task of textual scene decomposition: given a \emph{single} image of a scene that may contain \emph{several} concepts, we aim to extract a distinct text token for each concept, enabling fine-grained control over the generated scenes. To this end, we propose augmenting the input image with masks that indicate the presence of target concepts. These masks can be provided by the user or generated automatically by a pre-trained segmentation model. We then present a novel two-phase customization process that optimizes a set of dedicated textual embeddings (handles), as well as the model weights, striking a delicate balance between accurately capturing the concepts and avoiding overfitting. We employ a masked diffusion loss to enable handles to generate their assigned concepts, complemented by a novel loss on cross-attention maps to prevent entanglement. We also introduce union-sampling, a training strategy aimed to improve the ability of combining multiple concepts in generated images. We use several automatic metrics to quantitatively compare our method against several baselines, and further affirm the results using a user study. Finally, we showcase several applications of our method.
\end{abstract}

%% file: figures/teaser/fig.tex
\begin{teaserfigure}
    \centering
    \includegraphics[width=\linewidth]{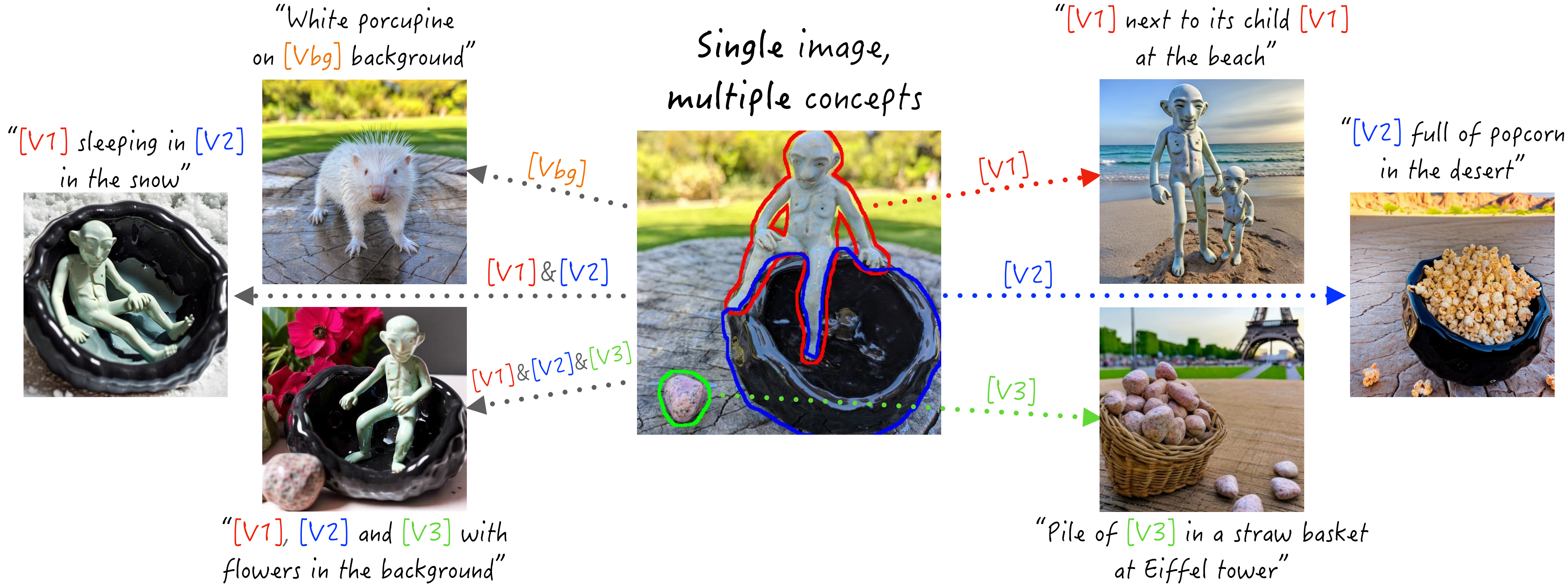} 
    \caption{\textbf{Break-A-Scene}: Given a \textit{single} image with \textit{multiple} concepts, annotated by loose segmentation masks (middle), our method can learn a distinct token for each concept, and use natural language guidance to re-synthesize the individual concepts (right) or combinations of them (left) in various contexts.
    }
    \label{fig:teaser}
\end{teaserfigure}

%% file: sections/introduction.tex
\section{Introduction}
\label{sec:introduction}

\input{tables/methods_overview.tex}

Humans have a natural ability to decompose complex scenes into their constituent parts and envision them in diverse contexts. For instance, given a photo of a ceramic artwork depicting a creature seated on a bowl (\Cref{fig:teaser}), one can effortlessly imagine the \emph{same creature} in a variety of different poses and locations, or envision the \emph{same bowl} in a new setting. However, today's generative models struggle when confronted with this type of task.

Recent works \cite{Ruiz2022DreamBoothFT, Gal2022AnII} suggested personalizing large-scale text-to-image models \cite{Rombach2021HighResolutionIS,Saharia2022PhotorealisticTD}: given \emph{several} images of a \emph{single} concept, they optimize newly-added dedicated text embeddings \cite{Gal2022AnII} or fine-tune the model weights \cite{Ruiz2022DreamBoothFT} in order to enable synthesizing instances of this concept in novel contexts. These works initiated a vibrant research field, surveyed in more detail in \Cref{sec:related_work} and summarized in \Cref{tab:previous_methods_overview}.

In this work, we introduce the new scenario of \emph{textual scene decomposition}: given a \emph{single} image of a scene that may contain \emph{multiple} concepts of different kinds, our goal is to extract a dedicated text token for each concept. This enables generation of novel images from textual prompts, featuring individual concepts or combinations of multiple concepts, as demonstrated in \Cref{fig:teaser}.

The personalization task can be inherently ambiguous: it is not always clear which concepts we intend to extract/learn. Previous works \cite{Ruiz2022DreamBoothFT,Gal2022AnII} resolve this ambiguity by extracting a single concept at a time, utilizing several different images that depict the concept in different contexts. However, when switching to a single image scenario, other means are necessary to disambiguate the task. 
Specifically, we propose to augment the input image with a set of masks, indicating the concepts that we aim to extract. These masks may be loose masks provided by the user, or generated by an automatic segmentation method (e.g., \cite{kirillov2023segment}). However, as demonstrated in \Cref{fig:reconstruction_editability_tradeoff}, adapting the two main approaches, TI \cite{Gal2022AnII} and DB \cite{Ruiz2022DreamBoothFT}, to this setting reveals a reconstruction-editability tradeoff: while TI fails to accurately reconstruct the concepts in a new context, DB loses the ability to control the context due to overfitting.

\input{figures/reconstruction_editability_tradeoff/fig.tex}

In this work, we propose a novel customization pipeline that effectively balances the preservation of learned concept identity with the avoidance of overfitting. Our pipeline, depicted in \Cref{fig:method_overview}, consists of two phases.
In the first phase, we designate a set of dedicated text tokens (handles), freeze the model weights, and optimize the handles to reconstruct the input image. In the second phase, we switch to fine-tuning the model weights, while continuing to optimize the handles.

We also recognize that in order to generate images exhibiting combinations of concepts, the customization process cannot be carried out separately for each concept. This observation leads us to introduce \emph{union-sampling}, a training strategy that addresses this requirement and enhances the generation of concept combinations.

A crucial focus of our approach is on disentangled concept extraction, \ie, ensuring that each handle is associated with only a single target concept. To achieve this, we employ a  masked version of the standard diffusion loss, which guarantees that each custom handle can generate its designated concept; however, this loss does not penalize the model for associating a handle with multiple concepts. Our main insight is that we can penalize such entanglement by additionally imposing a loss on the cross-attention maps, known to correlate with the scene layout \cite{Hertz2022PrompttoPromptIE}. This additional loss ensures that each handle attends only to the areas covered by its target concept.

We propose several automatic metrics for our task and use them to compare our method against the baselines. In addition, we conduct a user study and show that our method is also preferred by human evaluators. Finally, we present several applications of our method.

In summary, our contributions are: (1) we introduce the new task of textual scene decomposition, (2) propose a novel approach for this setting, which learns a set of disentangled concept handles, while balancing between concept fidelity and scene editability, and (3) propose several automatic evaluation metrics and use them, in addition to a user study, to demonstrate the effectiveness of our method.

%% file: tables/methods_overview.tex
\begin{table}
    \centering
    \caption{\textbf{Scenarios of previous work on model personalization.} Our method is the first to offer personalization of \emph{multiple concepts} given a \emph{single input image}. An extended version of this table, that also includes the concurrent works, is available in the supplementary materials.}
    \resizebox{1\columnwidth}{!}{
        \begin{tabular}{>{\columncolor[gray]{0.95}}lcc}
            \toprule
            
            \textbf{Method} & 
            Single input &
            Multi-concept
            \\

            &
            image &
            output
            \\
            
            \midrule

            Textual Inversion \cite{Gal2022AnII} &
            \xmark &
            \xmark
            \\

            DreamBooth \cite{Ruiz2022DreamBoothFT} &
            \xmark &
            \xmark
            \\

            Custom Diffusion \cite{Kumari2022MultiConceptCO} &
            \xmark & 
            \cmark
            \\

            ELITE \cite{Wei2023ELITEEV} &
            \cmark &
            \xmark
            \\

            E4T \cite{Gal2023EncoderbasedDT} &
            \cmark & 
            \xmark
            \\

            \midrule
            Ours &
            \cmark &
            \cmark
            \\
            
            \bottomrule
        \end{tabular}
    }
    \label{tab:previous_methods_overview}
\end{table}

%% file: figures/reconstruction_editability_tradeoff/fig.tex
\begin{figure}[t]
    \centering
    \setlength{\ww}{\columnwidth}
    
    \includegraphics[width=\ww]{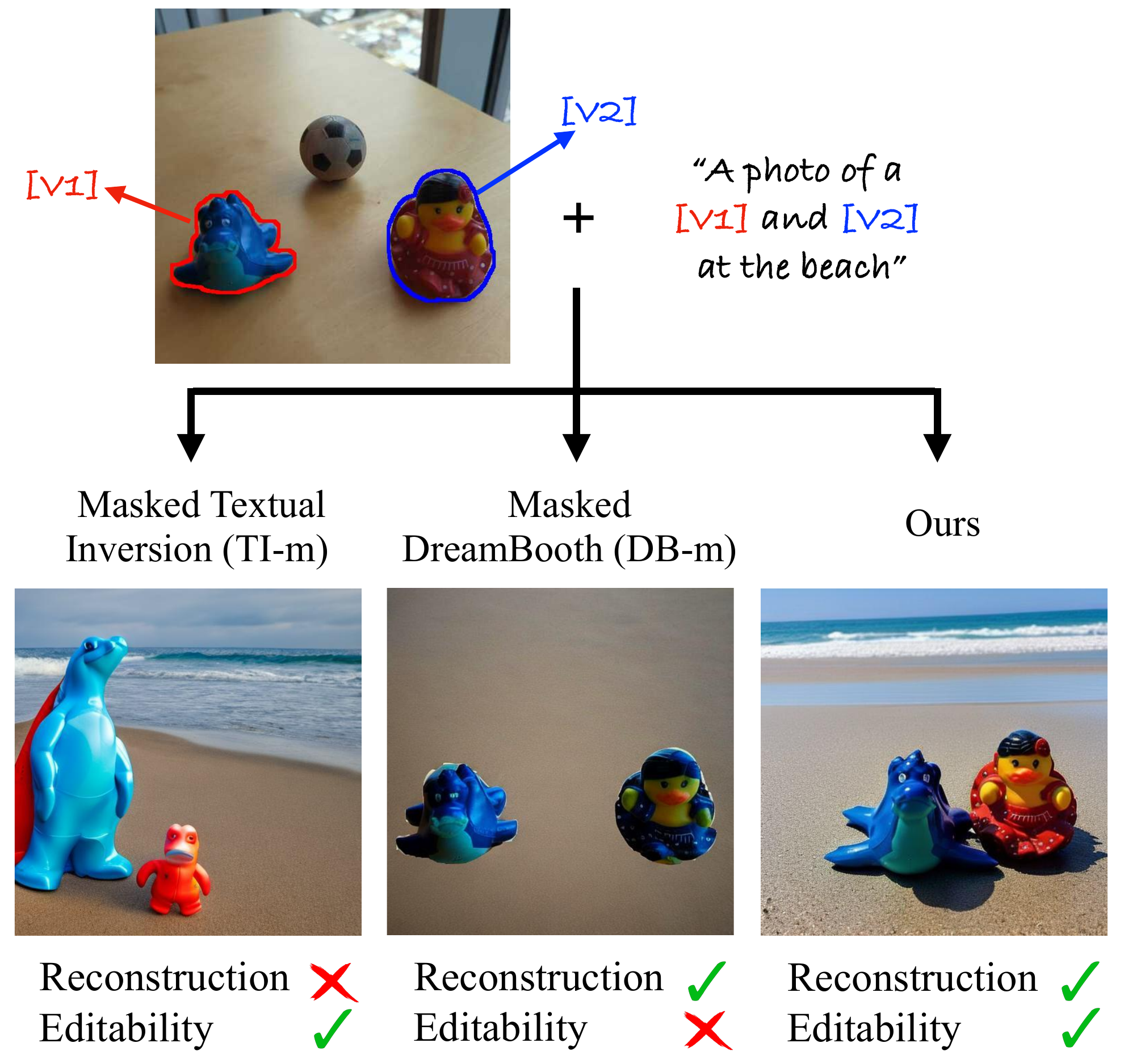}
    
    \caption{\textbf{Reconstruction-editability tradeoff:} 
    Given a single input image along with masks, and the prompt ``A photo of a \tokena and \tokenb at the beach'', (masked) Textual Inversion~\cite{Gal2022AnII} generates an image of two objects on a beach, but fails to preserve their identities. (Masked) DreamBooth~\cite{Ruiz2022DreamBoothFT} preserves the identities well, but fails to place them on a beach. Our method is able to generate a convincing image of two objects on a beach, which closely resemble the objects in the input image.}
    \label{fig:reconstruction_editability_tradeoff}
\end{figure}

%% file: sections/related_work.tex
\section{Related Work}
\label{sec:related_work}

\paragraph{\textbf{Text-to-image synthesis}} The field of text-to-image synthesis has seen immense progress in recent years. The initial approaches utilized RNNs \cite{Mansimov2016GeneratingIF}, GANs \cite{reed2016generative, zhang2017stackgan, zhang2018stackgan++, xu2018attngan} and transformers \cite{Ramesh2021ZeroShotTG, gafni2022make}.
However, diffusion-based models \cite{ho2020denoising, sohl2015deep, song2020denoising, song2019generative} emerged as superior for text-to-image generation \cite{chang2023muse, ramesh2022hierarchical, Rombach2021HighResolutionIS, Saharia2022PhotorealisticTD, yu2022scaling}.

Alongside these advancements, text-driven image editing has emerged, enabling global \cite{meng2021sdedit, crowson2022vqgan, kwon2021clipstyler, patashnik2021styleclip, brooks2022instructpix2pix, pnpDiffusion2022, valevski2022unitune} and local manipulations \cite{avrahami2022blended, avrahami2022blendedlatent, nichol2021glide, couairon2022diffedit, bar2022text2live, wang2022imagen, bau2021paint, sheynin2022knn, Kawar2022ImagicTR,Patashnik2023LocalizingOS}. In addition, diffusion models have also been employed for video generation \cite{Singer2022MakeAVideoTG, Ho2022ImagenVH}, video editing \cite{Molad2023DreamixVD}, scene generation \cite{Avrahami2022SpaTextSR, BarTal2023MultiDiffusionFD}, mesh texturing \cite{Richardson2023TEXTureTT}, typography generation \cite{Iluz2023WordAsImageFS}, and solving inverse problems \cite{Saharia2021ImageSV, Horwitz2022ConffusionCI, Saharia2021PaletteID}. 

\paragraph{\textbf{Cross-attention.}}
Prompt-to-prompt \cite{Hertz2022PrompttoPromptIE} utilizes cross-attention maps in text-to-image diffusion models for manipulating generated images, later extended to handle real images through inversion \cite{mokady2022null}. Attend-and-excite \cite{Chefer2023AttendandExciteAS} use cross-attention maps as an explainability-based technique \cite{Chefer2020TransformerIB, Chefer2021GenericAE} to adjust text-to-image generations. In our work, we employ cross-attention maps to disentangle learned concepts; however, our work focuses on extracting textual handles from a scene and remixing them into completely novel scenes, rather than editing the input image.

\paragraph{\textbf{Inversion.}}
In the realm of generative models, \emph{inversion} \cite{Xia2021GANIA} is the task of finding a code within the latent space of a generator \cite{goodfellow2014generative,karras2019style, karras2020analyzing} that faithfully reconstructs a given image. Inversion may be accomplished via direct optimization of the latent code \cite{abdal2019image2stylegan, abdal2020image2stylegan++, Zhu2020ImprovedSE} or by training a dedicated encoder \cite{Richardson2020EncodingIS,Tov2021DesigningAE, Pidhorskyi2020AdversarialLA, zhu2020domain, alaluf2021hyperstyle}. PTI \cite{Roich2021PivotalTF} follows the latent optimization with refinement of %
the model weights \cite{Bau2019SemanticPM}. In this study, we also employ a two-stage approach wherein we first optimize only the textual embeddings of the target concepts, followed by jointly training the embeddings and the model weights.

\paragraph{\textbf{Personalization.}}
The task of \emph{personalization} aims to identify a user-provided concept that is not prevalent in the training data for discriminative \cite{Cohen2022ThisIM} or generative \cite{nitzan2022mystyle} tasks. Textual Inversion (TI) \cite{Gal2022AnII}, and DreamBooth (DB) \cite{Ruiz2022DreamBoothFT} are two seminal works that address personalization of text-to-image models: given \emph{several} images of a \emph{single} visual concept, they learn to generate this concept in different contexts. TI introduces a new learnable text token and optimizes it to reconstruct the concept using the standard diffusion loss, while keeping the model weights frozen. DB, on the other hand, reuses an existing rare token, and fine-tunes the model weights to reconstruct the concept. Custom Diffusion \cite{Kumari2022MultiConceptCO}  fine-tunes only a subset of the layers, while LoRA \cite{lora,lora_diffusion} restricts their updates to rank 1. Perfusion \cite{Tewel2023KeyLockedRO} also performs a rank 1 update along with an attention key locking mechanism.

Concurrently with our work, SVDiff \cite{Han2023SVDiffCP} introduces an efficient personalization method in the parameter space based on singular-value decomposition of weight kernels. They also propose a mixing and unmixing regularization that enables generating two concepts next to each other. In contrast to our method, SVDiff requires \emph{several} images for each of the concepts, while we operate on a \emph{single} image containing \emph{multiple} concepts. Furthermore, SVDiff's automatic augmentation allows for the placement of two objects side by side, while our method enables arbitrary placement of up to four objects.

Most recently, fast personalization methods were introduced that employ dedicated encoders \cite{Gal2023EncoderbasedDT, Wei2023ELITEEV, Chen2023SubjectdrivenTG, Jia2023TamingEF, Shi2023InstantBoothPT} and can also handle a single image. Among these, only ELITE \cite{Wei2023ELITEEV} is publicly available, and we include it in our comparisons in \Cref{sec:comparisons}. XTI \cite{Voynov2023PET} extends TI to utilize a richer inversion space. As shown in \Cref{tab:previous_methods_overview}, our approach stands out from the existing personalization methods by addressing the challenge of coping with \emph{multiple} concepts within a \emph{single} image. To the best of our knowledge, this is the first work to tackle this task.

%% file: sections/method.tex
\section{Method}
\label{sec:method}
\input{figures/method_overview/fig.tex}
\input{figures/cross_attention_comparison/fig.tex}

Given a single input image $I$ and a set of $N$ masks $\{M_i\}_{i=1}^{N}$, indicating concepts of interest in the image, we aim to extract $N$ \emph{textual handles} $\{v_i\}_{i=1}^{N}$, s.t. the $i$th handle, $v_i$, represents the concept indicated by $M_i$. The resulting handles can then be used in text prompts to guide the synthesis of new instances of each concept, or novel combinations of several concepts, as demonstrated in \Cref{fig:teaser}.

Attempting to adapt TI or DB to extraction of multiple concepts from a single image (by using masks, as explained in \Cref{sec:experiments}), reveals an inherent reconstruction-editability tradeoff. As demonstrated in \Cref{fig:reconstruction_editability_tradeoff}, TI enables embedding the extracted concepts in a new context, but fails to faithfully preserve their identity, while fine-tuning the model in DB captures the identity, at the cost of losing editability, to a point of failing to comply with the guiding text prompt. We observe that optimizing only individual tokens is not expressive enough for good reconstruction, while fine-tuning the model using a single image is extremely prone to overfitting. In this work, we strive for a ``middle ground'' solution that would combine the best of both worlds, \ie, would be able to capture the identity of the target concepts without relinquishing editability. Our approach combines four key components, as depicted in \Cref{fig:method_overview} and described below.

\paragraph{\textbf{Balancing between reconstruction and editability:}}
We optimize both the text embeddings and the model's weights \cite{lora_diffusion}, but do so in two different phases. In the first phase, the model is frozen, while the text embeddings corresponding to the masked concepts are optimized \cite{Gal2022AnII} using a high learning rate. Thus, an initial approximate embedding is achieved quickly without detracting from the generality of the model, which then serves as a good starting point for the next phase. In the second phase, we unfreeze the model weights and optimize them along with the text tokens, using a significantly lower learning rate. This gentle fine-tuning of the weights and the tokens enables faithful reconstruction of the extracted concepts in novel contexts, with minimal editability degradation.

\paragraph{\textbf{Union-sampling:}}
We further observe that if the above process considers each concept separately, the resulting customized model struggles to generate images that exhibit a combination of several concepts (see \Cref{fig:autoamtic_dataset_qualitative_comparison_ablation_mini} and \Cref{sec:comparisons}).
Thus, we propose \emph{union-sampling} for each of the two optimization phases.
Specifically, we start by designating an initial textual embedding (handle) $v_i$ for each concept indicated by mask $M_i$. Next, at each training step, we randomly select a \emph{subset} of $k \leq N$ concepts, $s = \{i_1,\ldots,i_k\} \subseteq [N]$, and construct a text prompt ``a photo of $[v_{i_1}]$ and $\ldots [v_{i_k}]$''. The optimization losses described below are then computed with respect to the union of the corresponding masks, $M_s = \bigcup M_{i_k}$.

\paragraph{\textbf{Masked diffusion loss:}}
The handles (and the model weights, in the second phase) are optimized using a masked version of the standard diffusion loss \cite{lora_diffusion}, \ie, by penalizing only over the pixels covered by the concept masks: 
\begin{equation}
    \mathcal{L}_{\mathrm{rec}} = \mathbb{E}_{z, s, \epsilon \sim \mathcal{N}(0, 1), t }\Big[ \Vert \epsilon \odot M_s - \epsilon_\theta(z_{t},t,p_s) \odot M_s \Vert_{2}^{2}\Big],
    \label{eq:masked_LDM_loss}
\end{equation}
where $z_t$ is the noisy latent at time step $t$, $p_s$ is the text prompt, $M_s$ is the union of the corresponding masks, $\epsilon$ is the added noise, and, $\epsilon_\theta$ is the denoising network.
Using the masked diffusion loss in pixel space encourages the process to faithfully reconstruct the concepts. However, no penalty is imposed for associating a single handle with multiple concepts, as demonstrated in \Cref{fig:autoamtic_dataset_qualitative_comparison_ablation_mini}.
Thus, with this loss alone, the resulting handles fail to cleanly separate between the corresponding concepts.

In order to understand the source of this issue, it is helpful to examine the cross-attention maps between the learned handles and the generated images, as visualized in \Cref{fig:cross_attention_comparison} (top row).
It may be seen that both handles \tokena and \tokenb attend to the union of the areas containing the two concepts in the generated image, instead of each handle attending to just one concept, as we would have liked.

\paragraph{\textbf{Cross-Attention loss:}}
We therefore introduce another loss term that encourages the model to not only reconstruct the pixels of the learned concepts, but also ensures that each handle attends only to the image region occupied by the corresponding concept.
Specifically, as illustrated in \Cref{fig:method_overview} (right), we utilize the cross-attention maps for the newly-added tokens and penalize their MSE deviation from the input masks. Formally, we add the following term to loss in both training phases:
\begin{equation}
    \mathcal{L}_{\mathrm{attn}} = \mathbb{E}_{z, k, t}\Big[ \Vert CA_{\theta}(v_i, z_t) - M_{i_k} \Vert_{2}^{2}\Big],
    \label{eq:attention_loss}
\end{equation}
where $CA_{\theta}(v_i, z_t)$ is the cross-attention map between the token $v_i$ and the noisy latent $z_t$.
The cross attention maps are calculated over several layers of the denoising UNet model (for more details, please see the supplementary material). Thus, the total loss used is:
\begin{equation}
    \mathcal{L}_{\mathrm{total}} = \mathcal{L}_{\mathrm{rec}} + \lambda_{\mathrm{attn}} \mathcal{L}_{\mathrm{attn}},
    \label{eq:total_loss}
\end{equation}
where $\lambda_{\mathrm{attn}} = 0.01$. As can be seen in \Cref{fig:cross_attention_comparison} (bottom row), the addition of $\mathcal{L}_{\mathrm{attn}}$ to the loss succeeds in ensuring that \tokena and \tokenb attend to two distinct regions, corresponding to the appropriate spatial locations in the generated image.

%% file: figures/method_overview/fig.tex
\begin{figure*}[ht]
    \centering    
    \includegraphics[width=\linewidth]{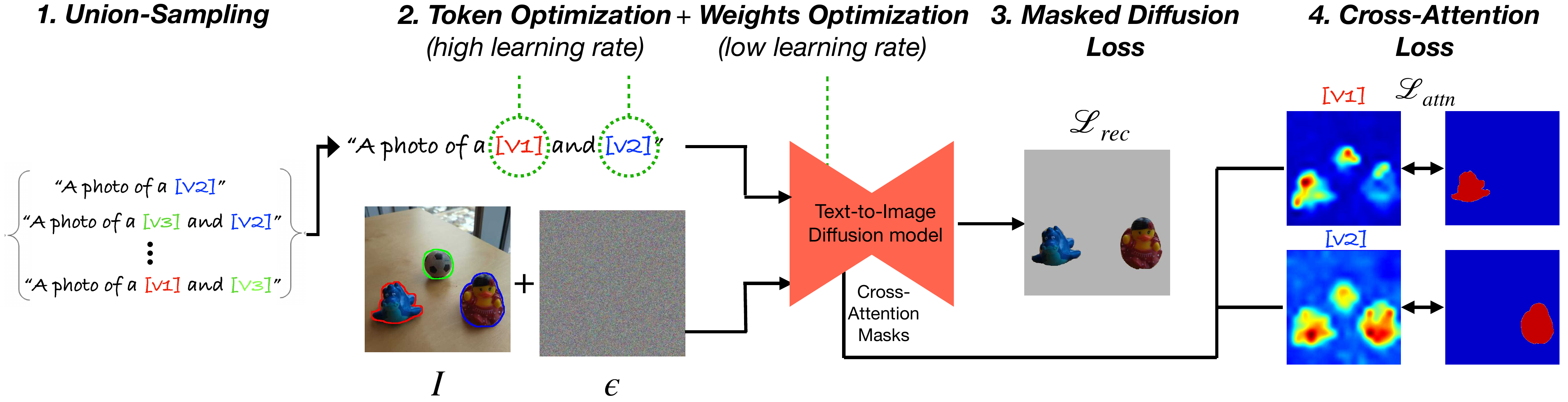}
    
    \caption{\textbf{Method overview:} our method consists of four key components: (1) in order to train the model to support different combinations of generated concepts, we employ a \emph{union-sampling} mechanism, where a random subset of the tokens is sampled each time. In addition, (2) in order to avoid overfitting, we use a \emph{two-phase training regime}, which starts by optimizing only the newly-added tokens, with a high learning rate, and in the second phase we also train the model weights, using a lower learning rate. A \emph{masked diffusion loss} (3) is used to reconstruct the desired concepts.
    Finally, (4) in order to encourage disentanglement between the learned concepts, we use a novel \emph{cross-attention loss}.}
    \label{fig:method_overview}
\end{figure*}

%% file: figures/cross_attention_comparison/fig.tex
\begin{figure}[t]
    \centering
    \setlength{\tabcolsep}{1.5pt}
    \renewcommand{\arraystretch}{0.5}
    \setlength{\ww}{0.23\columnwidth}
    \begin{tabular}{c @{\hspace{5\tabcolsep}} cccc}
        \raisebox{-0.7\height}[0pt][0pt]{\frame{\includegraphics[width=\ww]{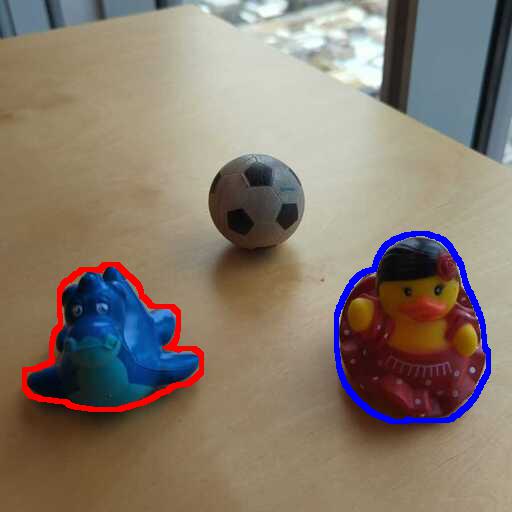}}} &
        \raisebox{1.5\height}[0pt][0pt]{\rotatebox[origin=c]{90}{\scriptsize{w/o attn. loss}}} &
        \frame{\includegraphics[width=\ww]{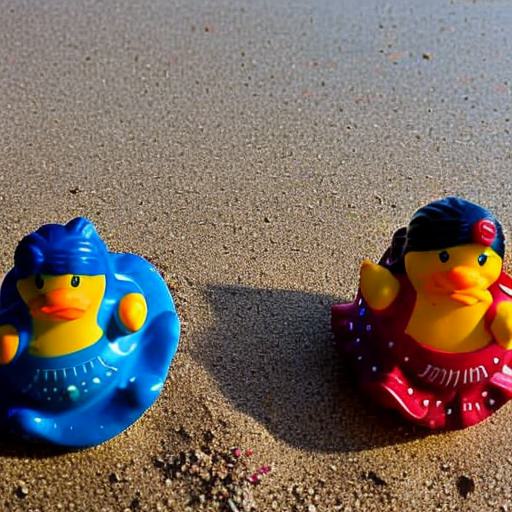}} &
        \frame{\includegraphics[width=\ww]{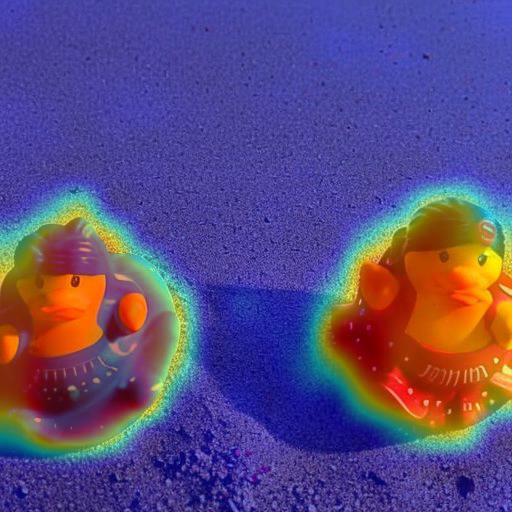}} &
        \frame{\includegraphics[width=\ww]{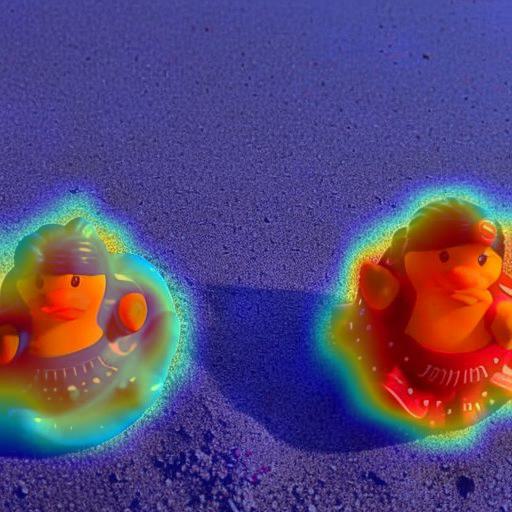}}
        \\
        \\

        \vspace{2px} &
        \raisebox{1.5\height}[0pt][0pt]{\rotatebox[origin=c]{90}{\scriptsize{w attn. loss}}} &
        \frame{\includegraphics[width=\ww]{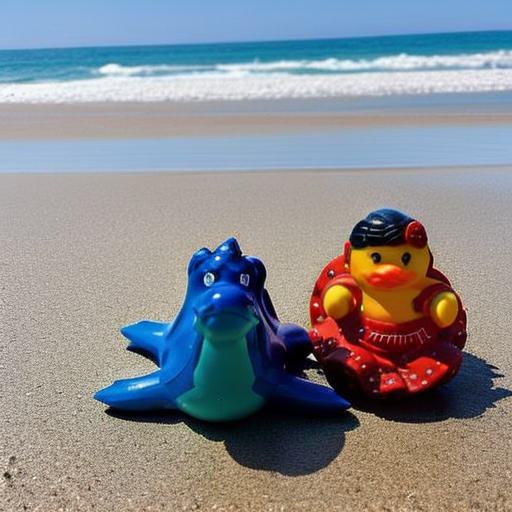}} 
        &
        \frame{\includegraphics[width=\ww]{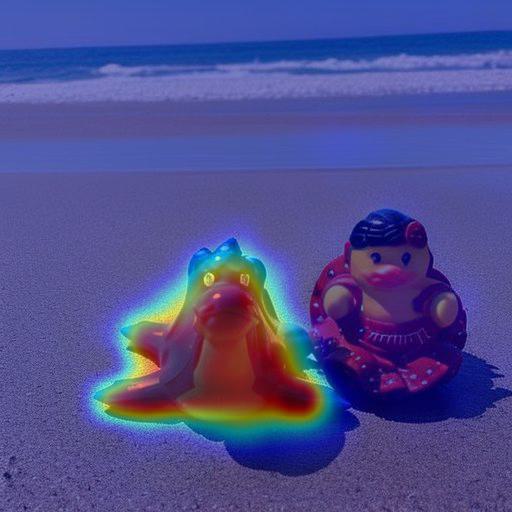}} &
        \frame{\includegraphics[width=\ww]{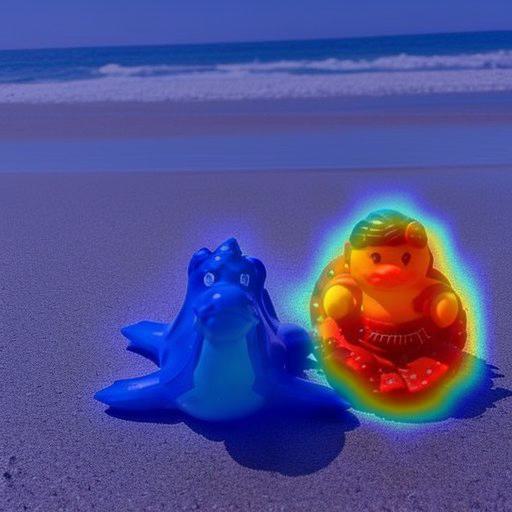}}
        \\

        \scriptsize{Inputs} &
        &
        \scriptsize{Output} &
        \scriptsize{\tokena} &
        \scriptsize{\tokenb}
        \\
    \end{tabular}
    
    \caption{\textbf{Cross-attention loss:} given the input scene and the text prompt ``a photo of \tokena and \tokenb at the beach'', we visualize the cross-attention maps of the generated images by our method with only the masked reconstruction loss of \Cref{eq:masked_LDM_loss} (top row), and after adding the cross-attention loss of \Cref{eq:attention_loss} (bottom row). Adding the cross-attention loss encourages each of the handles \tokena and \tokenb to attend only to its corresponding concept, which results with a disentangled concepts' generation.}
    \label{fig:cross_attention_comparison}
\end{figure}

%% file: sections/experiments.tex
\section{Experiments}
\label{sec:experiments}

\input{figures/quantitative_comparison/fig.tex}
\input{figures/dataset_synthesis/fig.tex}
\input{figures/automatic_dataset_comparison/fig_ablation_mini.tex}

This section begins by adapting current text-to-image personalization methods to suit our single-image problem setting, followed by a qualitative comparison with our method. Next, we establish an automatic pipeline to evaluate the effectiveness of our method and compare it quantitatively to the baseline methods. Additionally, a user study is conducted to substantiate the claim that our method outperforms the baselines. Finally, we explore several applications, demonstrating the versatility and usefulness of our approach.

\subsection{Comparisons}
\label{sec:comparisons}

Existing personalization methods, such as DreamBooth \cite{Ruiz2022DreamBoothFT} and Textual Inversion \cite{Gal2022AnII}, take multiple images as input, rather than a single image with masks indicating the target concepts. Applying such methods to a single image without such indication results in tokens that do not necessarily correspond to the concepts that we wish to learn. For more details and examples, please see the supplementary material.

Thus, in order to conduct a meaningful comparison with these previous methods, we first adapt them to our problem setting. This is achieved by converting a single input image with several concept masks into a small collection of image-text pairs, as shown in \Cref{fig:dataset_synthesis}. Specifically, each pair is constructed by randomly choosing a subset of concepts $i_1,\ldots,i_k$, and placing them on a random solid color background with a random flip augmentation. The text prompt accompanying each such image is ``a photo of $[v_{i_1}]$ and ... $[v_{i_k}]$''.
We refer to DB and TI trained on such image collections as \maskedDB and \maskedTI, respectively.

Another personalization approach, Custom Diffusion \cite{Kumari2022MultiConceptCO} (CD), optimizes only the cross-attention weights of the denoising model, as well as a newly-added text token. We adapt CD to our problem setting using the same approach as above, and refer to the adapted version as \maskedCD. In addition, ELITE \cite{Wei2023ELITEEV}, trains encoders on a \emph{single image} to allow fast personalization, and also supports input masks. We use the official implementation of ELITE to compare it with our method.

\paragraph{\textbf{Qualitative comparisons.}}
We start with a qualitative comparison between our method and the baselines. As demonstrated in \Cref{fig:qualitative_comparison}, \maskedTI and \maskedCD are able to generate images that follow the text prompt, but struggle with preserving the concept identities. \maskedDB preserves the identities well, but is not able to generate an image that complies with the rest of the prompt. ELITE preserves the identities better than \maskedTI and \maskedCD, but the reconstruction is still not faithful to the input image, especially when trying to generate more than one concept. Finally, our method is able to generate images that preserve the identity as well as follow the text prompt, and we demonstrate this ability with up to four different concepts in a single image.

\paragraph{\textbf{Quantitative comparisons.}}
In order to evaluate our method and the baselines quantitatively, we propose an automatic pipeline to generate a large number of inputs. As a source for these inputs, we use the COCO dataset \cite{Lin2014MicrosoftCC}, which contains images along with their instance segmentation masks. We crop COCO images into a square shape, and filter only those that contain at least two segments of distinct ``things'' type, with each segment occupying at least 15\% of the image. We also filter out concepts from COCO classes that do not have individual identities (\eg, oranges). Then, in order to create a larger dataset, we pair each of these inputs with a random text prompt and a random number of tokens, yielding a total number of 5400 image-text pairs per baseline. For more details and examples, please read the supplementary material.

For each of the baselines \maskedTI, \maskedCD, and \maskedDB, we convert each input image and masks into a small image collection, as described earlier. For ELITE, we used the official implementation that supports an input mask. Next, we generate the results for each of the input image-text pairs with all the baselines, as well as with our method.

We employ two evaluation metrics: prompt similarity and identity similarity. Prompt similarity measures the degree of correspondence between the input text prompt and the generated image. Specifically, we utilize the standard CLIP similarity metric \cite{Radford2021LearningTV}, \ie, the cosine between the normalized CLIP embeddings of the input prompt and the generated image. In each input prompt, the special $[v_i]$ tokens have been replaced with the text describing the corresponding class (\eg, ``a photo of a cat at the beach'' instead of ``a photo of \tokena at the beach'', which was used to create the image).

For the identity similarity metric, we must adapt the standard approach in order to deal with multiple subjects. A direct comparison between the input image and the generated image is bound to be imprecise, because either image may contain multiple concepts: the input image contains all the concepts, while the generated one will typically contain a subset of them. Therefore, for each generated image, we compare a masked version of the input image (using the input mask from the COCO dataset) with a masked version of the generated image. We obtained the masked version of the generated image by leveraging MaskFormer \cite{Cheng2021PerPixelCI}, a pre-trained image segmentation model.

In addition, following Ruiz \etal~\shortcite{Ruiz2022DreamBoothFT}, we chose to extract the image embeddings from the DINO model \cite{Caron2021EmergingPI}, as it was shown \cite{Ruiz2022DreamBoothFT} to better capture the identity of objects, which aligns with the goals of personalization.

As demonstrated in \Cref{fig:quantitative_comparison}(left), there is an inherent trade-off between identity similarity and prompt similarity, with \maskedDB on one end, preserving the identity well, while sacrificing prompt similarity. \maskedTI and \maskedCD are on the other end of the spectrum, exhibiting high prompt similarity but low identity preservation. It may be seen that ELITE also struggles with preservation of identities. Our method lies on the Pareto front, balancing between the two requirements.

\paragraph{\textbf{Ablation study.}}
In addition, we conducted an ablation study, which includes removing the first phase (TI) of our method, removing the masked diffusion loss in \Cref{eq:masked_LDM_loss}, removing the cross-attention loss in \Cref{eq:attention_loss}, and training the model to reconstruct a single concept at each sample, instead of union-sampling. As seen in \Cref{fig:quantitative_comparison}(left), removing the first phase causes a significant degradation in prompt similarity, as the model tends to overfit. Removing the masked loss also causes a significant prompt similarity degradation, as the model tends to learn also the background of the original image, which may override the guiding text prompt. Removing the cross-attention loss yields a degradation of identity similarities, because the model does not learn to disentangle the concepts, as explained in \Cref{sec:method}. Finally, removing union-sampling degrades the ability of the model to generate images with multiple concepts, thereby significantly reducing the identity preservation score. 

\Cref{fig:autoamtic_dataset_qualitative_comparison_ablation_mini} provides a visual comparison of the ablated cases. As can be seen, removing the first training phase causes the model to tend to ignore the target prompt. Removing the masked loss causes the model to extract elements from the background together with the masked concepts (note the vertical wooden poles present in the generated images). Removing the cross-attention loss causes the model to entangle between the concepts (the orange juice glass appears, even when the prompt only asks for the bear). When training without union sampling, the model struggles when asked generating more than one concept. For additional visual examples, please refer to the supplementary materials.

\paragraph{\textbf{User study.}}
Lastly, we conducted a user study using the Amazon Mechanical Turk (AMT) platform. We chose a random subset of the automatically generated inputs from COCO, and asked the evaluators to rate the identity preservation and the prompt similarity of each result on a Likert scale of 1--5. When rating the prompt similarity, evaluators were presented with the input text prompt where the special $[v_i]$ tokens have been replaced with the text describing the corresponding class (as we did for the automatic prompt similarity metric). The results of our method and all the baselines were presented on the same page, and the evaluators were asked to rate each of the images. For identity preservation, we showed a masked version of the input image, containing only the object being generated, next to each of the results, and asked the evaluator to rank on the scale of 1--5 whether the images contain the same object as in the masked input image. For more details and statistical significance analysis, read the supplementary materials. As can be seen in \Cref{fig:quantitative_comparison}(right), the human rankings provide an additional evidence that our method lies on the Pareto front, balancing identity preservation and prompt similarity.

\subsection{Applications}
\label{sec:applications}

In \Cref{fig:applications} we present several applications and use cases demonstrating the versatility of our method.

\paragraph{\textbf{Image variations.}}
Given a single image containing multiple concepts of interest, once these are extracted using our method, 
they can be used to
generate multiple variations of the image by simply prompting the model with ``a photo of \tokena and \tokenb and ...''. As demonstrated in \Cref{fig:applications}(a), the arrangement of the objects in the scene, as well as the background, are different in each generation.

\paragraph{\textbf{Entangled concept separation.}}
Given a single image with composite objects, one can decompose such objects into distinct concepts. For example, as shown in \Cref{fig:applications}(b), given a single image of a dog wearing a shirt, our method is able to separately learn the dog and the shirt concepts. Thus, it is possible to generate images of the dog without the shirt, or of a cat wearing that specific shirt. Note how the dog's body is not visible in the input image, yet the strong priors of the diffusion model enable it to generate a plausible body to go with the dog's head.

\paragraph{\textbf{Background extraction.}}
In addition to learning various foreground objects in the scene, we also learn the background as one of the visual concepts. The background mask is automatically defined as the complement of the union of all the input masks. As demonstrated in \Cref{fig:applications}(c), the user can extract the specific beach from the input image, and generate new objects on it. Please notice the correct water reflections of the newly generated objects. We used the same technique in \Cref{fig:teaser} (the white porcupine example). Note that this application is different from inpainting, as the model learns to generate variants of the background, e.g., the clouds in \Cref{fig:applications}(c) and the subtle background change in \Cref{fig:teaser}.

\paragraph{\textbf{Local image editing.}}
Once concepts have been extracted, one may utilize an off-the-shelf text-driven local image editing method in order to edit other images, \eg, Blended Latent Diffusion \cite{avrahami2022blended,avrahami2022blendedlatent}. This is demonstrated in \Cref{fig:applications}(d): after extracting the concepts from the input scene of \Cref{fig:teaser}, one may provide an image to edit, indicate the regions to be edited, and provide a guiding text prompt for each region. Then, by using Blended Latent Diffusion, we can embed the extracted concepts inside the indicated regions, while preserving the rest of the image. For more details on this approach, please refer to the supplementary material. This application is reminiscent of exemplar-based image editing methods \cite{Yang2022PaintBE, Song2022ObjectStitchGO} with two key differences: (1) our single example image may contain multiple concepts, and (2) we offer an additional fine-grained textual control over each of the edited regions.

%% file: figures/quantitative_comparison/fig.tex
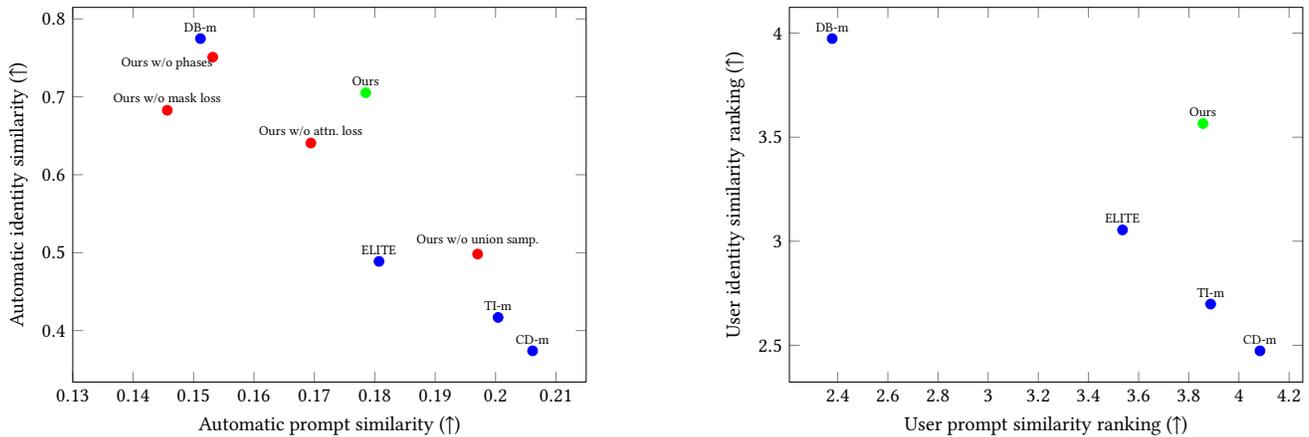
\begin{figure*}[h]
    \begin{tabular}{c @{\hspace{10\tabcolsep}} c}
        \begin{tikzpicture} [thick,scale=0.92, every node/.style={scale=0.92}]

            \def\MarkSize{.75em}
            \protected\def\ToWest#1{
              \llap{#1\kern\MarkSize}\phantom{#1}
            }
            \protected\def\ToSouth#1{
              \sbox0{#1}
              \smash{
                \rlap{
                  \kern-.5\dimexpr\wd0 + \MarkSize\relax
                  \lower\dimexpr.375em+\ht0\relax\copy0
                }
              }
              \hphantom{#1}
            }

            \begin{axis}[
                xlabel={Automatic prompt similarity ($\uparrow$)}, 
                ylabel={Automatic identity similarity ($\uparrow$)},
                compat=newest,
                xmin=0.13,
                xmax=0.215,
                width=9cm,
                height=7cm
            ]
                \addplot[
                    scatter/classes={a={blue}, b={red}, c={green}},
                    scatter,
                    mark=*, 
                    only marks, 
                    scatter src=explicit symbolic,
                    nodes near coords*={\Label},
                    visualization depends on={value \thisrow{label} \as \Label}
                ] table [meta=class] {
                    x y class label
                    0.2004144923 0.4169265571 a \scriptsize{\maskedTI}
                    0.1511466953 0.7747014945 a \scriptsize{\maskedDB}
                    0.206127391 0.3740189431 a \scriptsize{\maskedCD}
                    0.1807104829 0.4888174028 a \scriptsize{ELITE}
                    0.1694222071 0.6406772811 b \scriptsize{Ours w/o attn. loss}
                    0.1456436986 0.6827682648 b \scriptsize{Ours w/o mask loss}
                    0.1970327033 0.4982020243 b \scriptsize{Ours w/o union samp.}
                    0.153174692 0.7509141581 b \ToSouth{\scriptsize{Ours w/o phases}}
                    0.1785079577 0.7051955834 c \scriptsize{Ours}
                };
            \end{axis}
        \end{tikzpicture}
        &

        \begin{tikzpicture} [thick,scale=0.92, every node/.style={scale=0.92}]
            \begin{axis}[
                xlabel={User prompt similarity ranking ($\uparrow$)}, 
                ylabel={User identity similarity ranking ($\uparrow$)},
                compat=newest,
                width=9cm,
                height=7cm
            ]
                \addplot[
                    scatter/classes={a={blue}, b={red}, c={green}},
                    scatter,
                    mark=*, 
                    only marks, 
                    scatter src=explicit symbolic,
                    nodes near coords*={\Label},
                    visualization depends on={value \thisrow{label} \as \Label}
                ] table [meta=class] {
                    x y class label
                    3.887242798 2.697942387 a \scriptsize{\maskedTI}
                    2.377777778 3.973662551 a \scriptsize{\maskedDB}
                    4.083950617 2.473251029 a \scriptsize{\maskedCD}
                    3.535802469 3.054320988 a \scriptsize{ELITE}
                    3.856790123 3.565432099 c \scriptsize{Ours}
                };

            \end{axis}
        \end{tikzpicture}
    \end{tabular}

    \caption{\textbf{Quantitative comparison:} %
    (Left) A scatter plot of different personalization methods in terms of prompt similarity and identity similarity, generated as described in \Cref{sec:experiments}.
    \maskedDB preserves the identities, while compromising the prompt similarity. \maskedTI and \maskedCD follow the prompt, while sacrificing identity similarity. Our method lies on the Pareto front by balancing between the two extremes. We also plot ablated versions of our method: removing the first phase of our method reduces prompt similarity, removing the masked diffusion loss significant degrades prompt similarity, while removing the cross-attention loss or union sampling both degrade identity similarity. (Right) A scatter plot of human rankings of identity and prompt similarities (collected using Amazon Mechanical Turk) exhibits similar trends.}
    \label{fig:quantitative_comparison}
\end{figure*}

%% file: figures/dataset_synthesis/fig.tex
\begin{figure}
    \centering
    \setlength{\ww}{\columnwidth}
    
    \includegraphics[width=\ww]{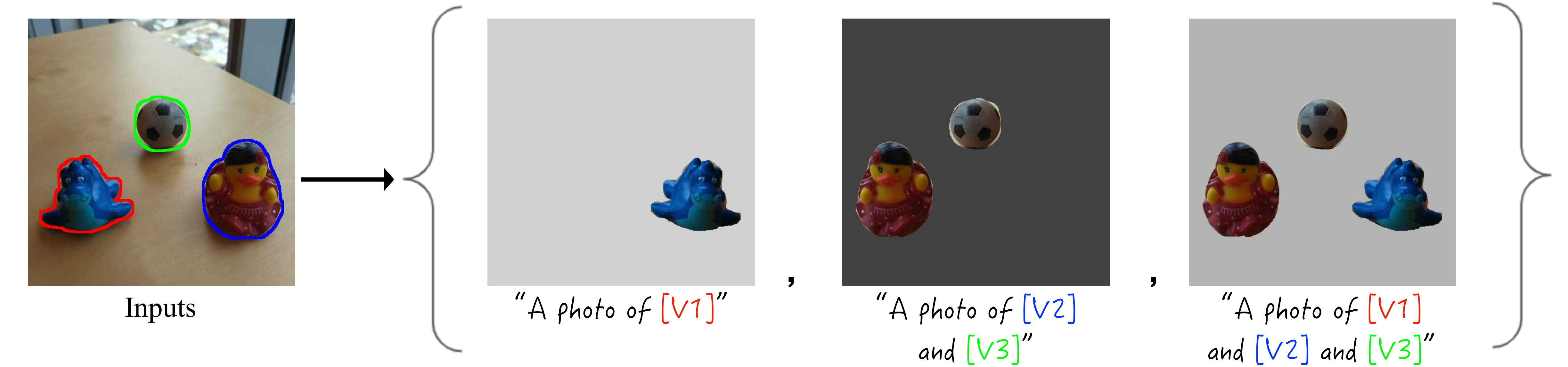}
    
    \caption{\textbf{Baseline adaptation:} Given a single image with concept masks, we construct a small collection of image-text pairs by selecting a random subset of tokens each time, creating a text prompt in the format of ``A photo of $[v_x]$ and $[v_y]$ ...'', masking out the background using the provided masks and applying a random solid background.}
    \label{fig:dataset_synthesis}
\end{figure}

%% file: figures/automatic_dataset_comparison/fig_ablation_mini.tex
\begin{figure*}[t]
    \centering
    \setlength{\tabcolsep}{1.5pt}
    \renewcommand{\arraystretch}{0.5}
    \setlength{\ww}{0.3\columnwidth}
    \begin{tabular}{c @{\hspace{10\tabcolsep}} @{\hspace{10\tabcolsep}} ccccc}

        \textbf{Inputs} &
        \textbf{Ours w/o} &
        \textbf{Ours w/o} &
        \textbf{Ours w/o} &
        \textbf{Ours w/o} &
        \textbf{Ours}
        \\

        &
        \textbf{two phases} &
        \textbf{masked loss} &
        \textbf{attention loss} &
        \textbf{union sampling} &
        \\

        \\
        \raisebox{-0.5\height}[0pt][0pt]{\frame{\includegraphics[width=\ww]{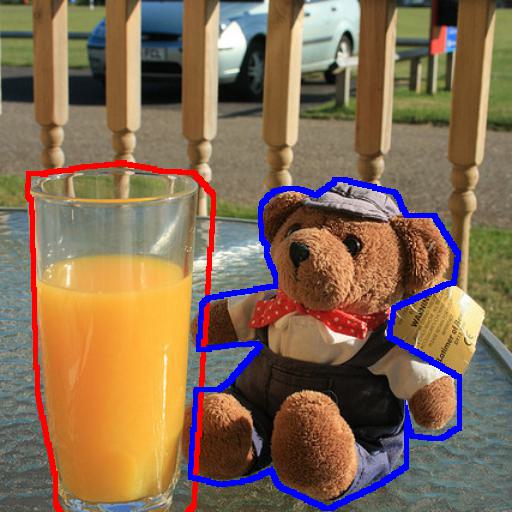}}} &
        \frame{\includegraphics[width=\ww]{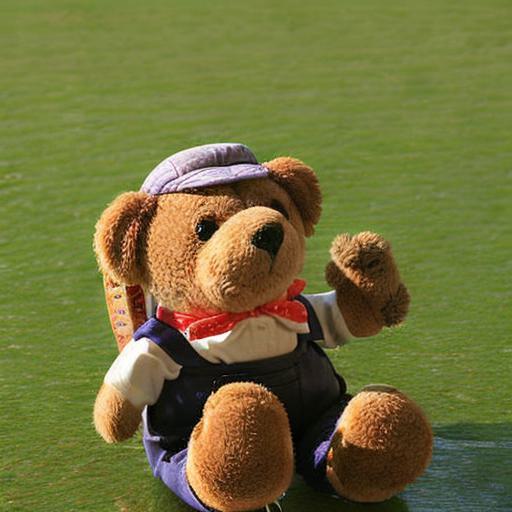}} &
        \frame{\includegraphics[width=\ww]{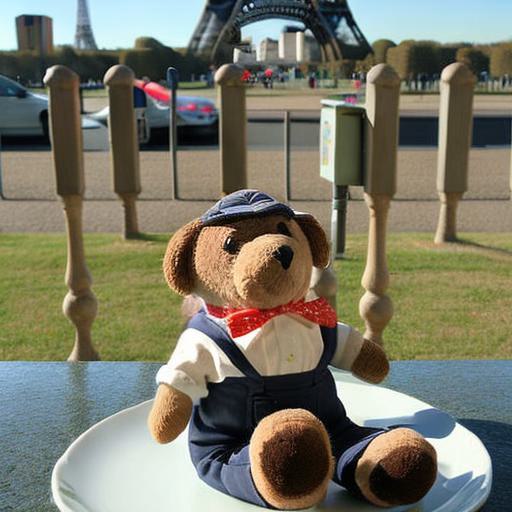}} &
        \frame{\includegraphics[width=\ww]{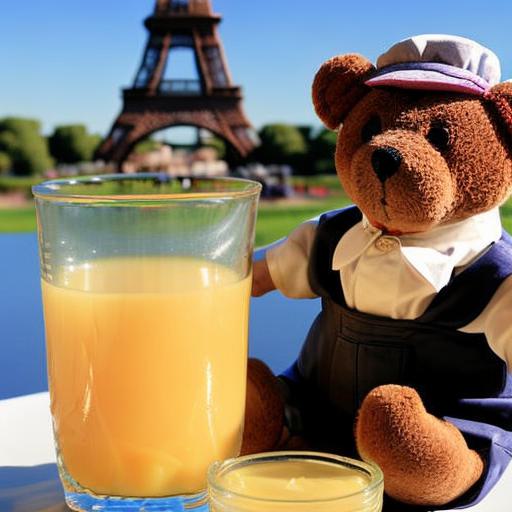}} &
        \frame{\includegraphics[width=\ww]{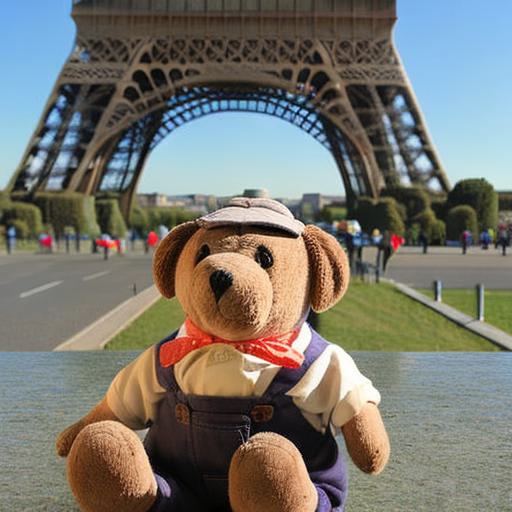}} &
        \frame{\includegraphics[width=\ww]{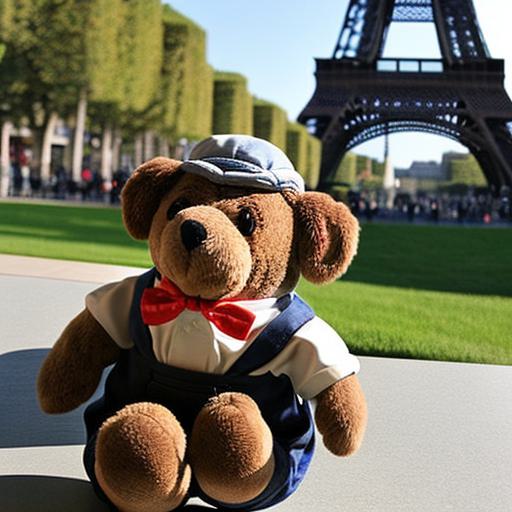}}
        \\
        \\

        &
        \multicolumn{5}{c}{``a photo of \tokenb with the Eiffel Tower in the background''}
        \\
        \\

        &
        \frame{\includegraphics[width=\ww]{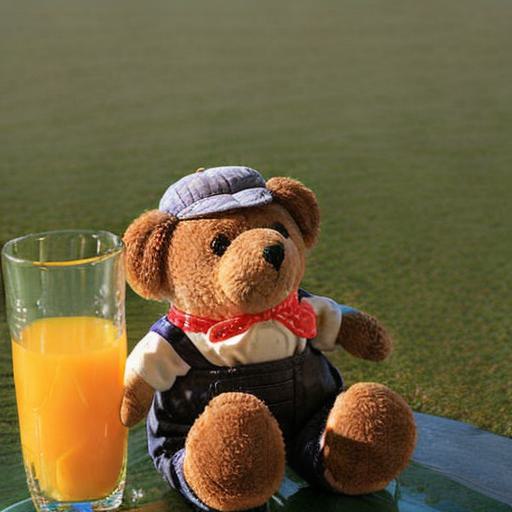}} &
        \frame{\includegraphics[width=\ww]{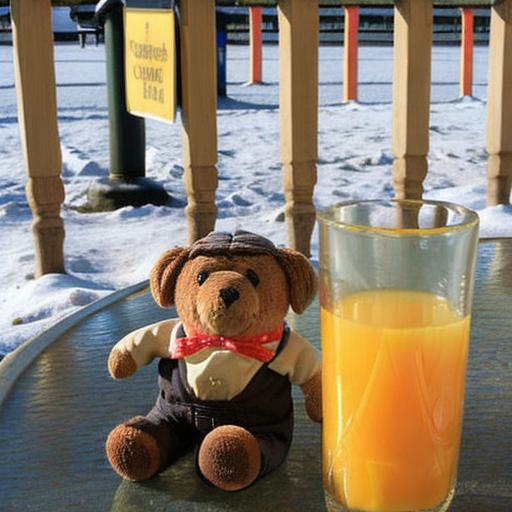}} &
        \frame{\includegraphics[width=\ww]{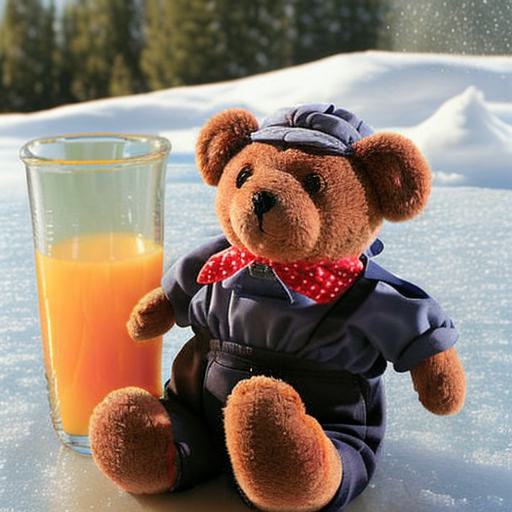}} &
        \frame{\includegraphics[width=\ww]{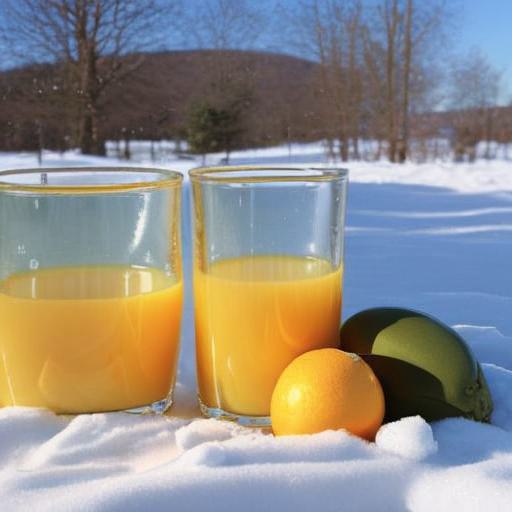}} &
        \frame{\includegraphics[width=\ww]{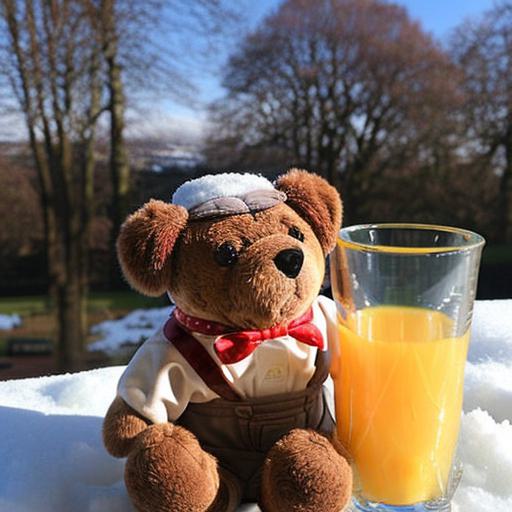}}
        \\
        \\

        &
        \multicolumn{5}{c}{``a photo of \tokena and \tokenb in the snow''}

    \end{tabular}
    
    \caption{\textbf{Qualitative ablation:} we ablate our approach by removing the first training phase, removing the masked diffusion loss, removing the cross-attention loss, and sampling a single concept at a time. As can be seen, when removing the first training phase, the model overfits and fails to follow the guiding prompt, when removing the masked loss, the model tends to learn also the background. Without the cross-attention loss, the model tends to entangle the concepts or replicate one of them. Finally, when sampling a single concept at a time, the model struggles with generating images with multiple concepts.}
    \label{fig:autoamtic_dataset_qualitative_comparison_ablation_mini}
\end{figure*}

%% file: sections/limitations.tex
\section{Limitations and Conclusions}
\label{sec:limitations}

\input{figures/limitations/fig.tex}

We found our method to suffer from the following limitations: (a) inconsistent lighting --- because the input to our method is a single image, our method sometimes struggles with disentangling the lighting from the learned identities, e.g., the input image in \Cref{fig:limitations}(a) was taken in broad daylight, and the model learns to generate the extracted concepts with daylight lighting, even when the user prompts it specifically with different environments (coral reef, dark cave and dark night). (b) Pose fixation --- another problem that stems from the single input is that sometimes the model learns to entangle between the object pose and its identity, e.g., the input image in \Cref{fig:limitations}(b) contains a dog looking upward with an open mouth, and the model generates the dog in this position in all the images, even when instructed specifically to refrain from doing so. (c) Underfitting of multiple concepts --- we found that our method works best when given up to four concepts, e.g., the input in \Cref{fig:limitations}(c) contains six objects, and the model struggles when learning that many identities. (d) Significant computational cost and parameter usage --- our method takes about 4.5 minutes to extract the concepts from a single scene and to fine-tune the entire model. Incorporating recent faster approaches that are more parameter-efficient (e.g., Custom Diffusion) did not work, which limits the applicability of this approach
in time-sensitive scenarios.
Improving the model cost is an appealing direction for further research.

In conclusion, in this paper we address the new scenario of extracting multiple concepts from a single image. We hope that it will serve as a building block for the future of the field, as generative AI continues to evolve and push the boundaries of what is possible in the realm of creative expression.

%% file: figures/limitations/fig.tex
\begin{figure}[t]
    \centering
    \setlength{\tabcolsep}{1pt}
    \renewcommand{\arraystretch}{0.5}
    \setlength{\ww}{0.24\columnwidth}
    \begin{tabular}{cc @{\hspace{2\tabcolsep}} ccc}
        \raisebox{1\height}[0pt]{\rotatebox[origin=c]{90}{\scriptsize{(a) Inconsist. Lighting}}} &
        \frame{\includegraphics[width=\ww]{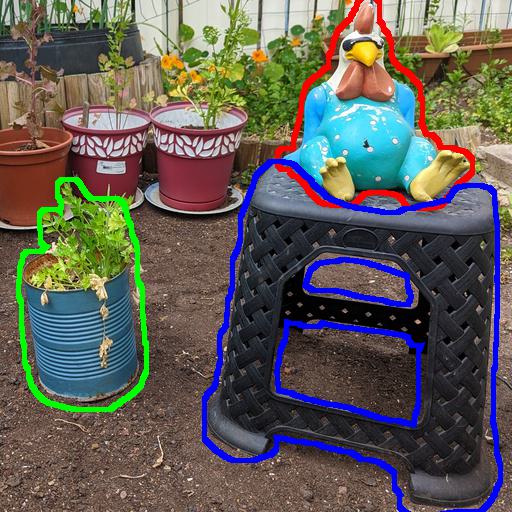}} 
        \vspace{2px} &
        \frame{\includegraphics[width=\ww]{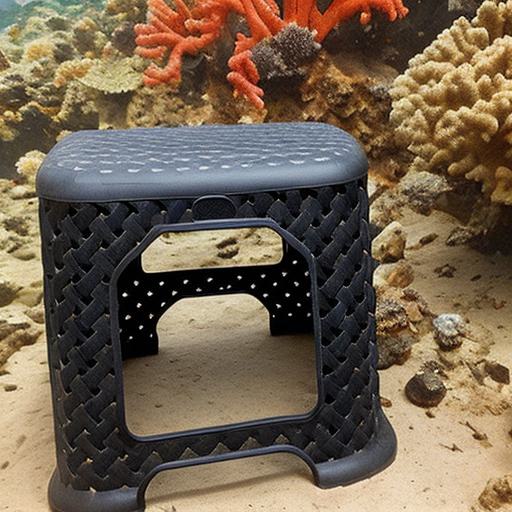}} &
        \frame{\includegraphics[width=\ww]{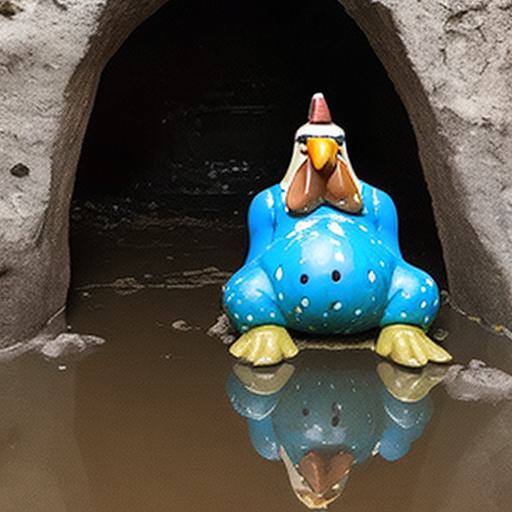}} &
        \frame{\includegraphics[width=\ww]{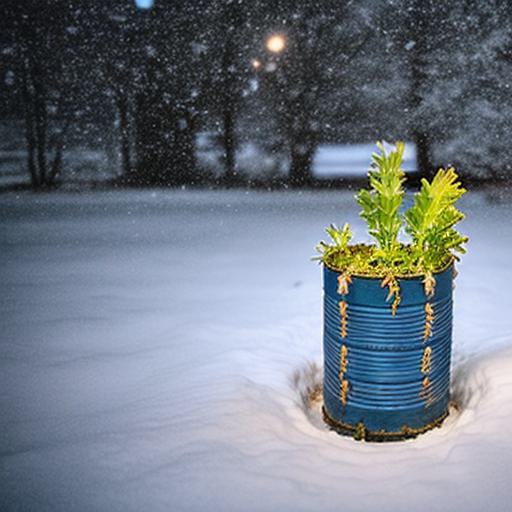}}
        \\

        &
        \scriptsize{Inputs} &
        \scriptsize{``a photo of \tokenb} &
        \scriptsize{``a photo of \tokena sitting} &
        \scriptsize{``a photo of \tokenc in a}
        \\

        &&
        \scriptsize{in a coral reef''} &
        \scriptsize{in a water puddle} &
        \scriptsize{snowy dark night''}
        \\

        &&&
        \scriptsize{inside a dark cave''} &
        
        \\
        \\
        
        \raisebox{1.5\height}[0pt]{\rotatebox[origin=c]{90}{\scriptsize{(b) Fixation}}} &
        \frame{\includegraphics[width=\ww]{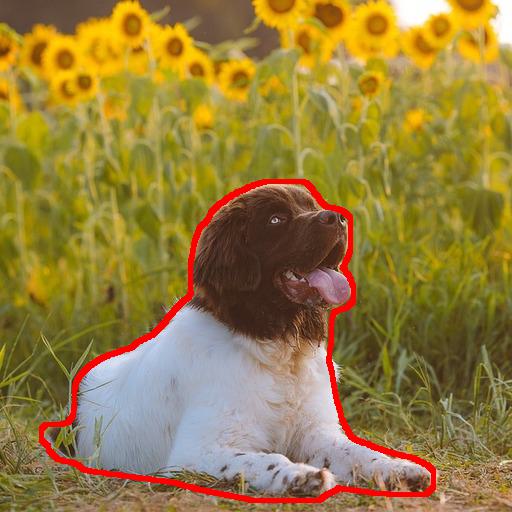}} 
        \vspace{2px} &
        \frame{\includegraphics[width=\ww]{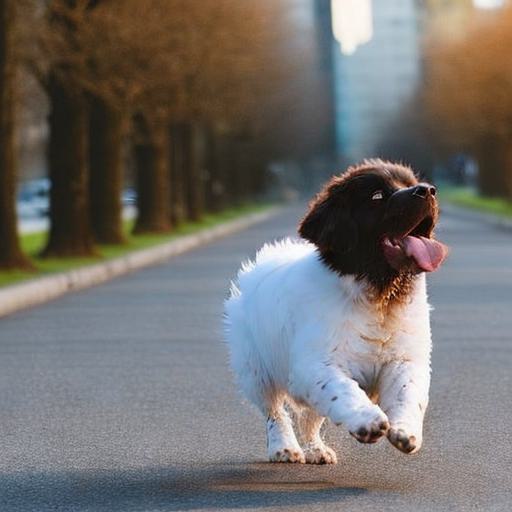}} &
        \frame{\includegraphics[width=\ww]{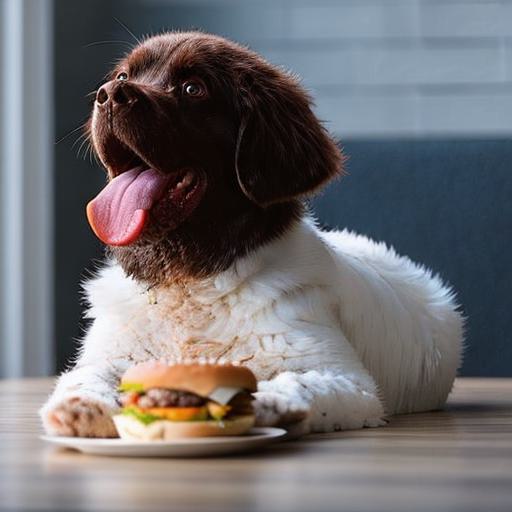}} &
        \frame{\includegraphics[width=\ww]{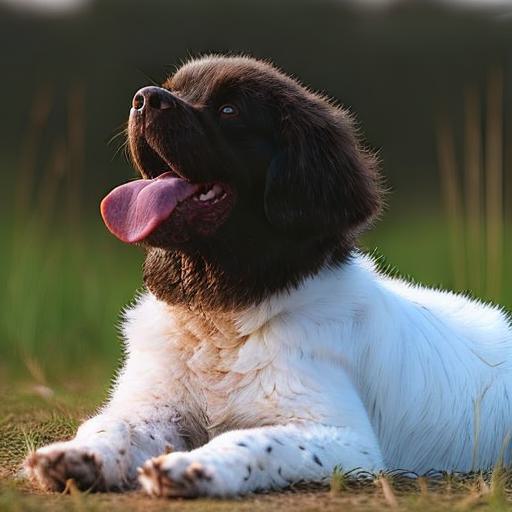}}
        \\

        &
        \scriptsize{Inputs} &
        \scriptsize{``a photo of \tokena} &
        \scriptsize{``a photo of \tokena} &
        \scriptsize{``a photo of \tokena}
        \\

        &&
        \scriptsize{running in the street''} &
        \scriptsize{eating a burger''} &
        \scriptsize{looking down with}
        \\

        &&&&
        \scriptsize{its mouth closed''}
        \\
        \\

        \raisebox{1.2\height}[0pt]{\rotatebox[origin=c]{90}{\scriptsize{(c) Many concepts}}} &
        \frame{\includegraphics[width=\ww]{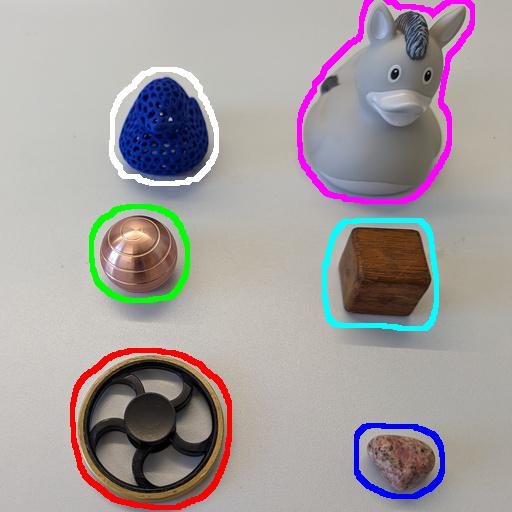}} 
        \vspace{2px} &
        \frame{\includegraphics[width=\ww]{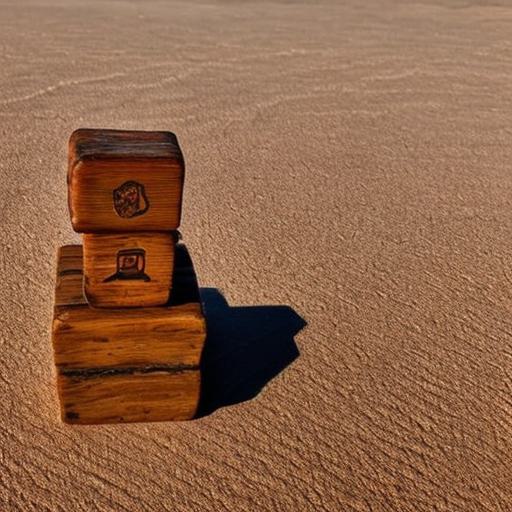}} &
        \frame{\includegraphics[width=\ww]{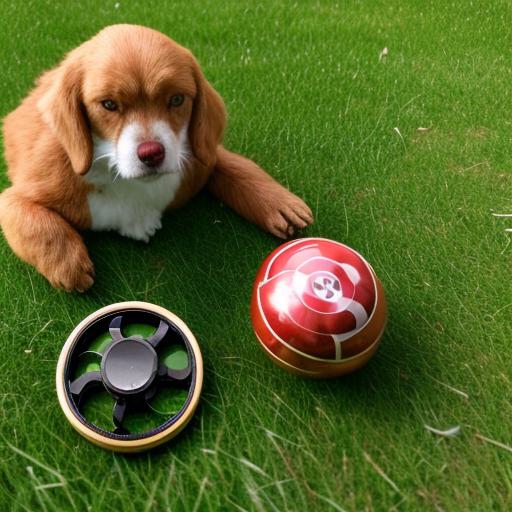}} &
        \frame{\includegraphics[width=\ww]{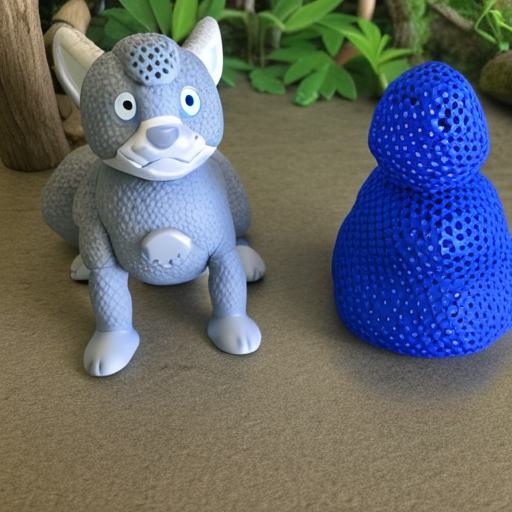}}
        \\

        &
        \scriptsize{Inputs} &
        \scriptsize{``a photo of \tokenc} &
        \scriptsize{``a photo of \tokena and} &
        \scriptsize{``a photo of \tokena and}
        \\

        &&
        \scriptsize{in the desert''} &
        \scriptsize{\tokenb on the grass''} &
        \scriptsize{\tokene and \tokenf in the}
        \\

        &&&&
        \scriptsize{the forest''}

    \end{tabular}
    
    \caption{\textbf{Limitations:} our method suffers from several limitations: (a) in some cases, the model does not learn to disentangle between the lighting of the scene in the original single image and the learned concepts, s.t. the lighting become inconsistent with the target prompt. (b) In other cases, the model learns to entangle between the pose of the objects in the single input image and their identities, s.t. it is not able to generate them with different poses, even when explicitly being told to do so. (c) We found our method to work best when used to extract up to four concepts; when trying to extract more than that, our method tends to fail in learning the objects' identities. Credits: RebaSpike @ pixabay}
    \label{fig:limitations}
\end{figure}

%% file: figures/qualitative_comparison/fig.tex
\begin{figure*}[t]
    \centering
    \setlength{\tabcolsep}{0.5pt}
    \renewcommand{\arraystretch}{0.5}
    \setlength{\ww}{0.285\columnwidth}
    \begin{tabular}{c @{\hspace{10\tabcolsep}} @{\hspace{10\tabcolsep}} ccccc}

        \textbf{Inputs} &
        \textbf{\maskedTI} &
        \textbf{\maskedDB} &
        \textbf{\maskedCD} &
        \textbf{ELITE} &
        \textbf{Ours}
        \\

        &
        \cite{Gal2022AnII} &
        \cite{Ruiz2022DreamBoothFT} &
        \cite{Kumari2022MultiConceptCO} &
        \cite{Wei2023ELITEEV}
        \\

        \\
        \raisebox{-0.6\height}[0pt][0pt]{\frame{\includegraphics[width=\ww]{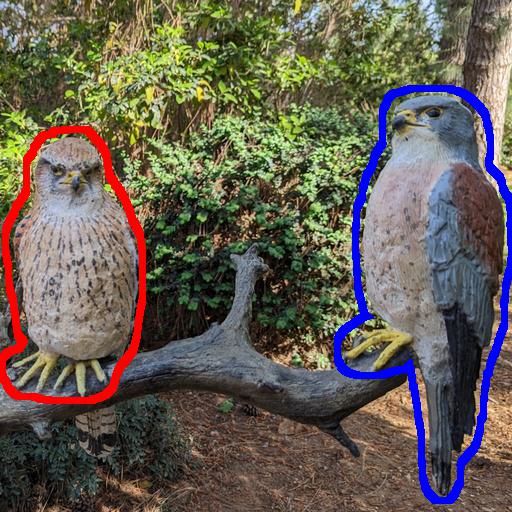}}} &
        \frame{\includegraphics[width=\ww]{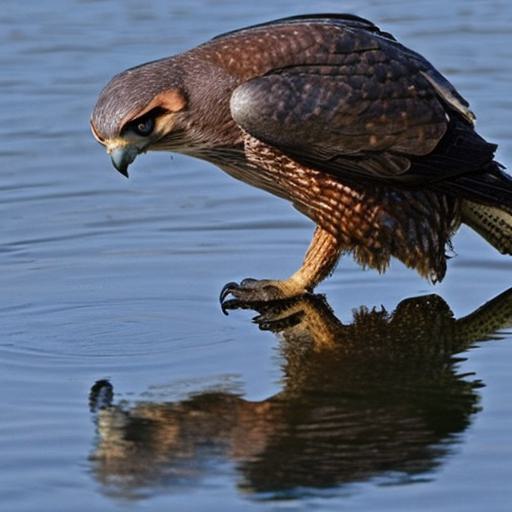}} &
        \frame{\includegraphics[width=\ww]{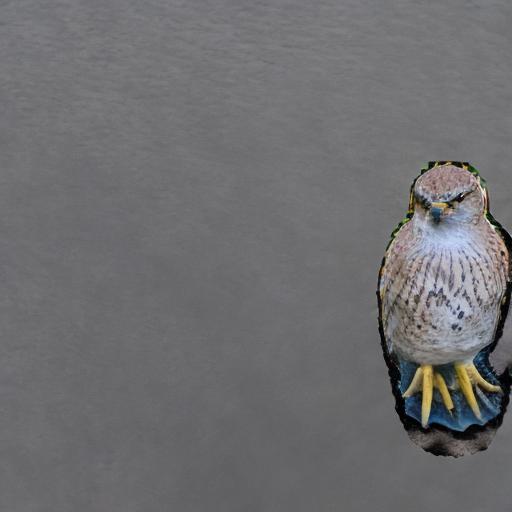}} &
        \frame{\includegraphics[width=\ww]{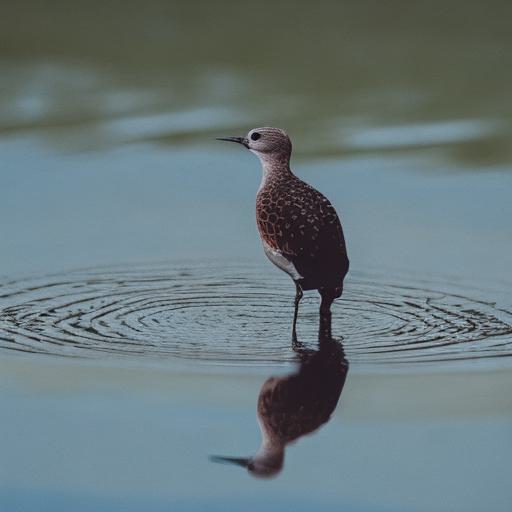}} &
        \frame{\includegraphics[width=\ww]{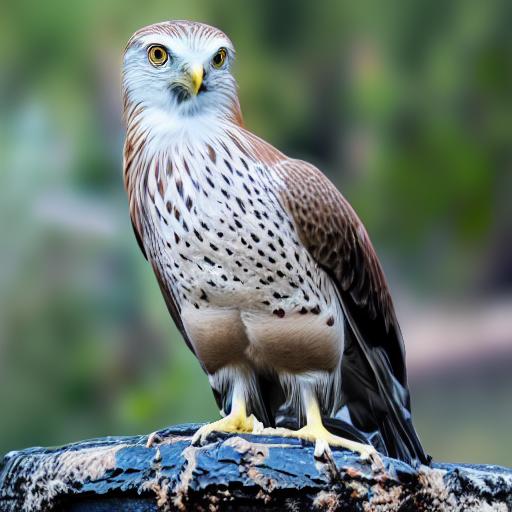}} &
        \frame{\includegraphics[width=\ww]{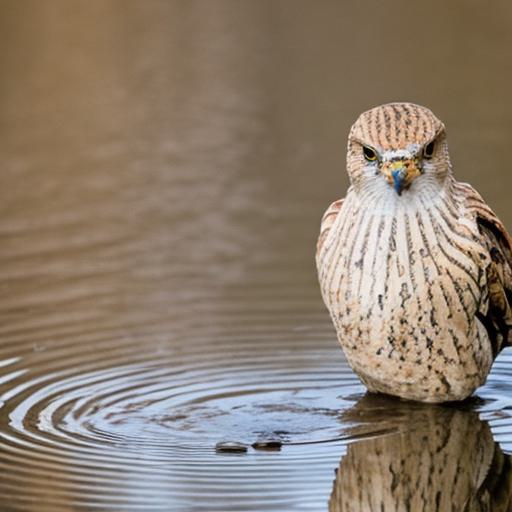}}
        \\
        \\

        &
        \multicolumn{5}{c}{``a photo of \tokena standing on top of water''}
        \\
        \\

        &
        \frame{\includegraphics[width=\ww]{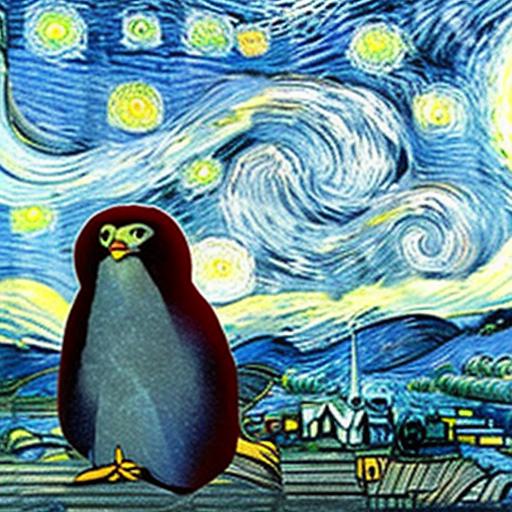}} &
        \frame{\includegraphics[width=\ww]{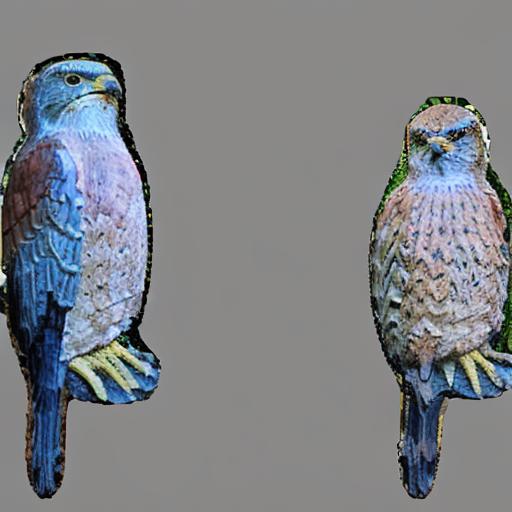}} &
        \frame{\includegraphics[width=\ww]{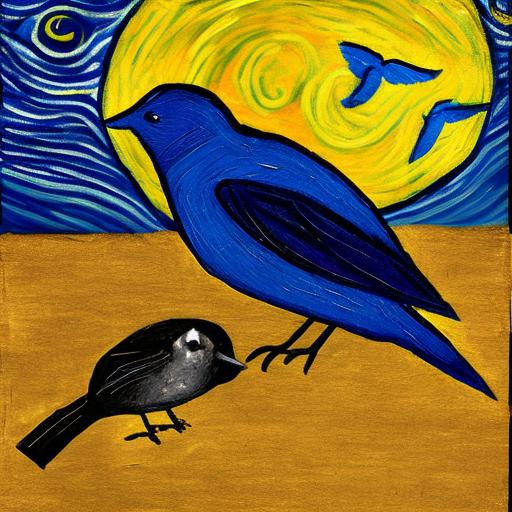}} &
        \frame{\includegraphics[width=\ww]{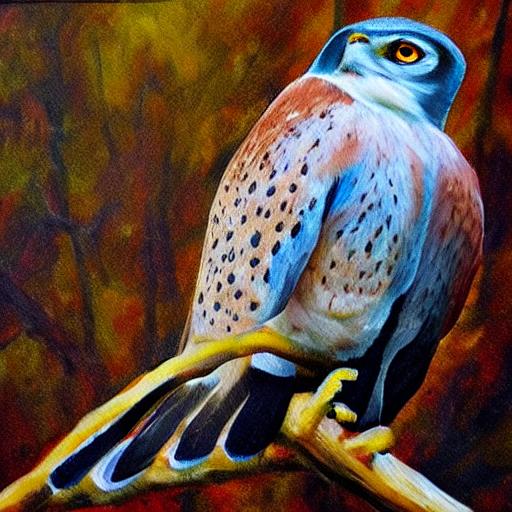}} &
        \frame{\includegraphics[width=\ww]{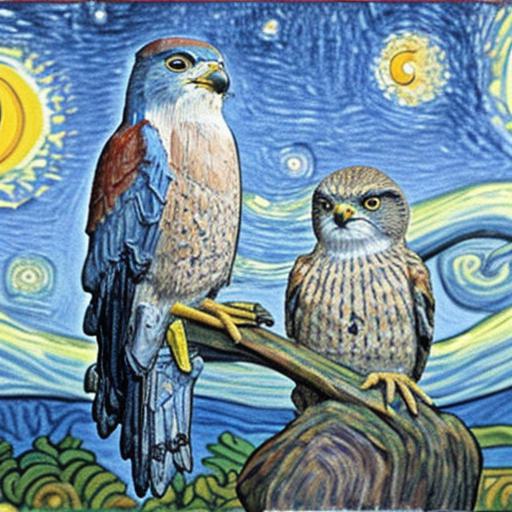}}
        \\
        \\

        &
        \multicolumn{5}{c}{``a painting of \tokena and \tokenb in the style of The Starry Night''}
        \\
        \\
        \midrule

        \\
        \raisebox{-0.6\height}[0pt][0pt]{\frame{\includegraphics[width=\ww]{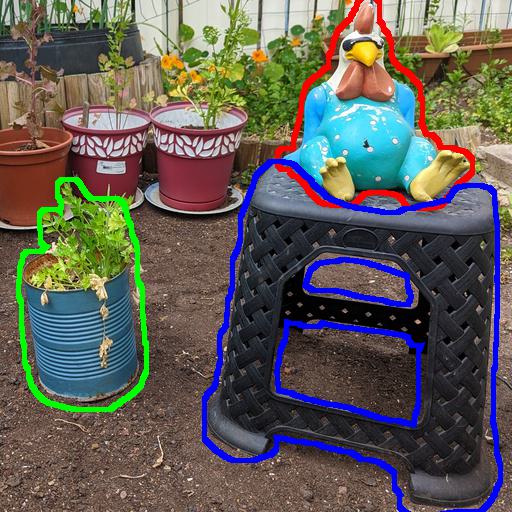}}} &
        \frame{\includegraphics[width=\ww]{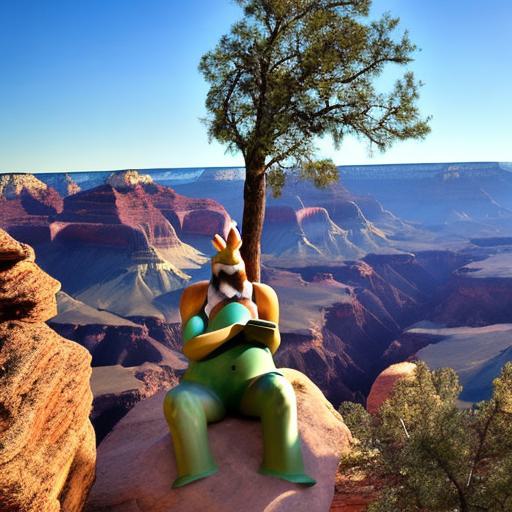}} &
        \frame{\includegraphics[width=\ww]{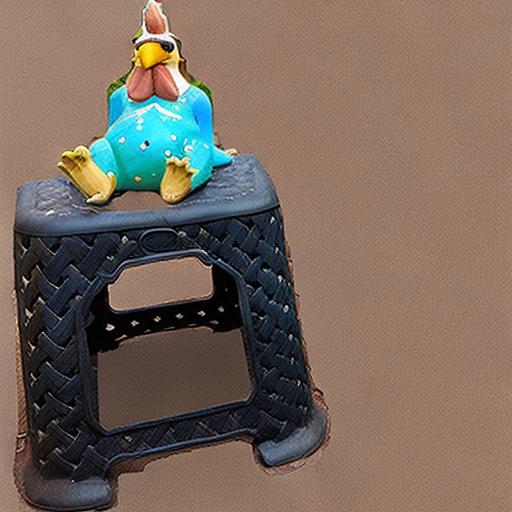}} &
        \frame{\includegraphics[width=\ww]{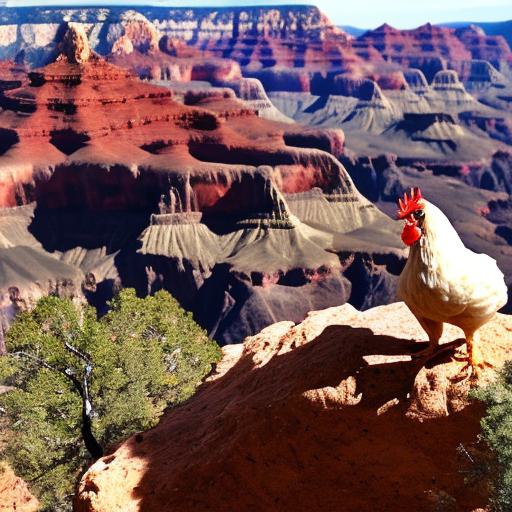}} &
        \frame{\includegraphics[width=\ww]{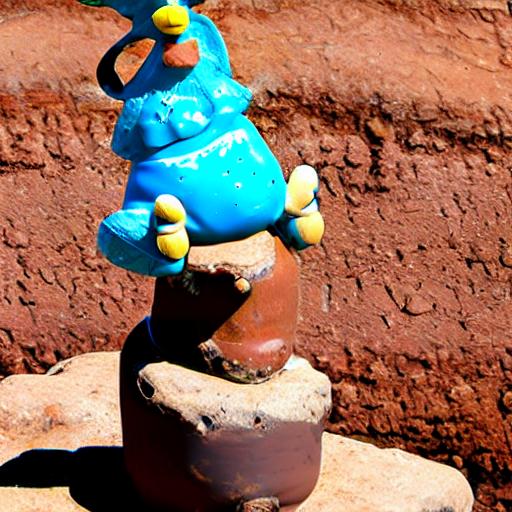}} &
        \frame{\includegraphics[width=\ww]{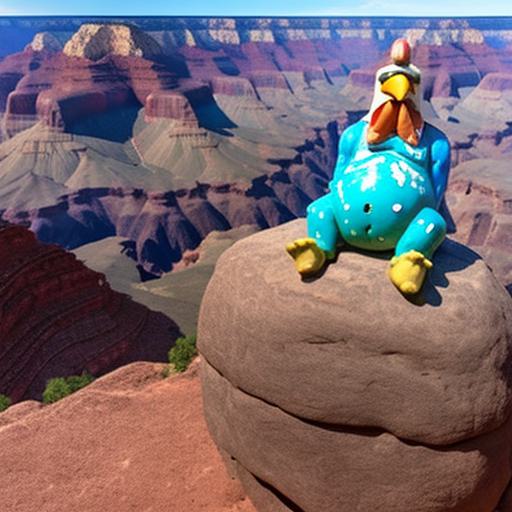}}
        \\
        \\

        &
        \multicolumn{5}{c}{``a photo of \tokena sitting on a rock in the Grand Canyon''}
        \\
        \\

        &
        \frame{\includegraphics[width=\ww]{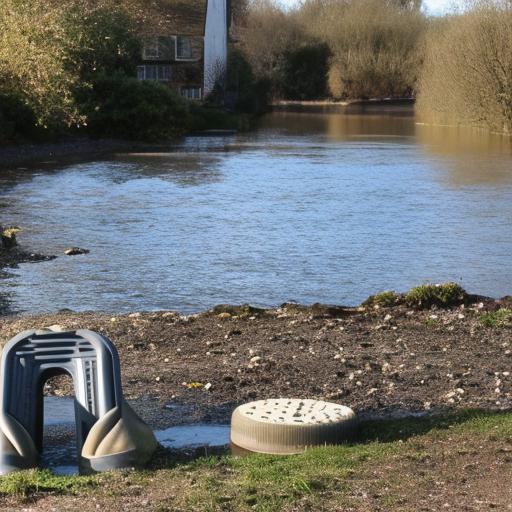}} &
        \frame{\includegraphics[width=\ww]{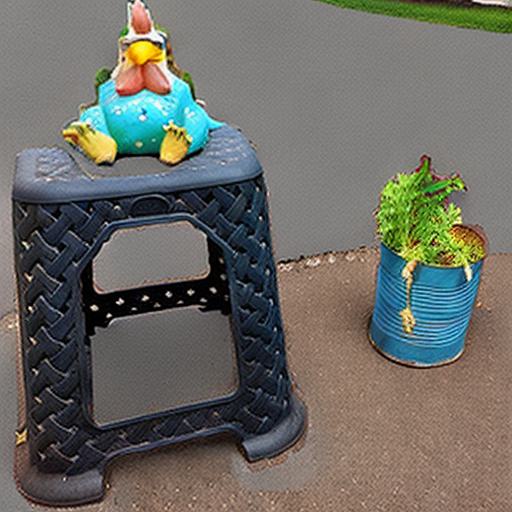}} &
        \frame{\includegraphics[width=\ww]{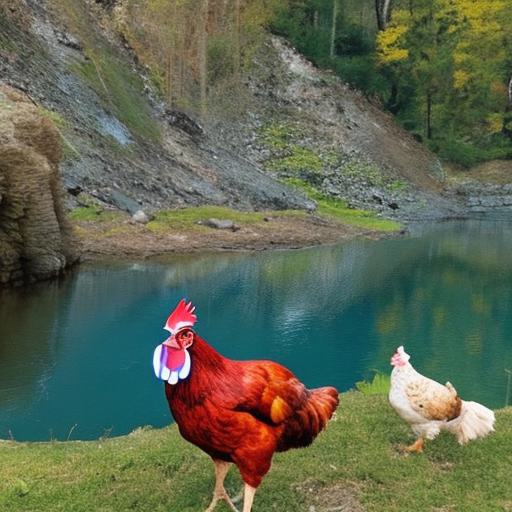}} &
        \frame{\includegraphics[width=\ww]{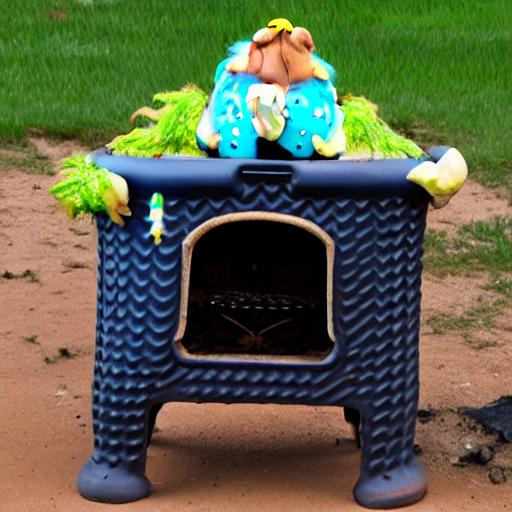}} &
        \frame{\includegraphics[width=\ww]{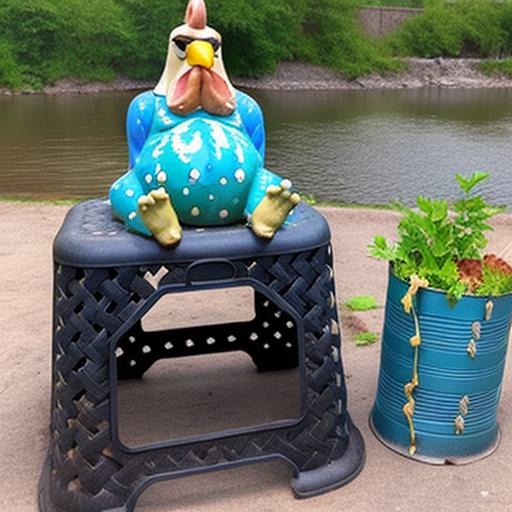}}
        \\
        \\

        &
        \multicolumn{5}{c}{``a photo of \tokena and \tokenb and \tokenc next to a river''}
        \\
        \\
        \midrule

        \\
        \raisebox{-0.6\height}[0pt][0pt]{\frame{\includegraphics[width=\ww]{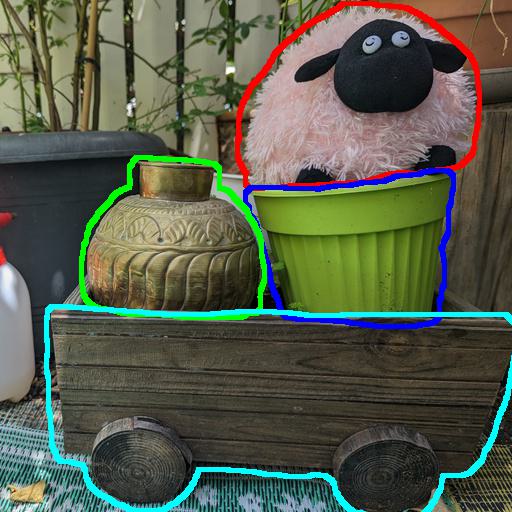}}} &
        \frame{\includegraphics[width=\ww]{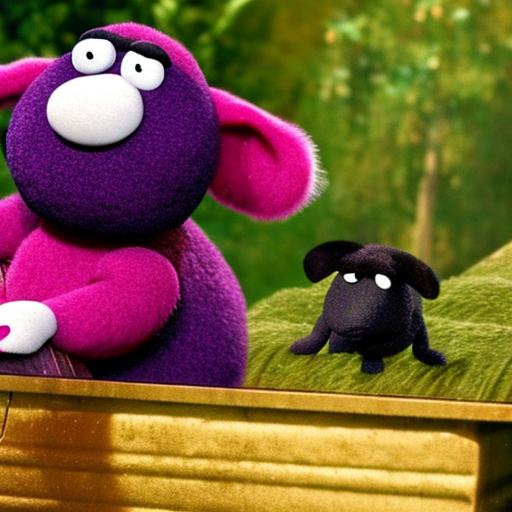}} &
        \frame{\includegraphics[width=\ww]{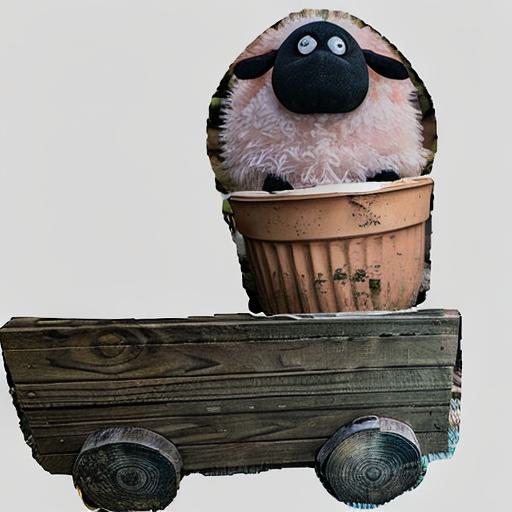}} &
        \frame{\includegraphics[width=\ww]{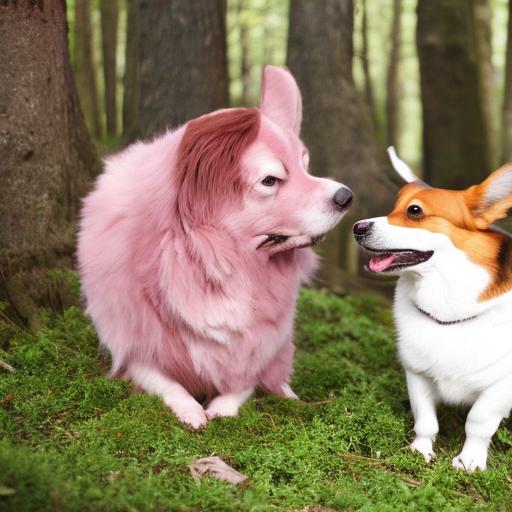}} &
        \frame{\includegraphics[width=\ww]{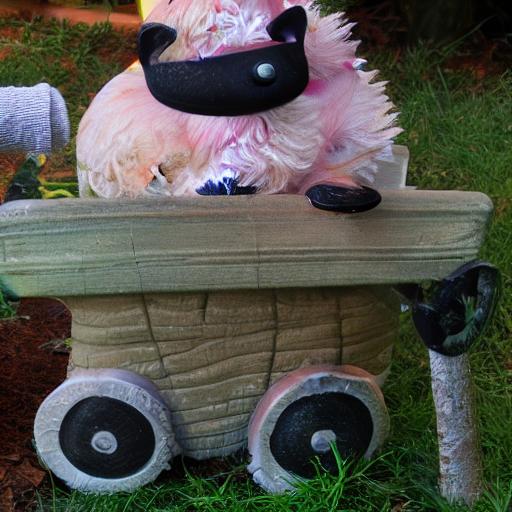}} &
        \frame{\includegraphics[width=\ww]{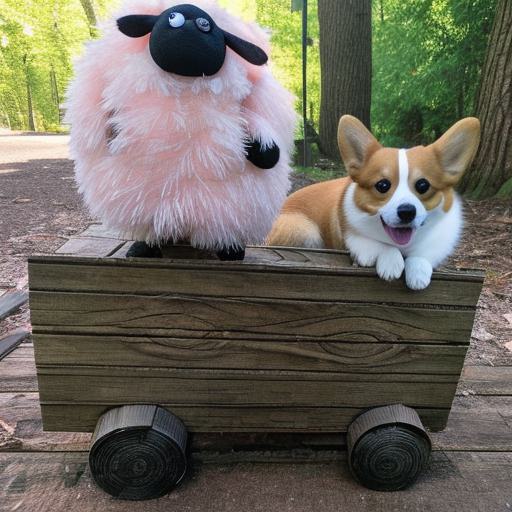}}
        \\
        \\

        &
        \multicolumn{5}{c}{``a photo of \tokena and a Corgi on \tokend in the forest''}
        \\
        \\

        &
        \frame{\includegraphics[width=\ww]{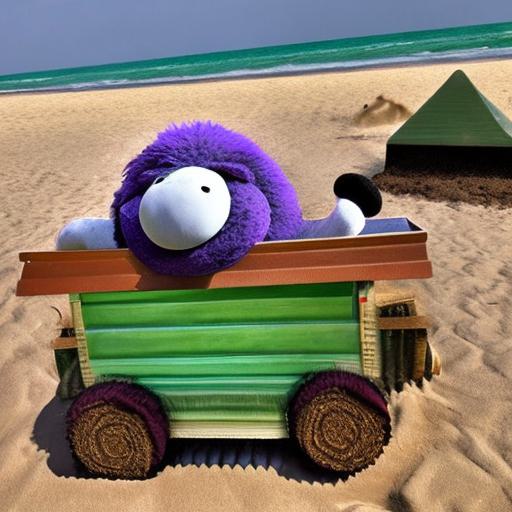}} &
        \frame{\includegraphics[width=\ww]{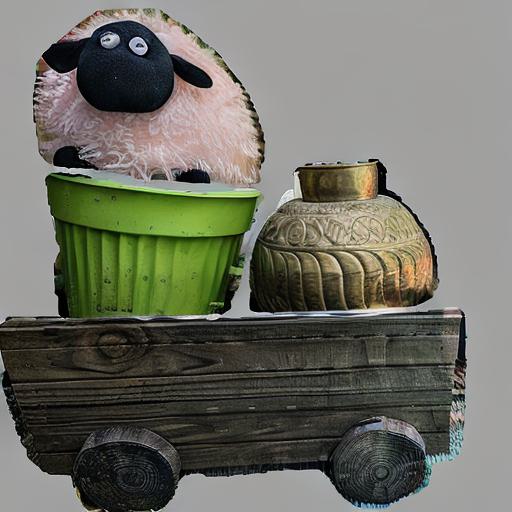}} &
        \frame{\includegraphics[width=\ww]{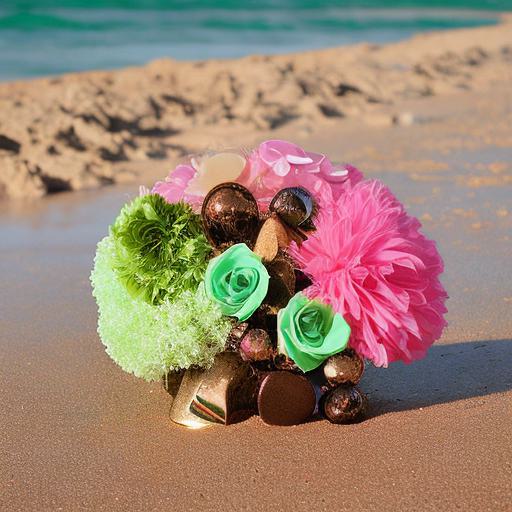}} &
        \frame{\includegraphics[width=\ww]{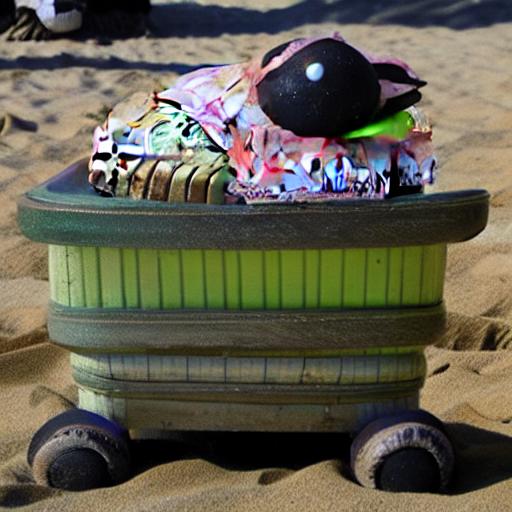}} &
        \frame{\includegraphics[width=\ww]{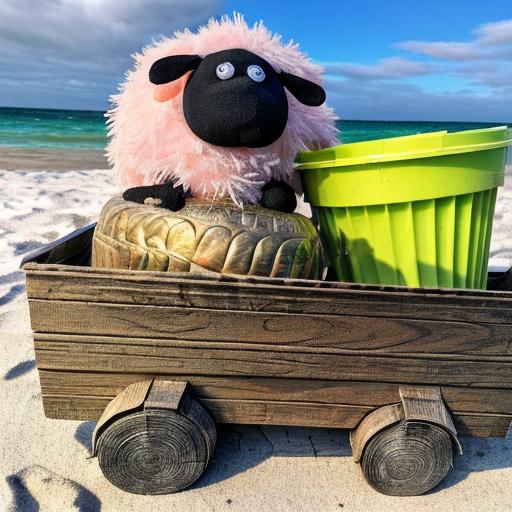}}
        \\
        \\

        &
        \multicolumn{5}{c}{``a photo of \tokena and \tokenb and \tokenc and \tokend at the beach''}

    \end{tabular}
    
    \caption{\textbf{A qualitative comparison} between several baselines and our method. \maskedTI and \maskedCD struggle with preserving the concept identities, while the images generated by \maskedDB effectively ignore the text prompt. ELITE preserves the identities better than \maskedTI/\maskedCD, but the concepts are still not recognizable enough, especially when more than one concept is generated. Finally, our method is able to preserve the identities as well as follow the text prompt, even when learning four different concepts (bottom row).}
    \label{fig:qualitative_comparison}
\end{figure*}

%% file: figures/applications/fig.tex
\begin{figure*}
    \centering
    \includegraphics[width=\linewidth]{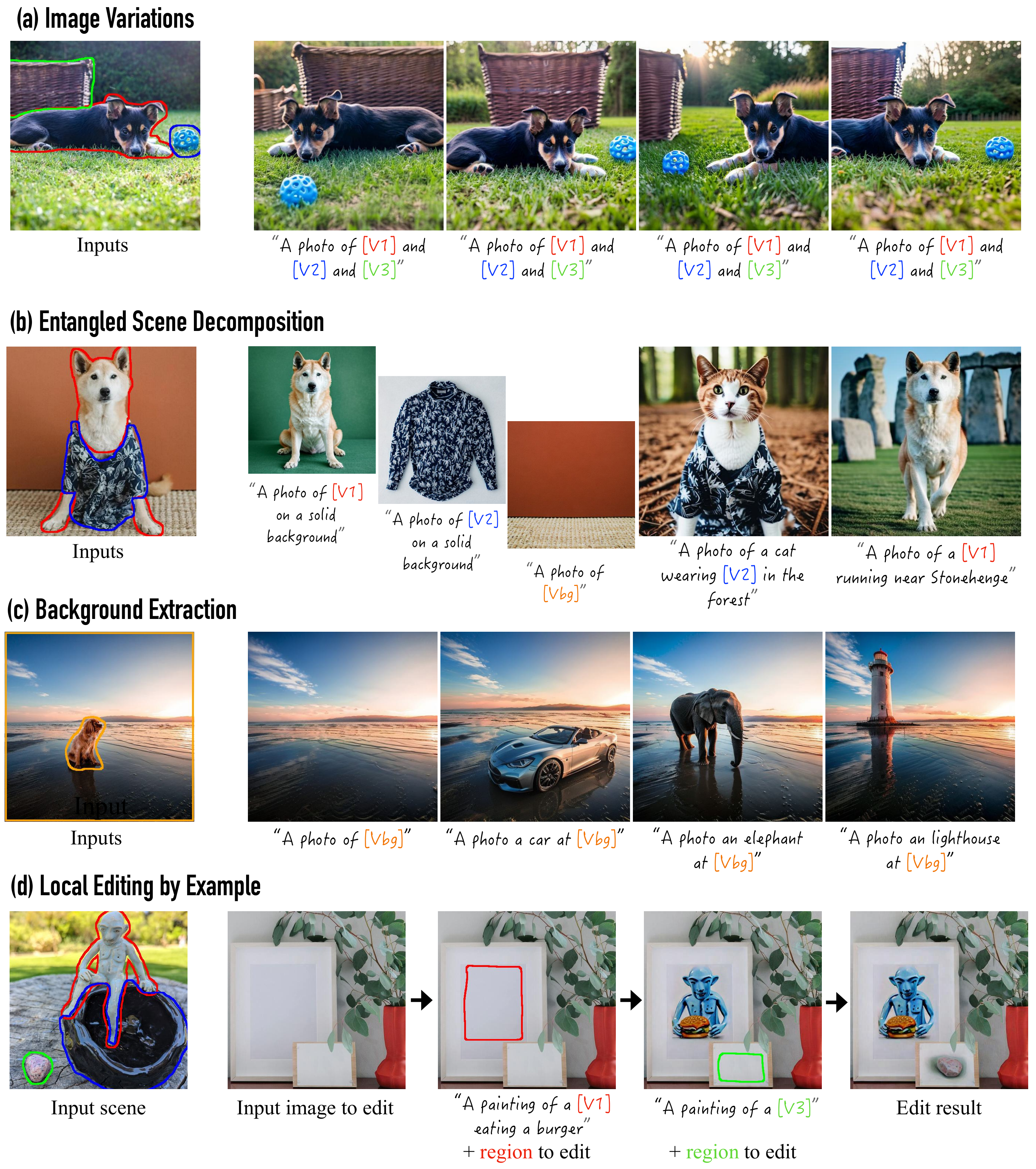} 
    \caption{\textbf{Applications:} our method can be used for other downstream tasks, such as generating image variations, decomposing entangled concepts into their components, extracting the background from an existing scene, and
    locally editing an existing image using off-the-shelf tools \cite{avrahami2022blended,avrahami2022blendedlatent}. Credits: Magda Ehlers @ pexels / Sam Lion @ pexels / pixabay / Angela Roma @ pexels}
    \label{fig:applications}
\end{figure*}

%% file: sections/appendix/additional_experiments.tex
\section{Additional Experiments}
\label{sec:additional_experiments}

\input{figures/teaser/fig_full.tex}
\input{figures/applications/fig_bld_additional.tex}
\input{figures/applications/fig_entangled_additional.tex}
\input{figures/applications/fig_variations_additional.tex}
\input{figures/naive_adaptation/fig.tex}
\input{figures/additional_qualitative_comparison/fig.tex}
\input{figures/automatic_dataset_comparison/fig.tex}
\input{figures/automatic_dataset_comparison/fig_ablation.tex}

In \Cref{sec:additional_examples} we start by providing additional results generated by our method. Then, in \Cref{sec:qualitative_ablation_study} we add additional qualitative comparisons from our ablation study. Finally, in \Cref{sec:naive_baselines} we show the results of a \naive{} application of TI \cite{Gal2022AnII} and DB \cite{Ruiz2022DreamBoothFT} to our problem setting (multiple concepts from a single image) without the adaptation discussed in our paper.

\subsection{Additional Results}
\label{sec:additional_examples}

In \Cref{fig:teaser_full} we provide additional results of breaking a scene into components and using them to re-synthesize novel images. Then, in \Cref{fig:bld_additional} we provide additional examples of the localized image editing application. Furthermore, in \Cref{fig:entangeld_scene_additional} we provide more examples of the entangled scene decomposition application. Then, in \Cref{fig:variations_additional} we provide more examples of the image variations application. Finally, in \Cref{fig:additional_qualitative_comparison} and \Cref{fig:autoamtic_dataset_qualitative_comparison} we provide additional qualitative comparisons of our method against the baselines.

\subsection{Qualitative Ablation Study Results}
\label{sec:qualitative_ablation_study}

As discussed in
Section 4.1 in the main paper,
we conducted an ablation study, which includes removing the first phase in our two-phase training scheme, removing the masked diffusion loss, removing the cross-attention loss, and removing the union-sampling. As seen in \Cref{fig:autoamtic_dataset_qualitative_comparison_ablation} when removing the first training phase, the model tends to generate images that do not correspond to the target text prompt. In addition, when removing the masked diffusion loss, the model tends to learn also the background of the original image, which overrides the target text prompt. Furthermore, when removing the cross-attention loss, the model tends to mix between the concepts or replicate one of them. Finally, removing the union-sampling degrades the ability of the model to generate images with multiple concepts. In addition, increasing the probability of only one concept during the union-sampling also has a similar effect of degrading the multiple concepts generation ability.

\subsection{\Naive{} Baselines}
\label{sec:naive_baselines}

Existing personalization methods, such as DreamBooth (DB) \cite{Ruiz2022DreamBoothFT} and Textual Inversion (TI) \cite{Gal2022AnII} take multiple images as input, rather than a single image with masks indicating the target concepts. Applying these methods to a single image without such indication results in tokens that do not necessarily correspond to the concepts that we wish to learn. In \Cref{fig:naive_adaptation} we provide a visual result of
training TI and DB on a single image with the text prompt ``a photo of \tokena and \tokenb''. As expected, these approach fails to disentangle between the concepts --- TI learns an arbitrary concept while DB overfits the input image.

%% file: figures/teaser/fig_full.tex
\begin{figure*}[t]
    \centering
    \setlength{\tabcolsep}{1.5pt}
    \renewcommand{\arraystretch}{0.5}
    \setlength{\ww}{0.4\columnwidth}
    \begin{tabular}{ccccc}
        \includegraphics[valign=c, width=\ww,frame]{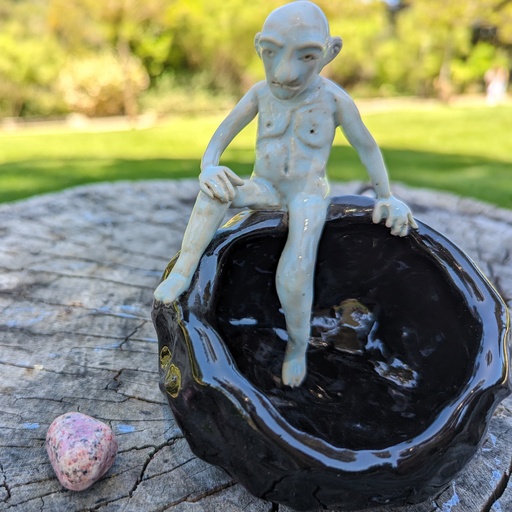} 
        \vspace{2px}&
        \includegraphics[valign=c, width=\ww,frame]{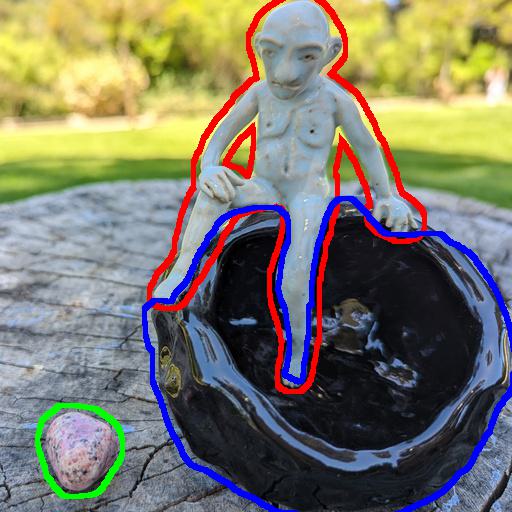} &
        \includegraphics[valign=c, width=\ww,frame]{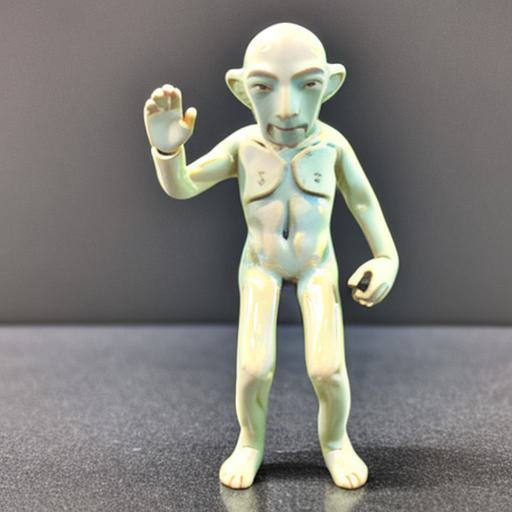} &
        \includegraphics[valign=c, width=\ww,frame]{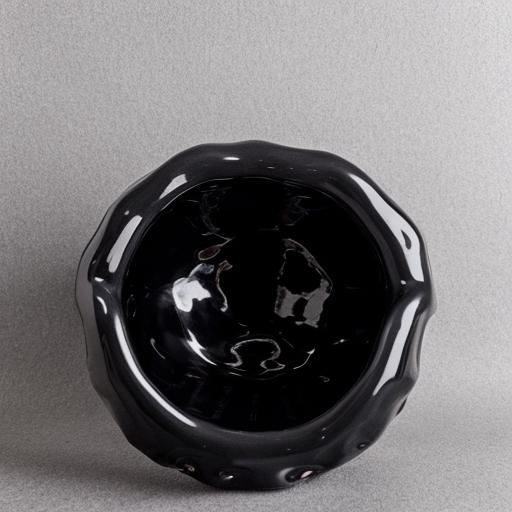} &
        \includegraphics[valign=c, width=\ww,frame]{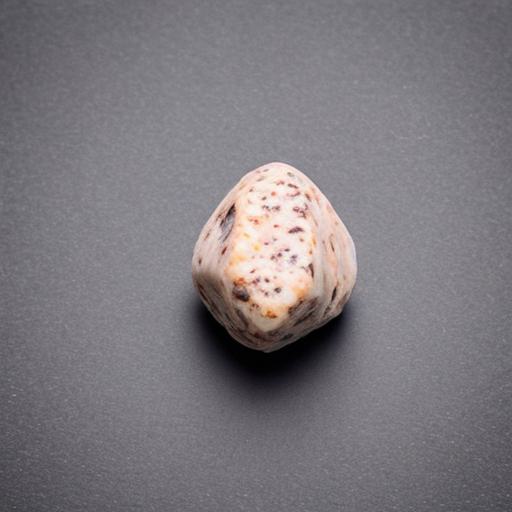}
        \\
        \emph{Single} input image &
        Input masks &
        ``A photo of \tokena on &
        ``A photo of \tokenb on &
        ``A photo of \tokenc on 
        \\
        &
        &
        a solid background'' &
        a solid background'' &
        a solid background''
        \\
        \\

        \includegraphics[valign=c, width=\ww,frame]{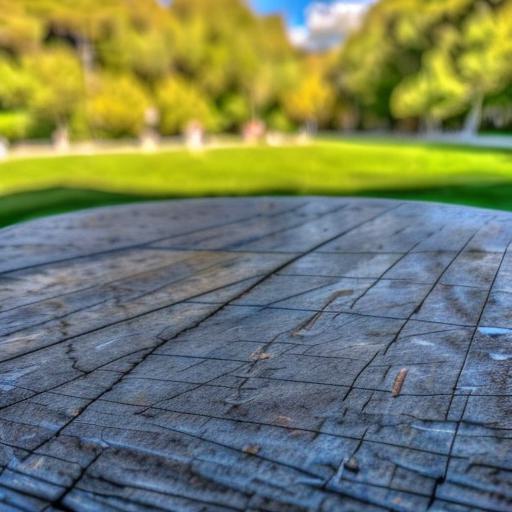} 
        \vspace{2px} &
        \includegraphics[valign=c, width=\ww,frame]{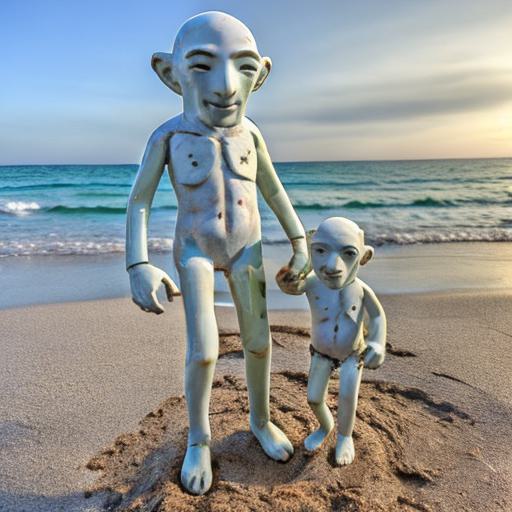} &
        \includegraphics[valign=c, width=\ww,frame]{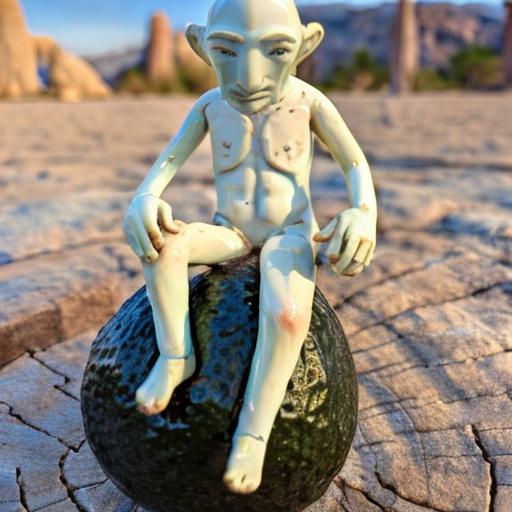} &
        \includegraphics[valign=c, width=\ww,frame]{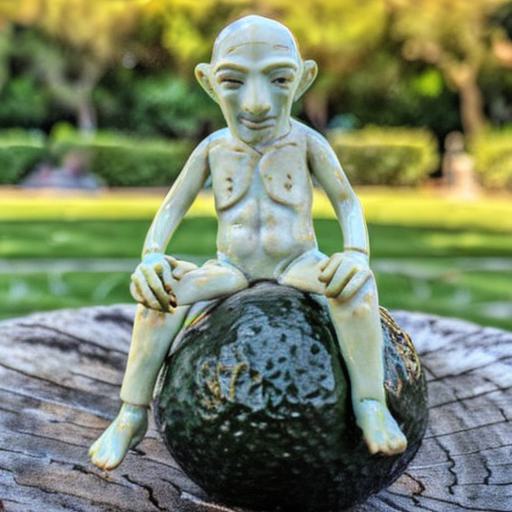} &
        \includegraphics[valign=c, width=\ww,frame]{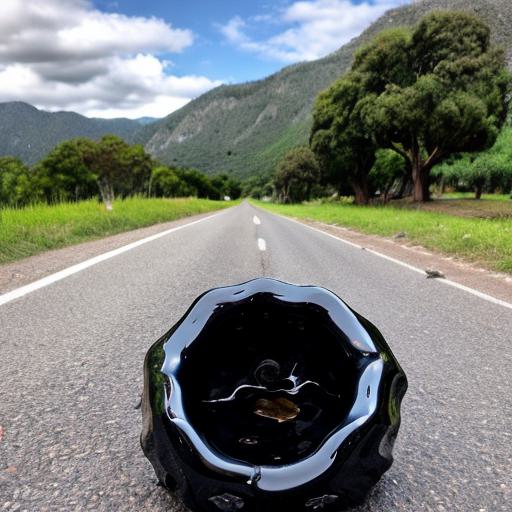}
        \\
        ``A photo of \tokenbg'' &
        ``A photo of \tokena and its &
        ``A photo of \tokena sitting on &
        ``A photo of \tokena sitting on &
        ``A photo of \tokenb on
        \\
        &
        child at the beach'' &
        an avocado in the desert'' &
        an avocado at \tokenbg'' &
        the road''
        \\
        \\

        \includegraphics[valign=c, width=\ww,frame]{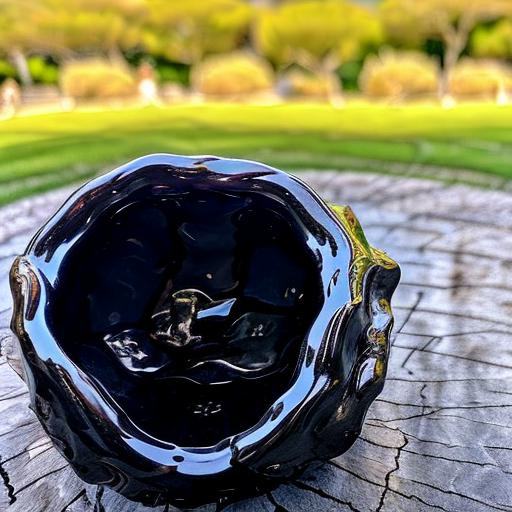} 
        \vspace{2px} &
        \includegraphics[valign=c, width=\ww,frame]{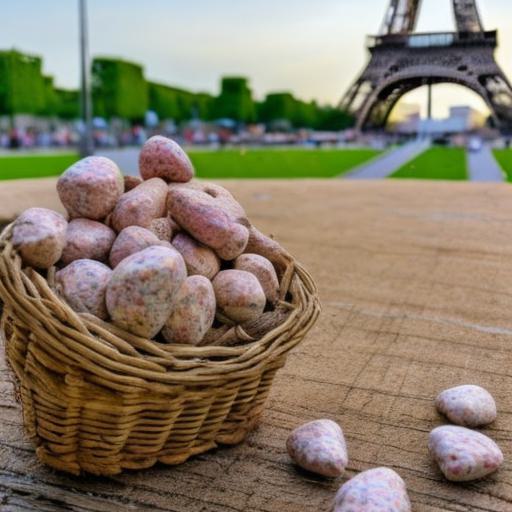} &
        \includegraphics[valign=c, width=\ww,frame]{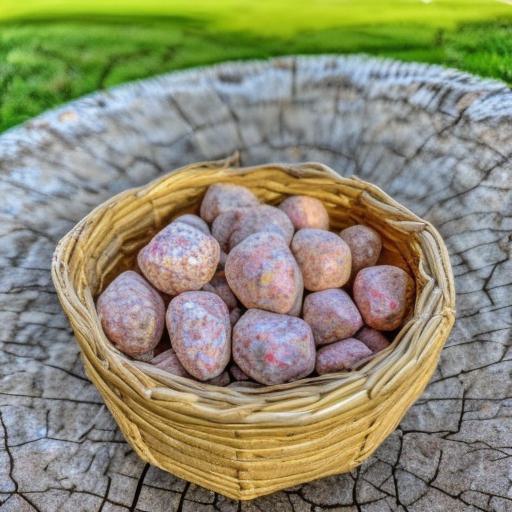} &
        \includegraphics[valign=c, width=\ww,frame]{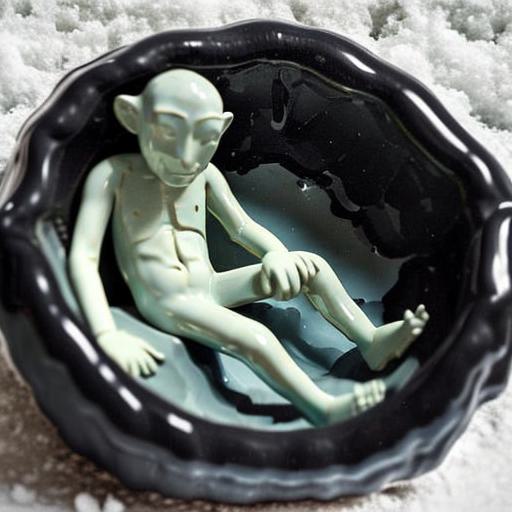} &
        \includegraphics[valign=c, width=\ww,frame]{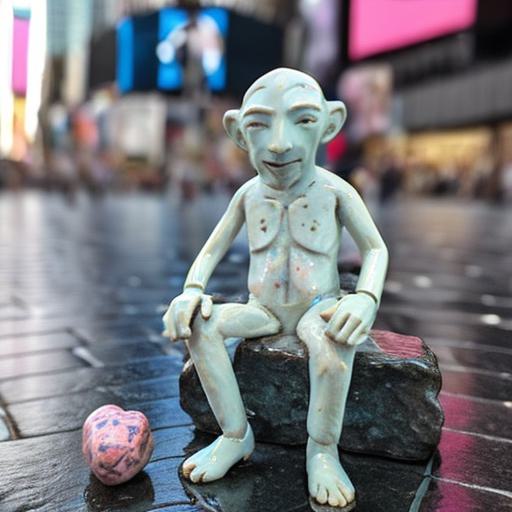}
        \\
        ``A photo of \tokenb &
        ``A photo of a pile of \tokenc &
        ``A photo of a pile of \tokenc &
        ``A photo of \tokena sleeping &
        ``A photo of \tokena and
        \\
        at \tokenbg'' &
        in a straw basket near &
        in a straw basket &
        inside \tokenb in the snow'' &
        \tokenc at Times Square''
        \\
        &
        the Eiffel Tower'' &
        at \tokenbg''
        \\
        \\

        \includegraphics[valign=c, width=\ww,frame]{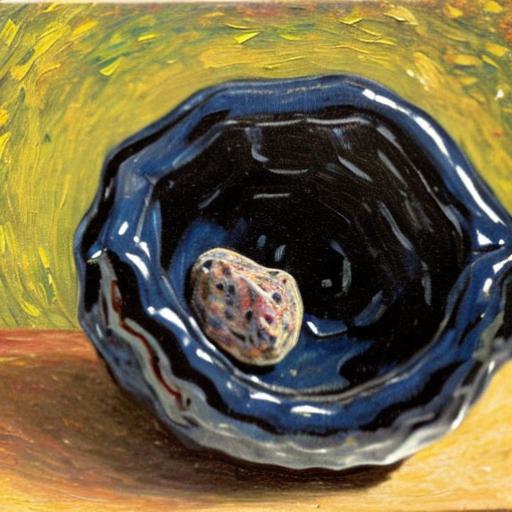} 
        \vspace{2px} &
        \includegraphics[valign=c, width=\ww,frame]{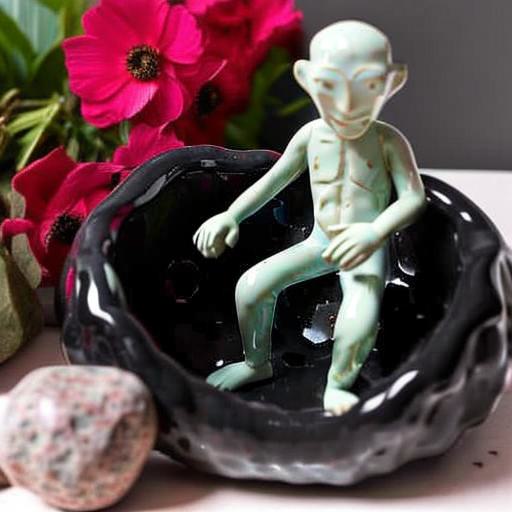} &
        \includegraphics[valign=c, width=\ww,frame]{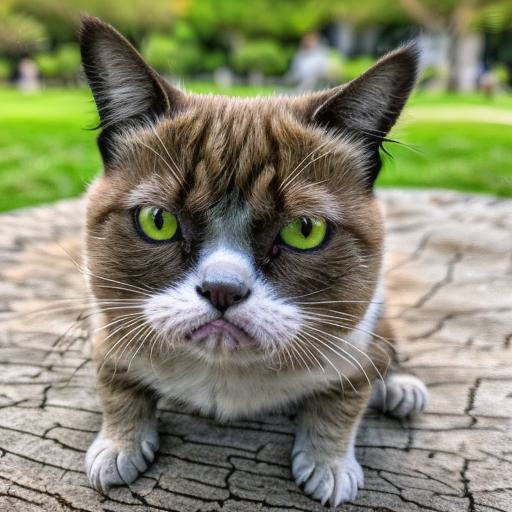} &
        \includegraphics[valign=c, width=\ww,frame]{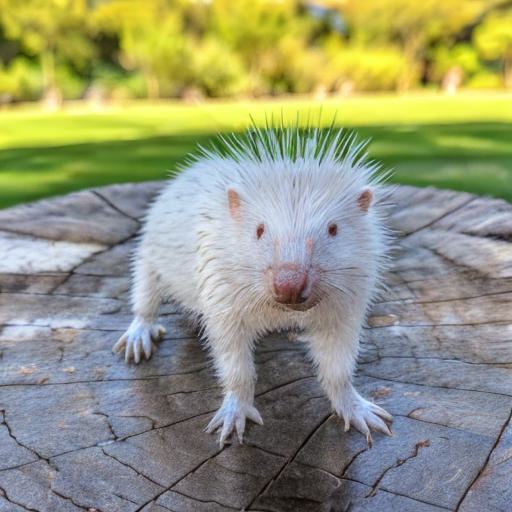} &
        \includegraphics[valign=c, width=\ww,frame]{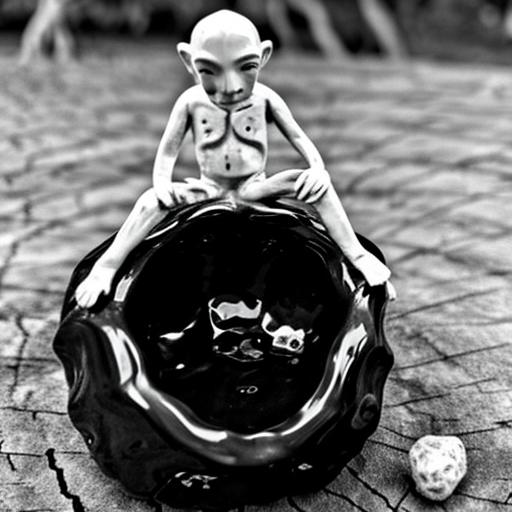}
        \\
        ``A painting of \tokenc &
        ``A photo of \tokena and \tokenb &
        ``A photo of a grumpy &
        ``A photo of a small &
        ``A black and white photo
        \\
        inside \tokenb &
        and \tokenc with flowers &
        cat at \tokenbg'' &
        albino porcupine at \tokenbg'' &
        of \tokena and \tokenb and \tokenc
        \\
        &
        in the background''
        &&&
        at \tokenbg''
    \end{tabular}
    \caption{\textbf{Additional break-a-scene results:} a scene decomposed into 3 parts and a background, which are then re-synthesized in different contexts and combinations.}
    \label{fig:teaser_full}
\end{figure*}
  

%% file: figures/applications/fig_bld_additional.tex
\begin{figure*}[th]
    \centering
    \setlength{\tabcolsep}{0.5pt}
    \renewcommand{\arraystretch}{0.5}
    \setlength{\ww}{0.4\columnwidth}
    \begin{tabular}{cccccc}
        \includegraphics[valign=c, width=\ww,frame]{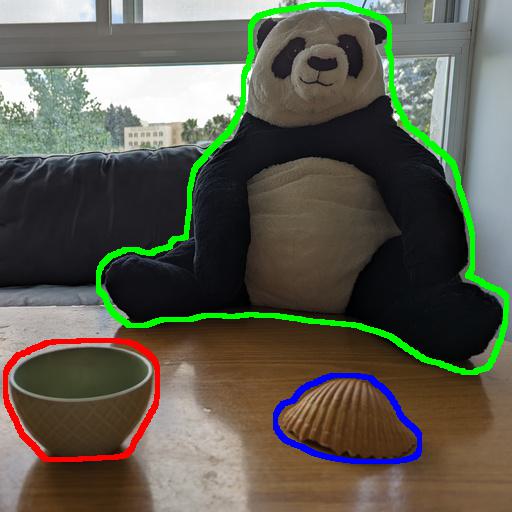} 
        \vspace{2px} &
        \includegraphics[valign=c, width=\ww,frame]{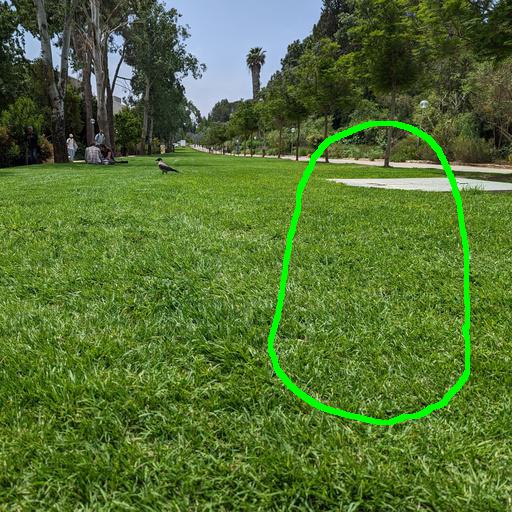} &
        \includegraphics[valign=c, width=\ww,frame]{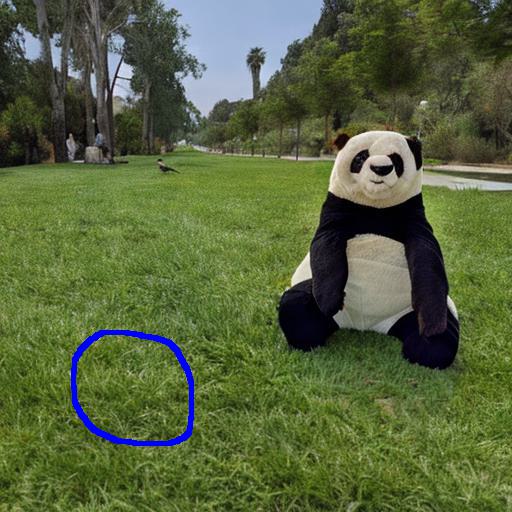} &
        \includegraphics[valign=c, width=\ww,frame]{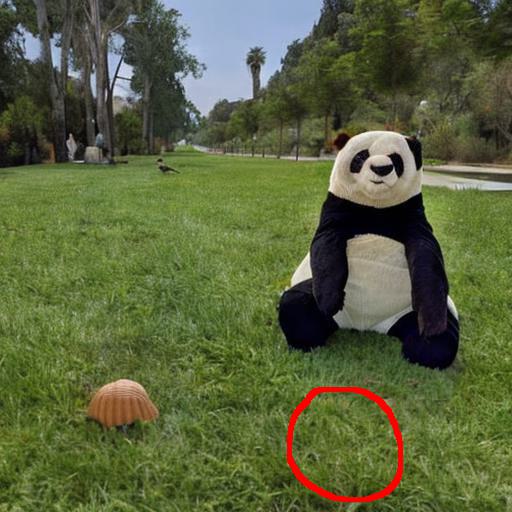} &
        \includegraphics[valign=c, width=\ww,frame]{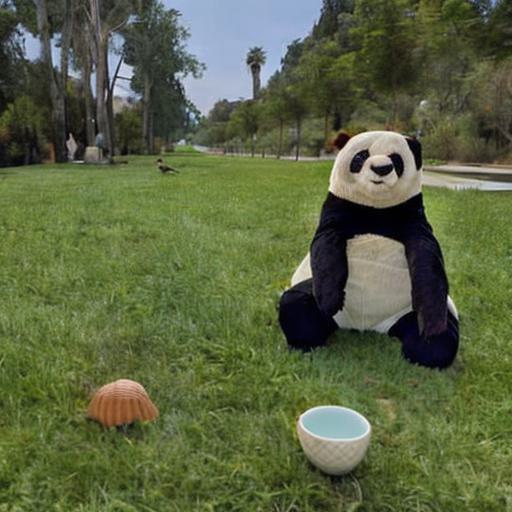}
        \\

        (a) Input scene &
        (b) Input image to edit &
        (c) Edit result 1 &
        (d) Edit result 2 &
        (e) Final result
        \\

        &
        + ``a photo of \tokenc'' &
        + ``a photo of \tokenb'' &
        + ``a photo of \tokena'' &
        \\

        &
        + mask 1 &
        + mask 2 &
        + mask 3 &
        \\
        \\

        \includegraphics[valign=c, width=\ww,frame]{figures/qualitative_comparison/assets/chicken/mask_overlay.jpg} 
        \vspace{2px} &
        \includegraphics[valign=c, width=\ww,frame]{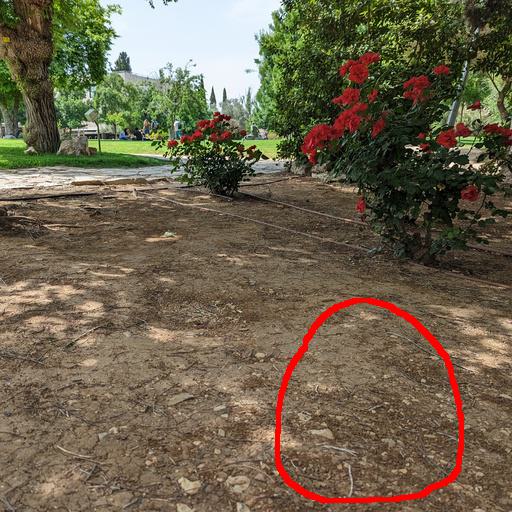} &
        \includegraphics[valign=c, width=\ww,frame]{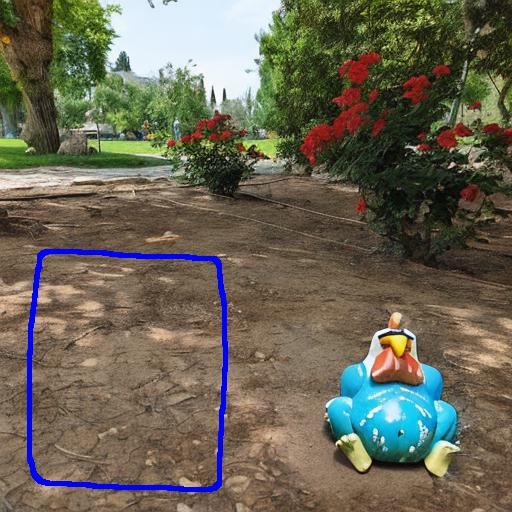} &
        \includegraphics[valign=c, width=\ww,frame]{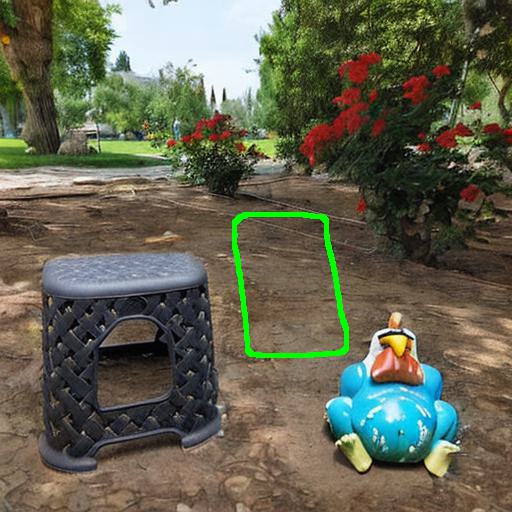} &
        \includegraphics[valign=c, width=\ww,frame]{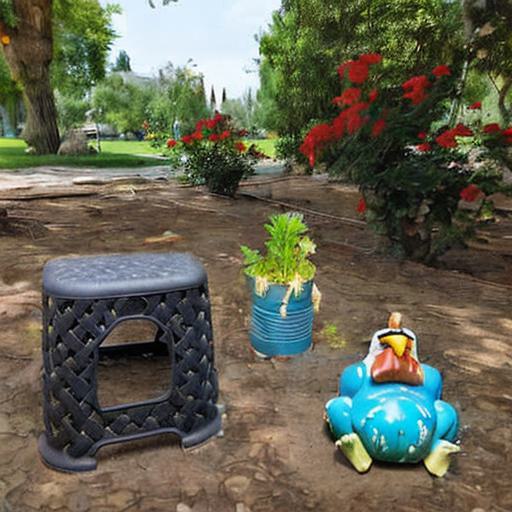}
        \\

        (a) Input scene &
        (b) Input image to edit &
        (c) Edit result 1 &
        (d) Edit result 2 &
        (e) Final result
        \\

        &
        + ``a photo of \tokena'' &
        + ``a photo of \tokenb'' &
        + ``a photo of \tokenc'' &
        \\

        &
        + mask 1 &
        + mask 2 &
        + mask 3 &
        \\
        \\

        \includegraphics[valign=c, width=\ww,frame]{figures/qualitative_comparison/assets/sheep/mask_overlay.jpg} 
        \vspace{2px} &
        \includegraphics[valign=c, width=\ww,frame]{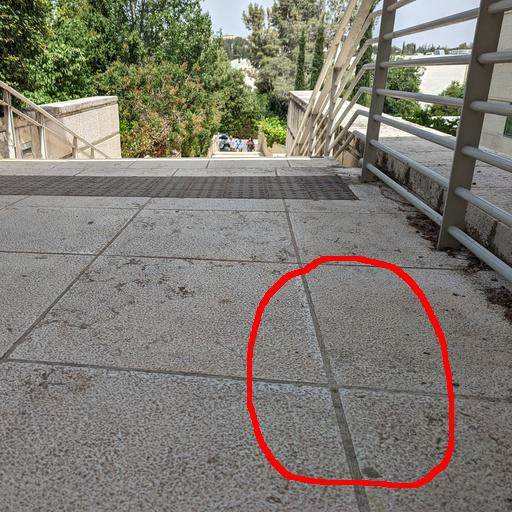} &
        \includegraphics[valign=c, width=\ww,frame]{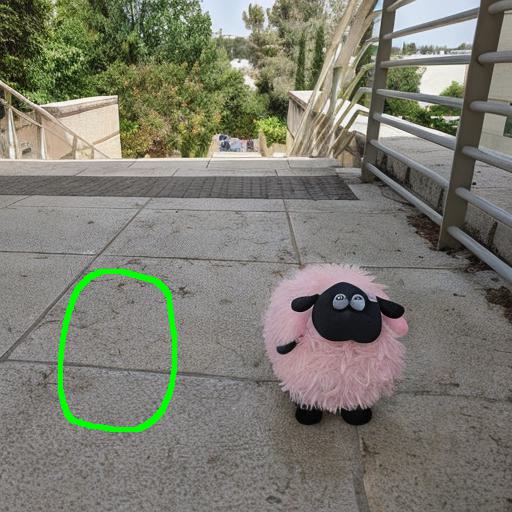} &
        \includegraphics[valign=c, width=\ww,frame]{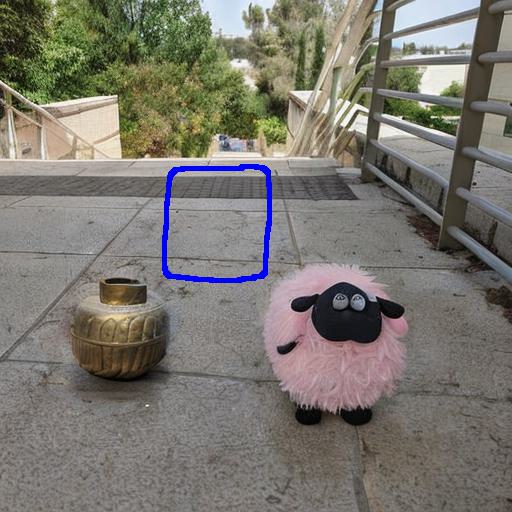} &
        \includegraphics[valign=c, width=\ww,frame]{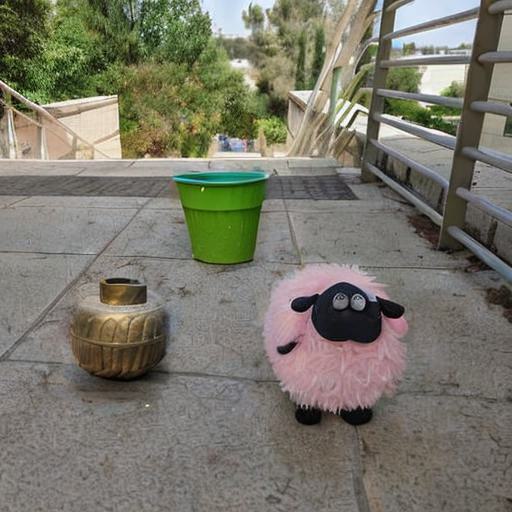}
        \\

        (a) Input scene &
        (b) Input image to edit &
        (c) Edit result 1 &
        (d) Edit result 2 &
        (e) Final result
        \\

        &
        + ``a photo of \tokena'' &
        + ``a photo of \tokenc'' &
        + ``a photo of \tokenb'' &
        \\

        &
        + mask 1 &
        + mask 2 &
        + mask 3 &
        \\

    \end{tabular}
    
    \caption{\textbf{Additional examples of local image editing:} given an input scene (a), we extract the indicated concepts using our method. Given an additional input image to edit (b) along with a mask indicating the edit area, and a guiding text prompt, we use Blended Latent Diffusion \cite{avrahami2022blended,avrahami2022blendedlatent} to obtain the first edit result (c). The process (provide mask and prompt, apply Blended Latent Diffusion) can be repeated (c--d), until the final outcome is obtained (e).}
    \label{fig:bld_additional}
\end{figure*}

%% file: figures/applications/fig_entangled_additional.tex
\begin{figure*}[th]
    \centering
    \setlength{\tabcolsep}{0.5pt}
    \renewcommand{\arraystretch}{0.5}
    \setlength{\ww}{0.4\columnwidth}
    \begin{tabular}{ccccc}
        \includegraphics[valign=c, width=\ww,frame]{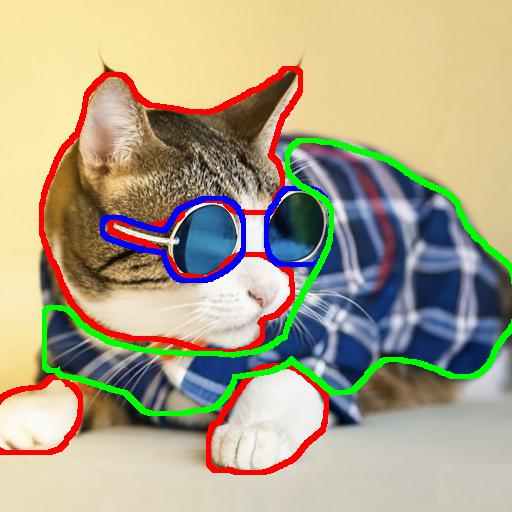}\vspace{2px} &
        \includegraphics[valign=c, width=\ww,frame]{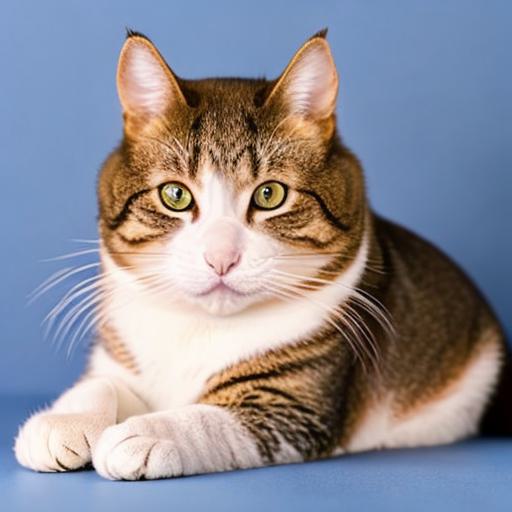} &
        \includegraphics[valign=c, width=\ww,frame]{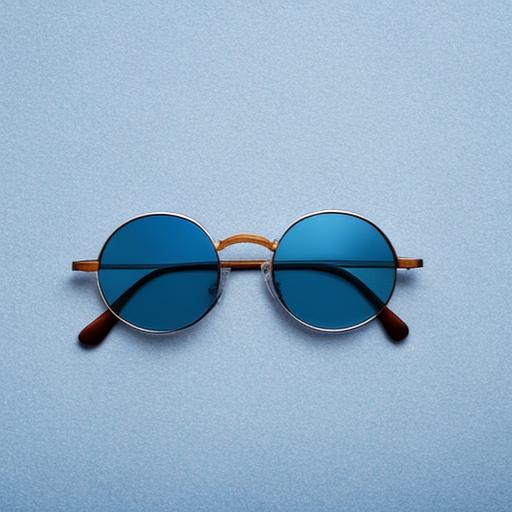} &
        \includegraphics[valign=c, width=\ww,frame]{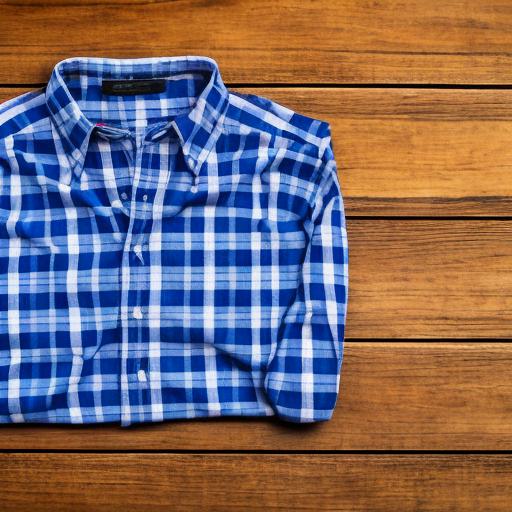} &
        \includegraphics[valign=c, width=\ww,frame]{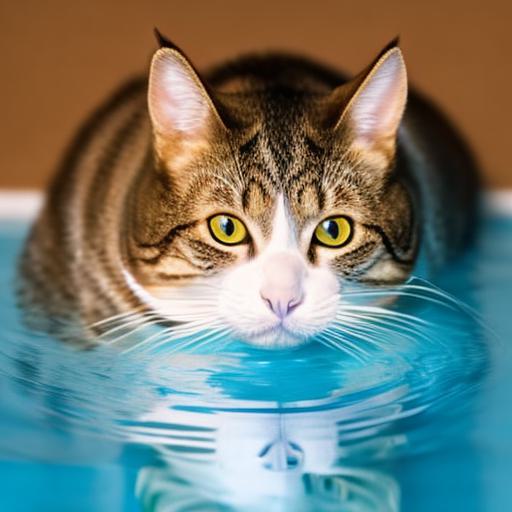}
        \\

        Input scene &
        ``a photo of \tokena on &
        ``a photo of \tokenb on &
        ``a photo of \tokenc on &
        ``a photo of \tokena
        \\

        &
        a solid background'' &
        a solid background'' &
        a table'' &
        swimming''
        \\
        \\

        \includegraphics[valign=c, width=\ww,frame]{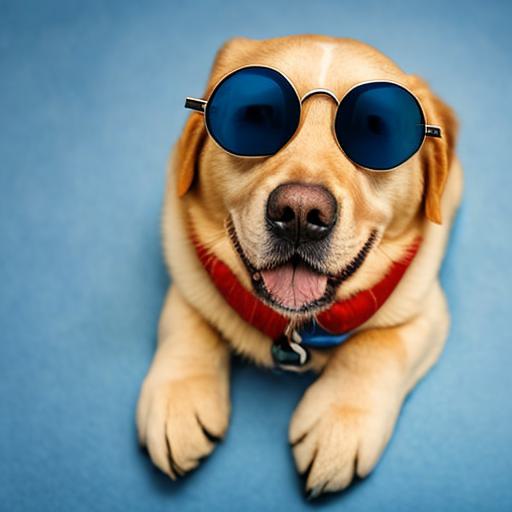}\vspace{2px} &
        \includegraphics[valign=c, width=\ww,frame]{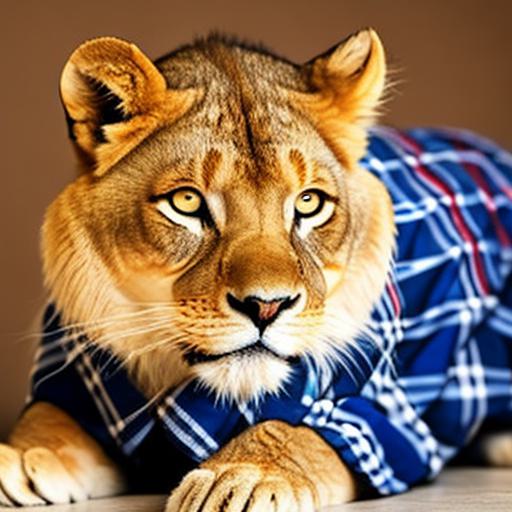} &
        \includegraphics[valign=c, width=\ww,frame]{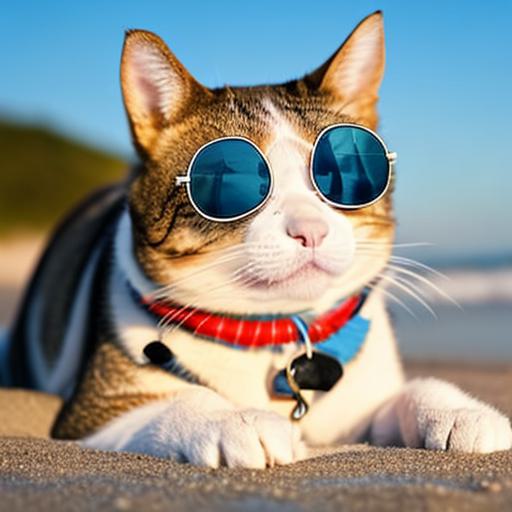} &
        \includegraphics[valign=c, width=\ww,frame]{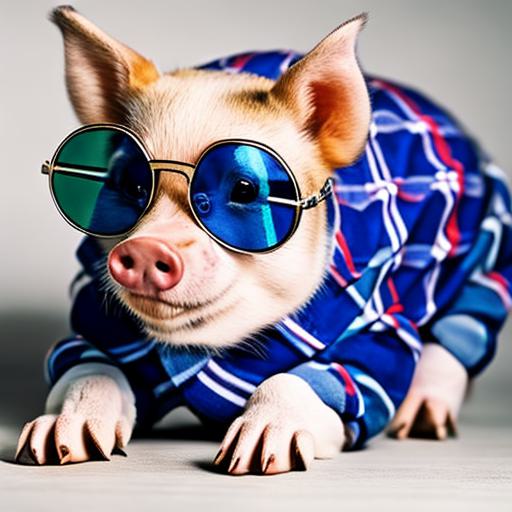} &
        \includegraphics[valign=c, width=\ww,frame]{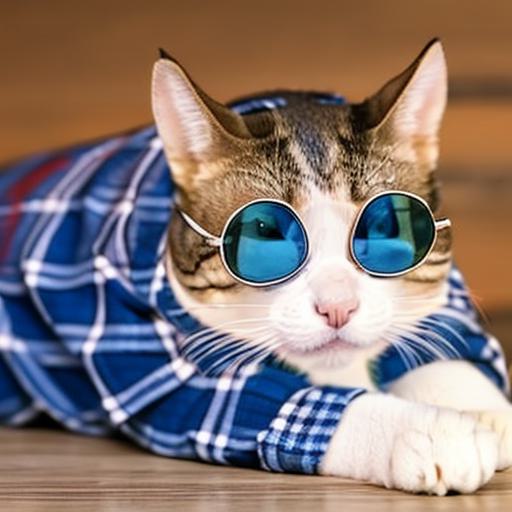}
        \\

        ``a photo of a Labrador &
        ``a photo of a lion &
        ``a photo of \tokena wearing &
        ``a photo of a pig &
        ``a photo of \tokena wearing
        \\

        wearing \tokenb'' &
        wearing \tokenc'' &
        \tokenb at the beach'' &
        wearing \tokenb and \tokenc'' &
        \tokenb and \tokenc on 
        \\

        &&&&
        a wooden floor''
        \\

        \\
        \midrule
        \\
        \\

        \includegraphics[valign=c, width=\ww,frame]{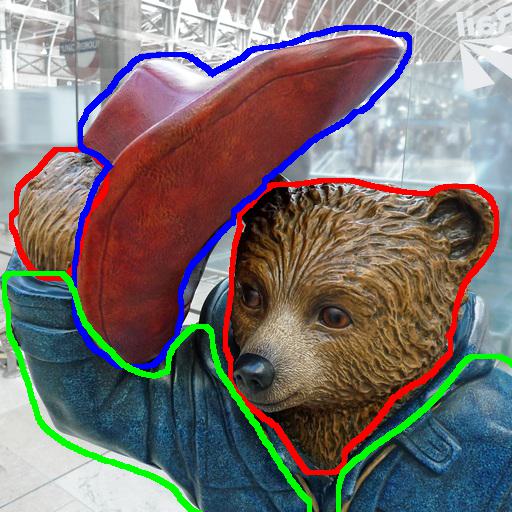}\vspace{2px} &
        \includegraphics[valign=c, width=\ww,frame]{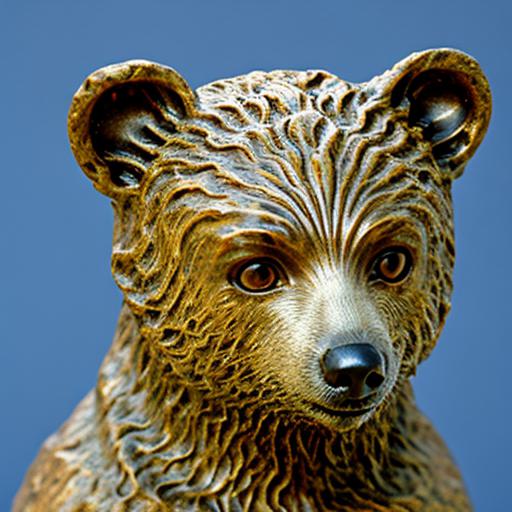} &
        \includegraphics[valign=c, width=\ww,frame]{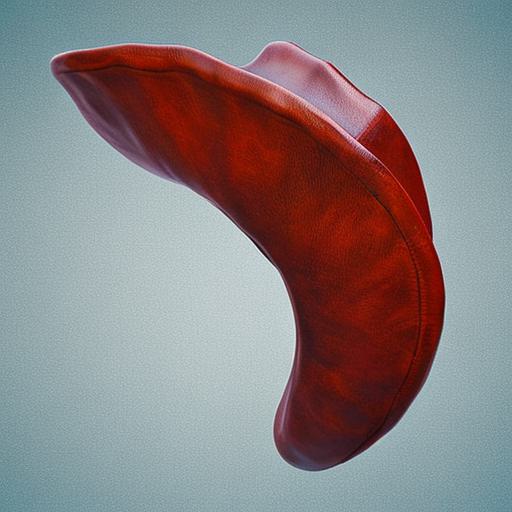} &
        \includegraphics[valign=c, width=\ww,frame]{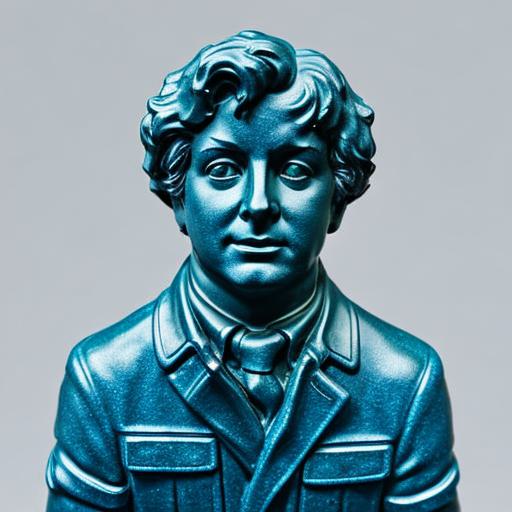} &
        \includegraphics[valign=c, width=\ww,frame]{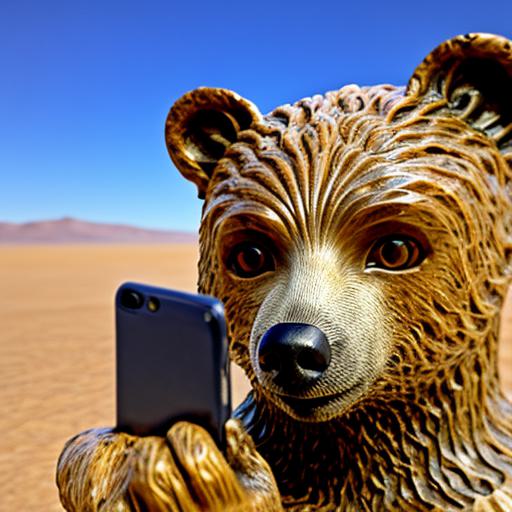}
        \\

        Input scene &
        ``a photo of \tokena on &
        ``a photo of \tokenb on &
        ``a photo of \tokenc on &
        ``a photo of \tokena taking
        \\

        &
        a solid background'' &
        a solid background'' &
        a solid background'' &
        a selfie in the desert''
        \\
        \\

        \includegraphics[valign=c, width=\ww,frame]{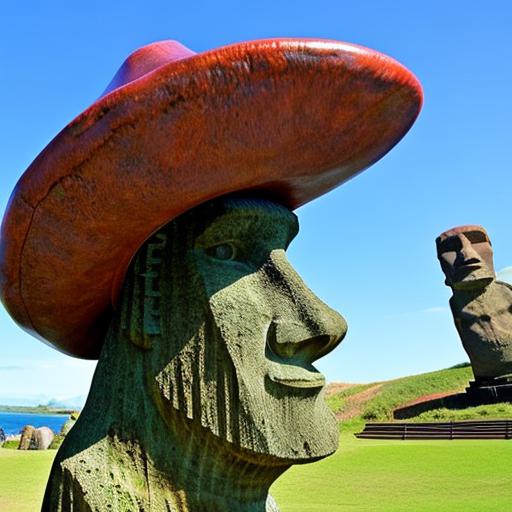}\vspace{2px} &
        \includegraphics[valign=c, width=\ww,frame]{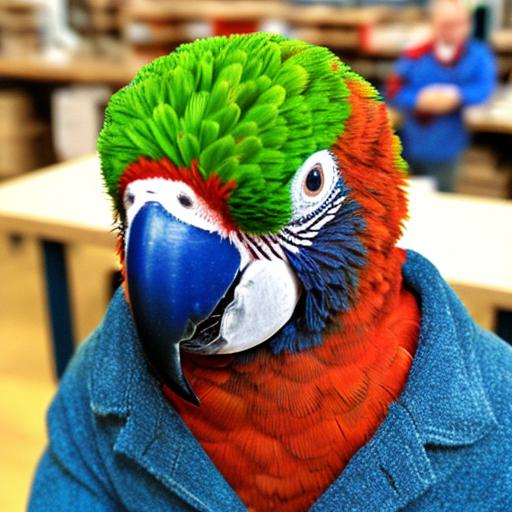} &
        \includegraphics[valign=c, width=\ww,frame]{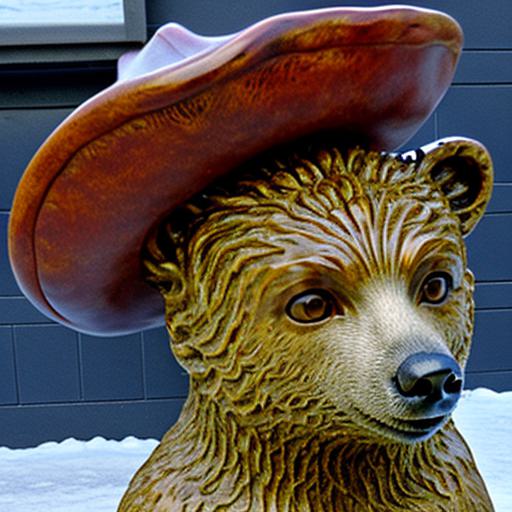} &
        \includegraphics[valign=c, width=\ww,frame]{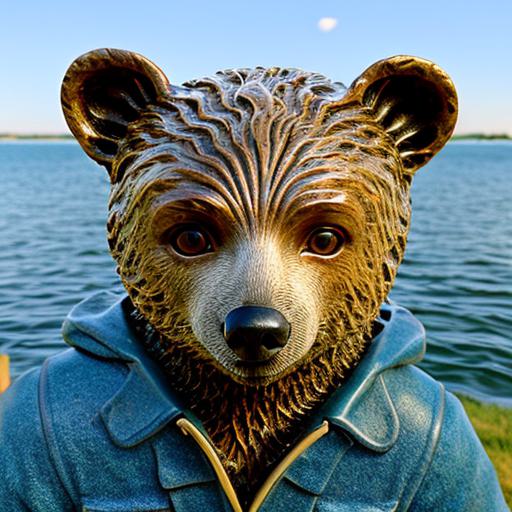} &
        \includegraphics[valign=c, width=\ww,frame]{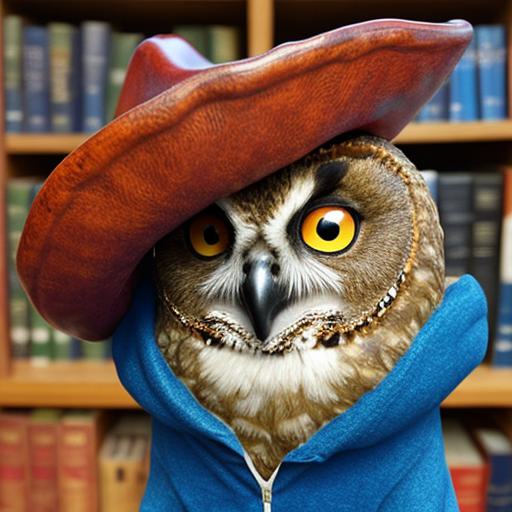}
        \\

        ``a photo of a Moai &
        ``a photo of a parrot &
        ``a photo of \tokena and &
        ``a photo of \tokena &
        ``a photo of an owl
        \\

        statue wearing \tokenb'' &
        wearing \tokenc'' &
        \tokenb in the snow'' &
        wearing \tokenc near a lake'' &
        wearing \tokenb and \tokenc''
        \\

    \end{tabular}
    
    \caption{\textbf{Additional examples of entangled scene decomposition:} given a single input image of several spatially-entangled concepts, our method is able to disentangle them and generate novel images with them, separately or jointly.}
    \label{fig:entangeld_scene_additional}
\end{figure*}

%% file: figures/applications/fig_variations_additional.tex
\begin{figure*}[th]
    \centering
    \setlength{\tabcolsep}{0.5pt}
    \renewcommand{\arraystretch}{0.5}
    \setlength{\ww}{0.4\columnwidth}
    \begin{tabular}{ccccc}
        \includegraphics[valign=c, width=\ww,frame]{figures/applications/entangled_scene/cat_glasses_shirt/mask_overlay.jpg}\vspace{2px} &
        \includegraphics[valign=c, width=\ww,frame]{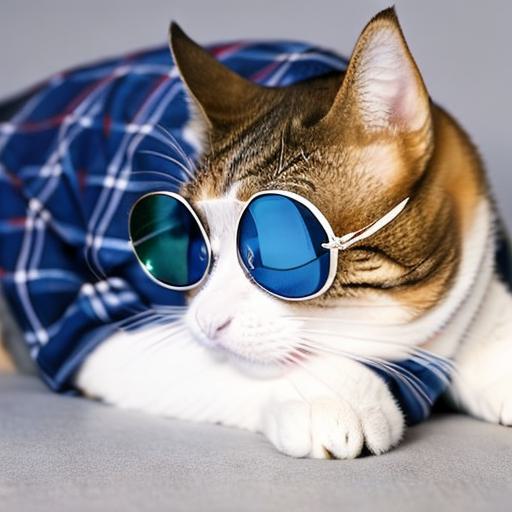} &
        \includegraphics[valign=c, width=\ww,frame]{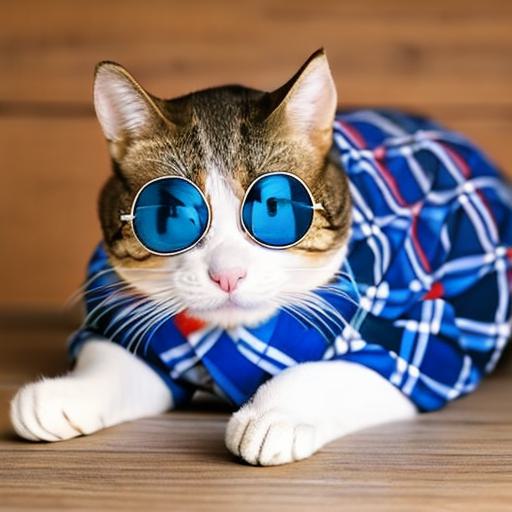} &
        \includegraphics[valign=c, width=\ww,frame]{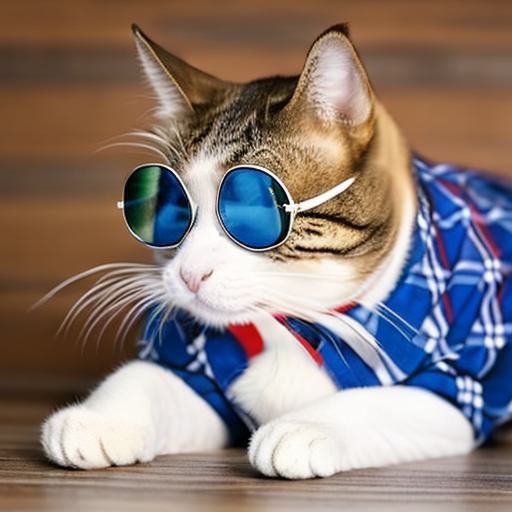} &
        \includegraphics[valign=c, width=\ww,frame]{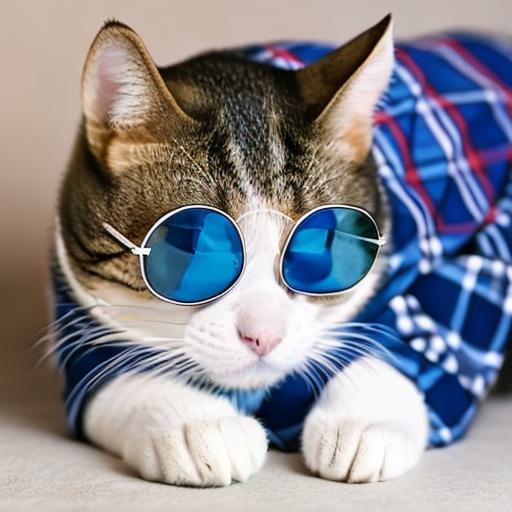}
        \\

        Input scene &
        ``a photo of \tokena and &
        ``a photo of \tokena and &
        ``a photo of \tokena and &
        ``a photo of \tokena and
        \\

        &
        \tokenb and \tokenc'' &
        \tokenb and \tokenc'' &
        \tokenb and \tokenc'' &
        \tokenb and \tokenc''
        \\
        \\

        \includegraphics[valign=c, width=\ww,frame]{figures/applications/entangled_scene/bear_statue/mask_overlay.jpg}\vspace{2px} &
        \includegraphics[valign=c, width=\ww,frame]{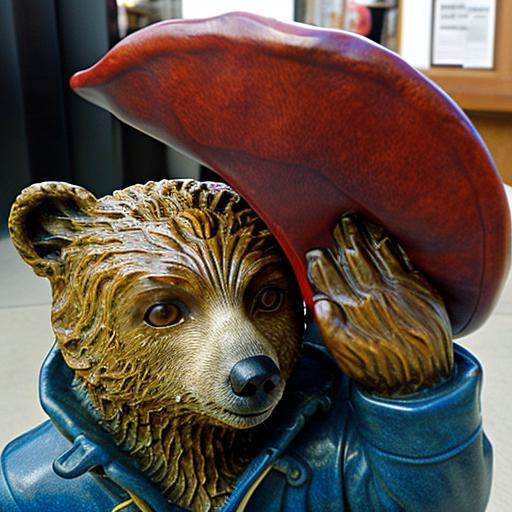} &
        \includegraphics[valign=c, width=\ww,frame]{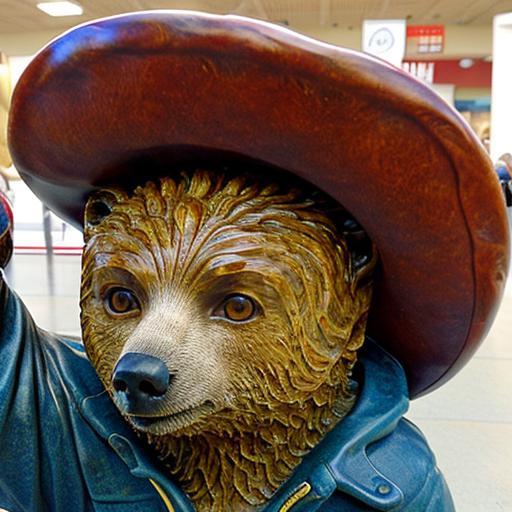} &
        \includegraphics[valign=c, width=\ww,frame]{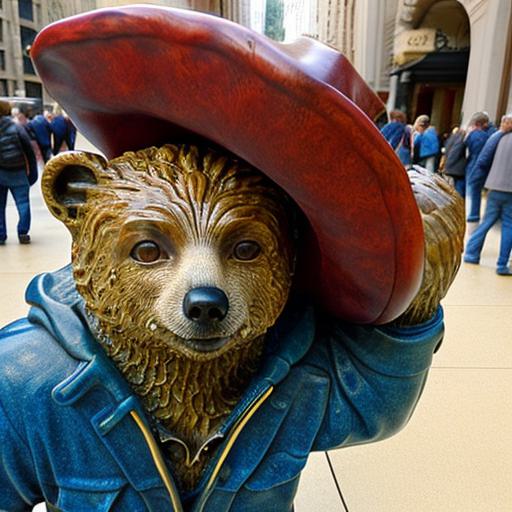} &
        \includegraphics[valign=c, width=\ww,frame]{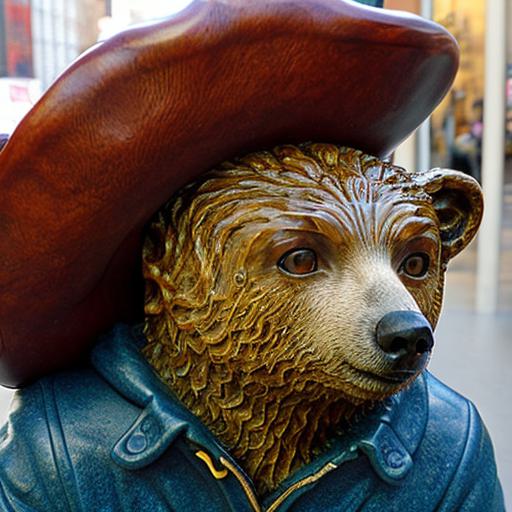}
        \\

        Input scene &
        ``a photo of \tokena and &
        ``a photo of \tokena and &
        ``a photo of \tokena and &
        ``a photo of \tokena and
        \\

        &
        \tokenb and \tokenc'' &
        \tokenb and \tokenc'' &
        \tokenb and \tokenc'' &
        \tokenb and \tokenc''
        \\
        \\

        \includegraphics[valign=c, width=\ww,frame]{figures/additional_qualitative_comparison/assets/bear/mask_overlay.jpg}\vspace{2px} &
        \includegraphics[valign=c, width=\ww,frame]{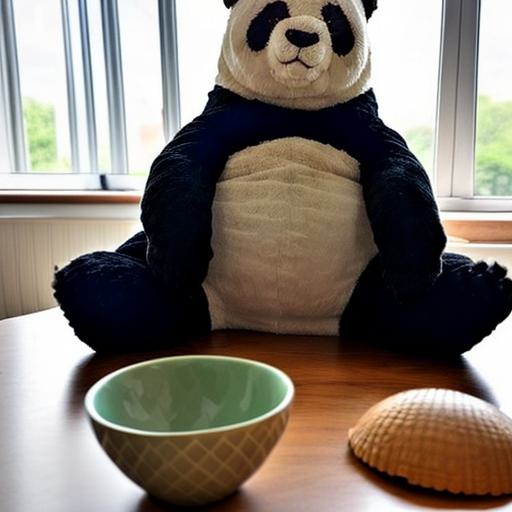} &
        \includegraphics[valign=c, width=\ww,frame]{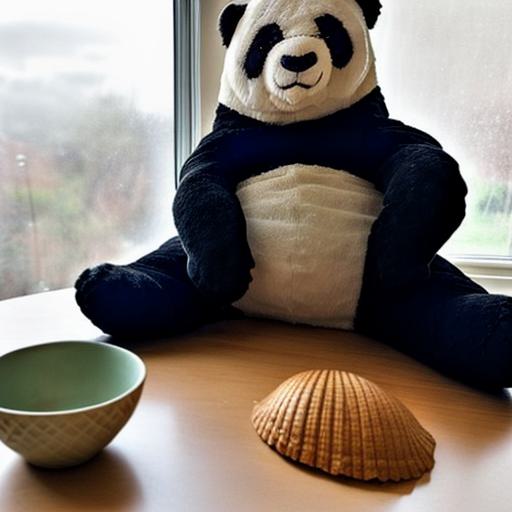} &
        \includegraphics[valign=c, width=\ww,frame]{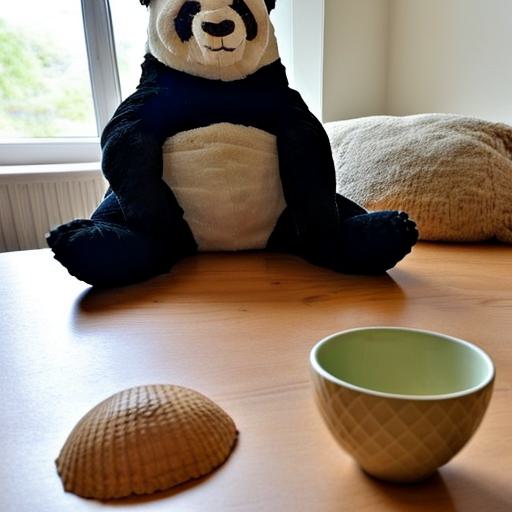} &
        \includegraphics[valign=c, width=\ww,frame]{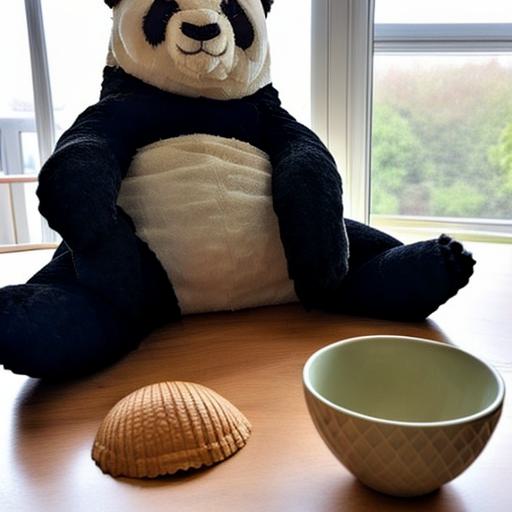}
        \\

        Input scene &
        ``a photo of \tokena and &
        ``a photo of \tokena and &
        ``a photo of \tokena and &
        ``a photo of \tokena and
        \\

        &
        \tokenb and \tokenc'' &
        \tokenb and \tokenc'' &
        \tokenb and \tokenc'' &
        \tokenb and \tokenc''
        \\
        \\

    \end{tabular}
    
    \caption{\textbf{Additional examples of image variations applications:} given a single input image of several concepts, our method is able to generate many variations of the image. Credits: pxhere}
    \label{fig:variations_additional}
\end{figure*}

%% file: figures/naive_adaptation/fig.tex
\begin{figure}[t]
    \centering
    \setlength{\tabcolsep}{0.5pt}
    \renewcommand{\arraystretch}{0.5}
    \setlength{\ww}{0.3\columnwidth}
    \begin{tabular}{cccc}
        \rotatebox[origin=c]{90}{(a) Inputs} &
        \includegraphics[valign=c, width=\ww,frame]{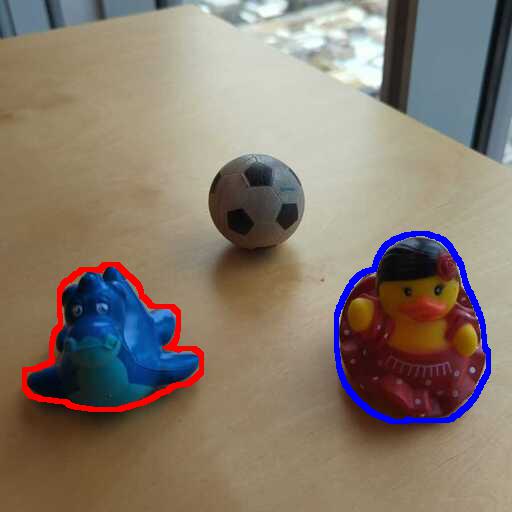} 
        \\
        \\

        \rotatebox[origin=c]{90}{(b) TI} &
        \includegraphics[valign=c, width=\ww,frame]{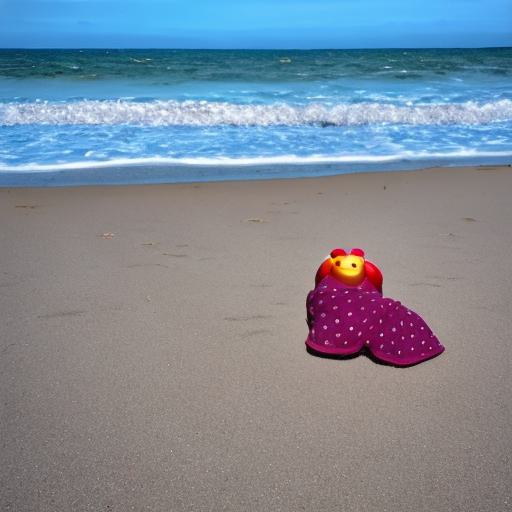} &
        \includegraphics[valign=c, width=\ww,frame]{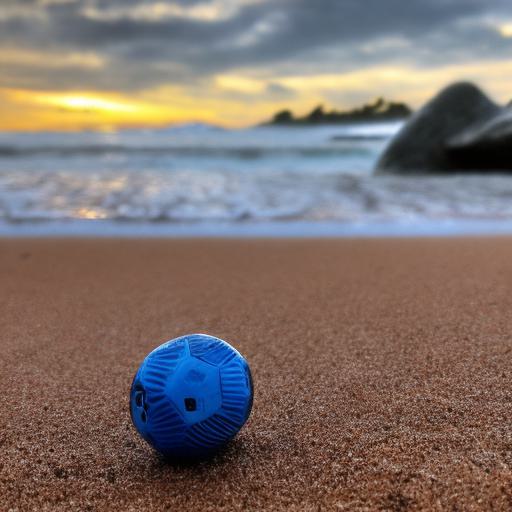} &
        \includegraphics[valign=c, width=\ww,frame]{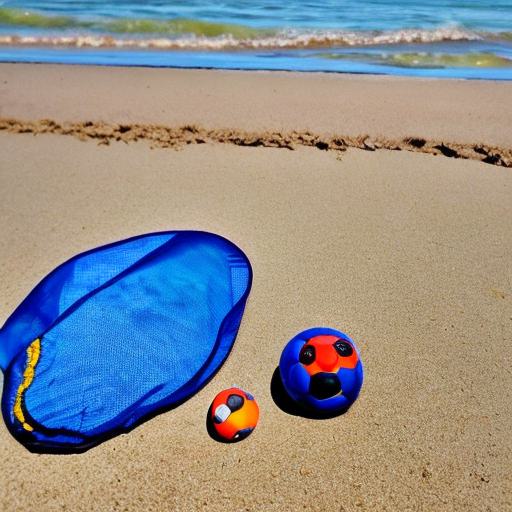}
        \\
        \\

        \rotatebox[origin=c]{90}{(c) DB} &
        \includegraphics[valign=c, width=\ww,frame]{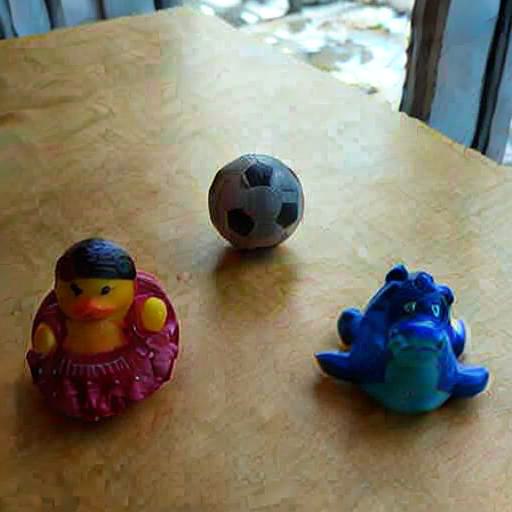} &
        \includegraphics[valign=c, width=\ww,frame]{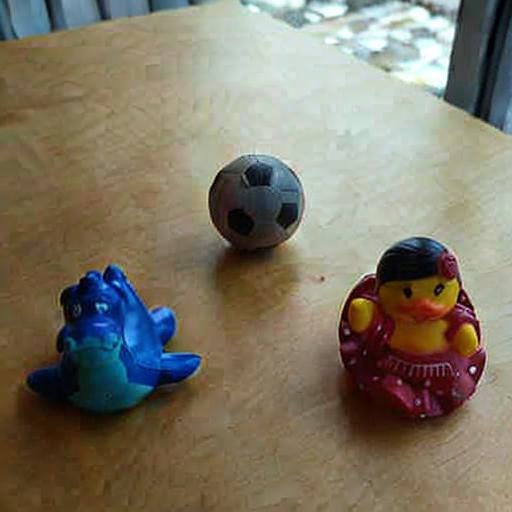} &
        \includegraphics[valign=c, width=\ww,frame]{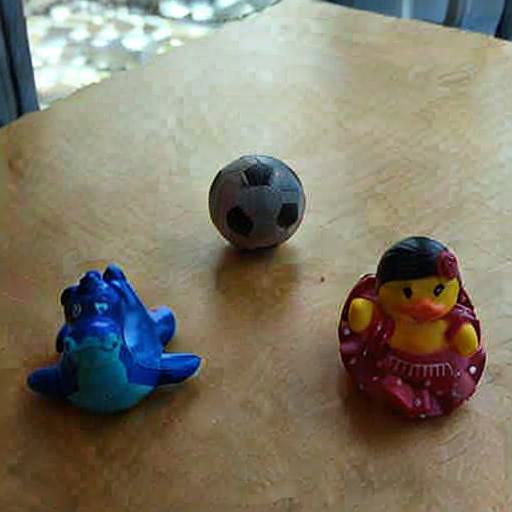}
        \\
        \\

        \rotatebox[origin=c]{90}{(d) Ours} &
        \includegraphics[valign=c, width=\ww,frame]{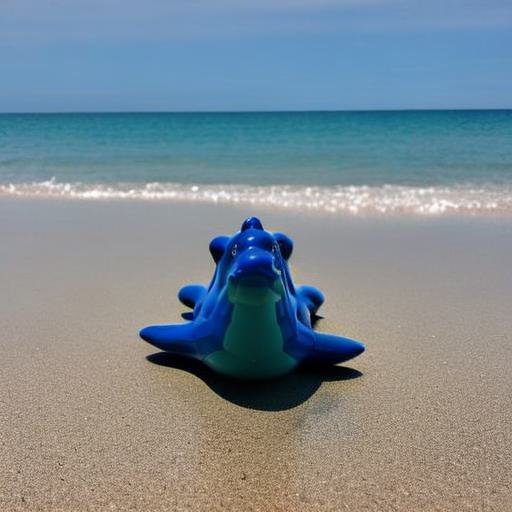}\vspace{2px} &
        \includegraphics[valign=c, width=\ww,frame]{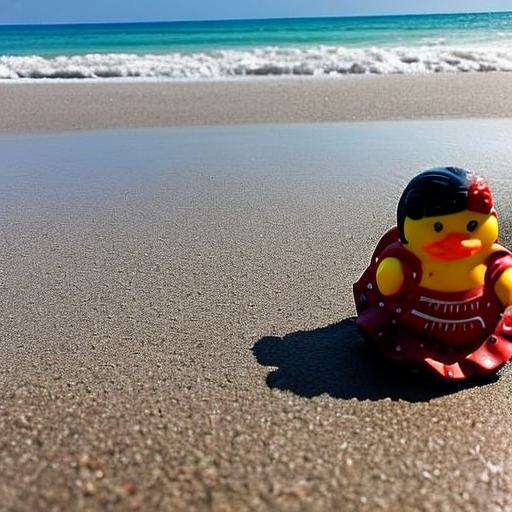} &
        \includegraphics[valign=c, width=\ww,frame]{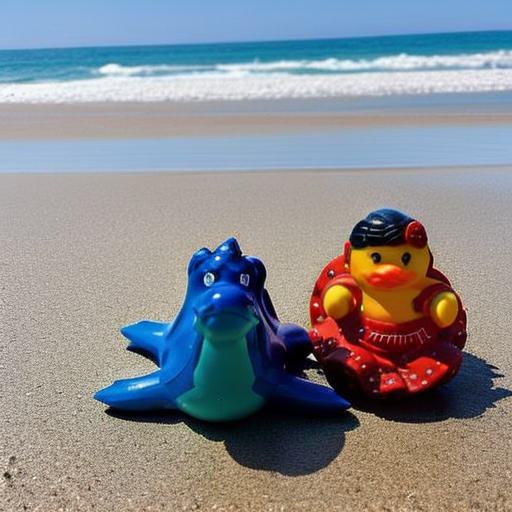}
        \\

        &
        ``a photo of \tokena &
        ``a photo of \tokenb &
        ``a photo of \tokena and
        \\

        &
        at the beach'' &
        at the beach'' &
        \tokenb at the beach''
        \\

    \end{tabular}
    
    \caption{\textbf{Naive TI and DB adaptations:} given an (a) input scene, trying \naive{}ly running (b) TI and (c) DB on the input image. As expected, these approach fails to disentangle between the concepts --- TI learns an arbitrary concept while DB overfits the input image. On the other hand, (d) our method is able to learn the identity of the concepts while taking into account the text prompt.}
    \label{fig:naive_adaptation}
\end{figure}

%% file: figures/additional_qualitative_comparison/fig.tex
\begin{figure*}[t]
    \centering
    \setlength{\tabcolsep}{0.5pt}
    \renewcommand{\arraystretch}{0.5}
    \setlength{\ww}{0.285\columnwidth}
    \begin{tabular}{c @{\hspace{10\tabcolsep}} @{\hspace{10\tabcolsep}} ccccc}

        \textbf{Inputs} &
        \textbf{\maskedTI} &
        \textbf{\maskedDB} &
        \textbf{\maskedCD} &
        \textbf{ELITE} &
        \textbf{Ours}
        \\

        &
        \cite{Gal2022AnII} &
        \cite{Ruiz2022DreamBoothFT} &
        \cite{Kumari2022MultiConceptCO} &
        \cite{Wei2023ELITEEV}
        \\

        \\
        \raisebox{-1.3\height}[0pt][0pt]{\includegraphics[valign=c, width=\ww,frame]{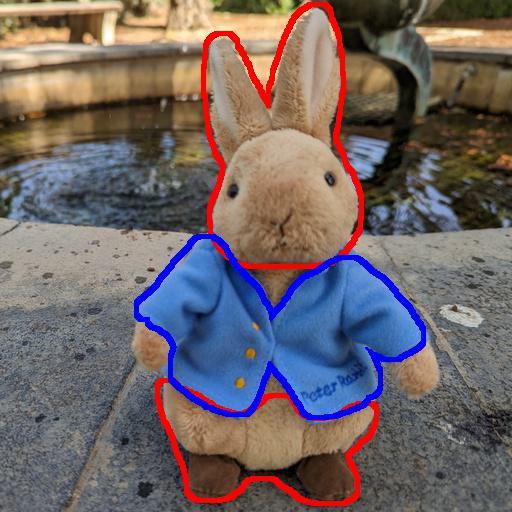}} &
        \includegraphics[valign=c, width=\ww,frame]{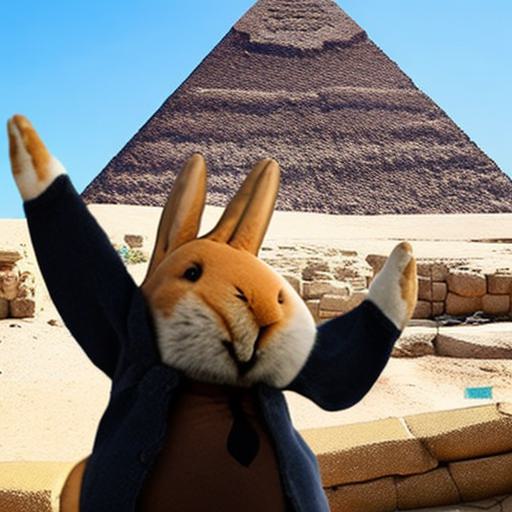} &
        \includegraphics[valign=c, width=\ww,frame]{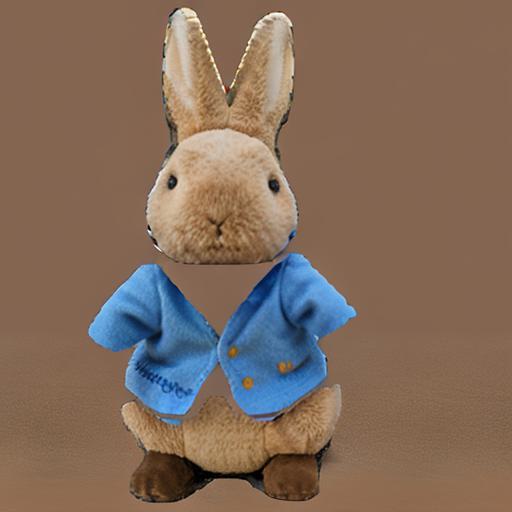} &
        \includegraphics[valign=c, width=\ww,frame]{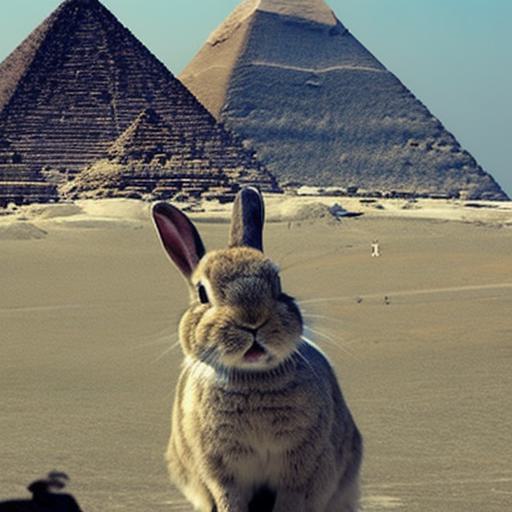} &
        \includegraphics[valign=c, width=\ww,frame]{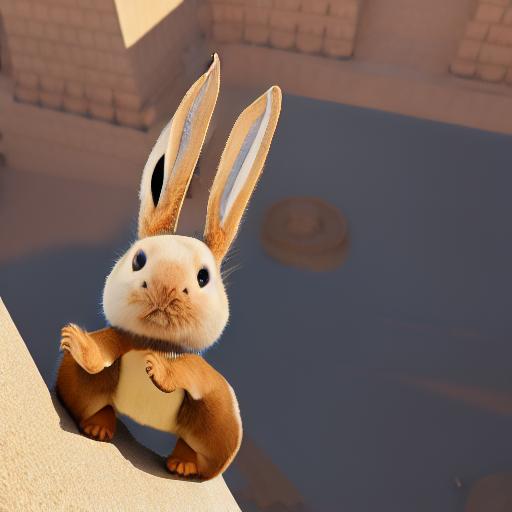} &
        \includegraphics[valign=c, width=\ww,frame]{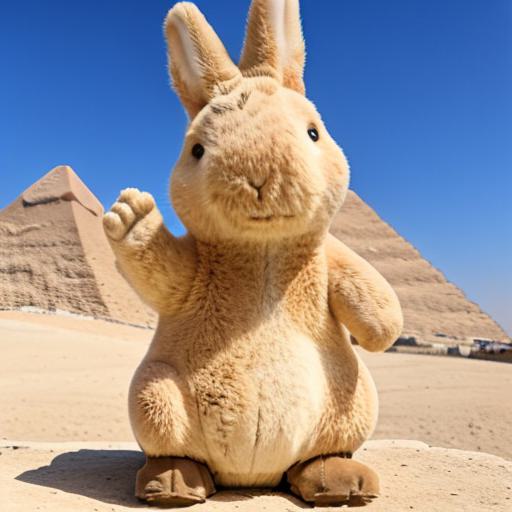}
        \\
        \\

        &
        \multicolumn{5}{c}{``a photo of \tokena raisng its hand near the Great Pyramid of Giza''}
        \\
        \\

        &
        \includegraphics[valign=c, width=\ww,frame]{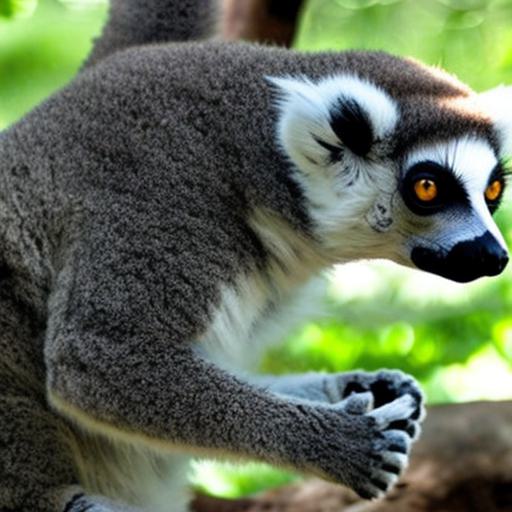} &
        \includegraphics[valign=c, width=\ww,frame]{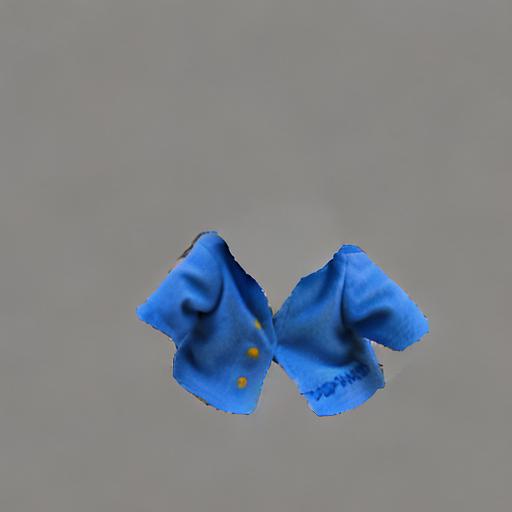} &
        \includegraphics[valign=c, width=\ww,frame]{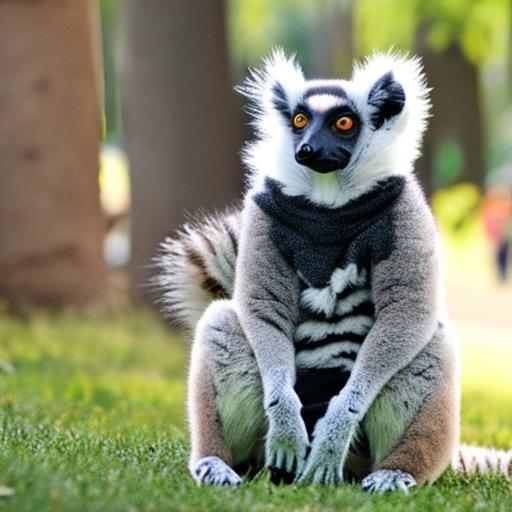} &
        \includegraphics[valign=c, width=\ww,frame]{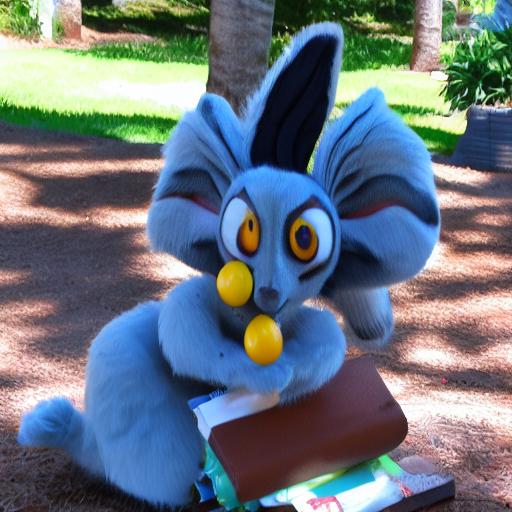} &
        \includegraphics[valign=c, width=\ww,frame]{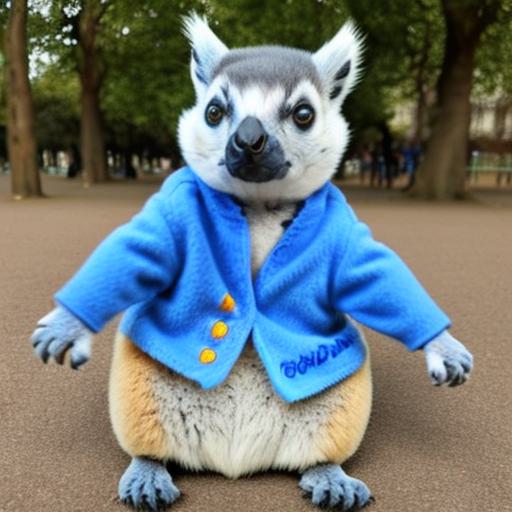}
        \\
        \\

        &
        \multicolumn{5}{c}{``a photo of a lemur earing \tokenb in the park''}
        \\
        \\

        \midrule  

        \\
        \raisebox{-1.3\height}[0pt][0pt]{\includegraphics[valign=c, width=\ww,frame]{figures/additional_qualitative_comparison/assets/bear/mask_overlay.jpg}} &
        \includegraphics[valign=c, width=\ww,frame]{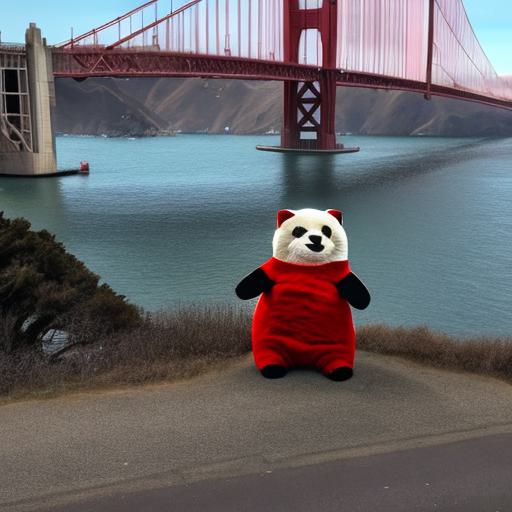} &
        \includegraphics[valign=c, width=\ww,frame]{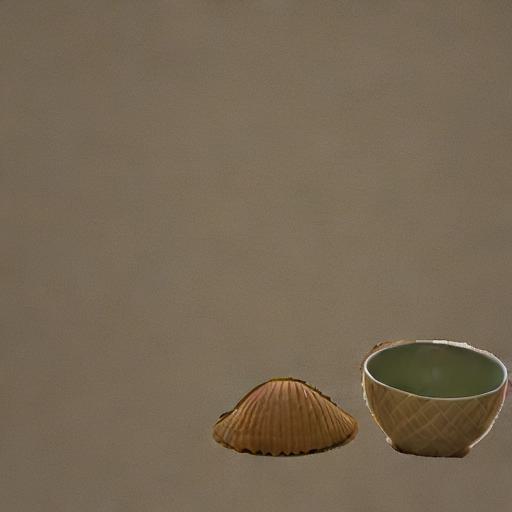} &
        \includegraphics[valign=c, width=\ww,frame]{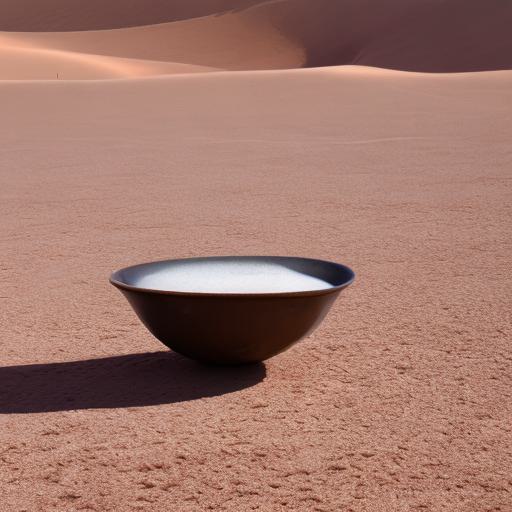} &
        \includegraphics[valign=c, width=\ww,frame]{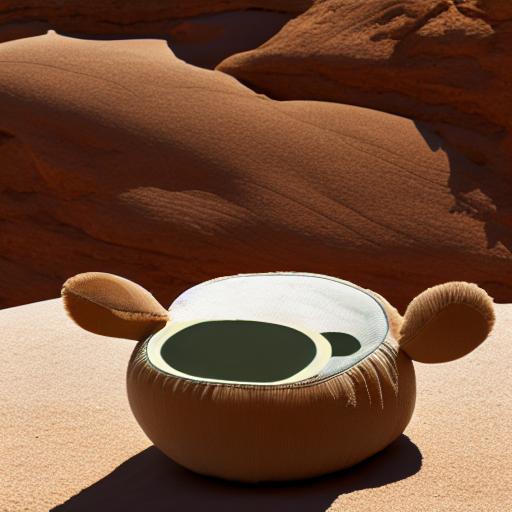} &
        \includegraphics[valign=c, width=\ww,frame]{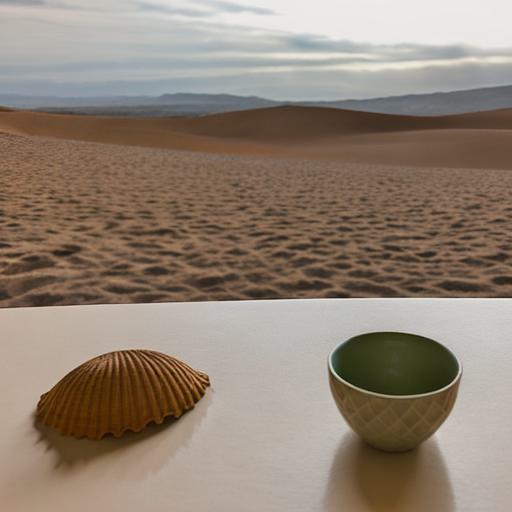}
        \\
        \\

        &
        \multicolumn{5}{c}{``a photo of \tokena and \tokenb in the desert''}
        \\
        \\

        &
        \includegraphics[valign=c, width=\ww,frame]{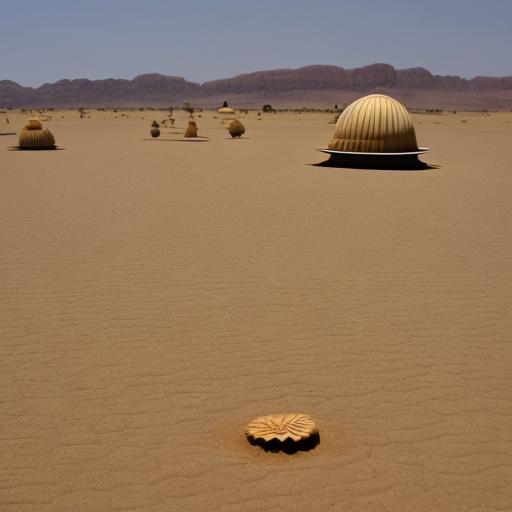} &
        \includegraphics[valign=c, width=\ww,frame]{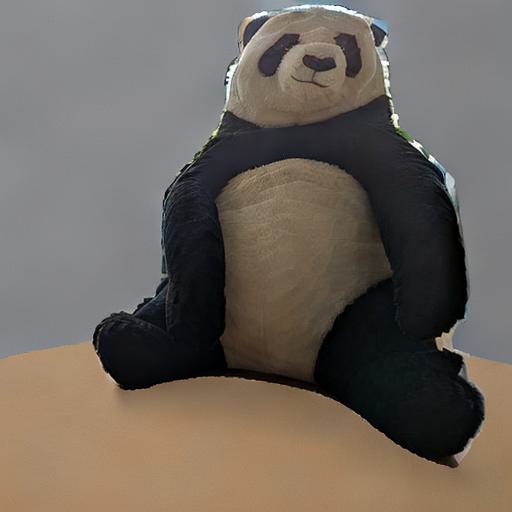} &
        \includegraphics[valign=c, width=\ww,frame]{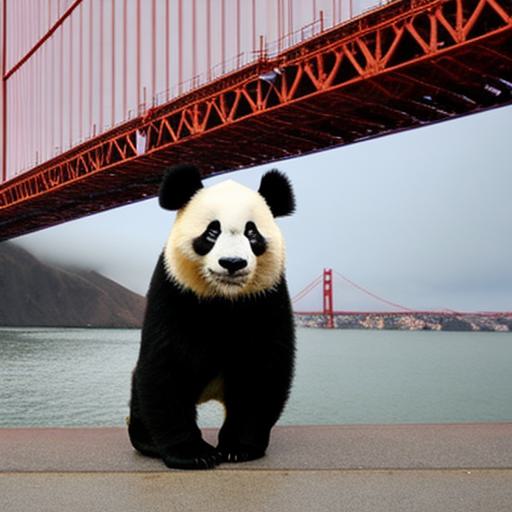} &
        \includegraphics[valign=c, width=\ww,frame]{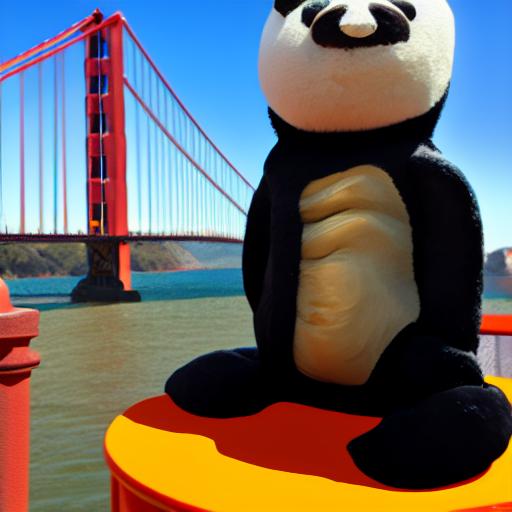} &
        \includegraphics[valign=c, width=\ww,frame]{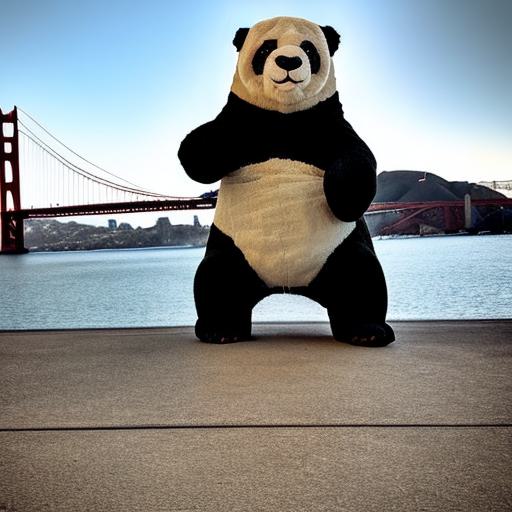}
        \\
        \\

        &
        \multicolumn{5}{c}{``a photo of \tokenc standing near the Golden Gate Bridge''}
        \\
        \\
        \midrule

        \\
        \raisebox{-1.3\height}[0pt][0pt]{\includegraphics[valign=c, width=\ww,frame]{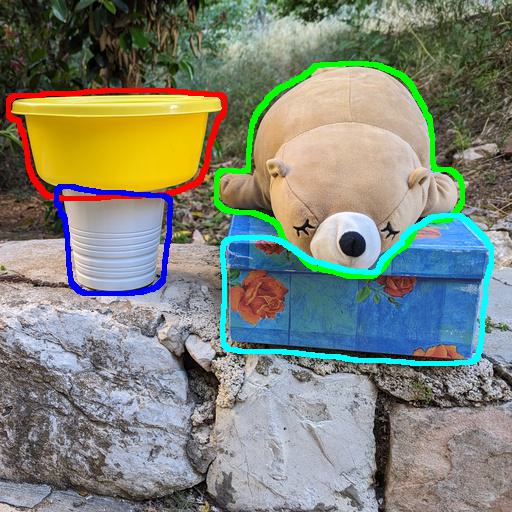}} &
        \includegraphics[valign=c, width=\ww,frame]{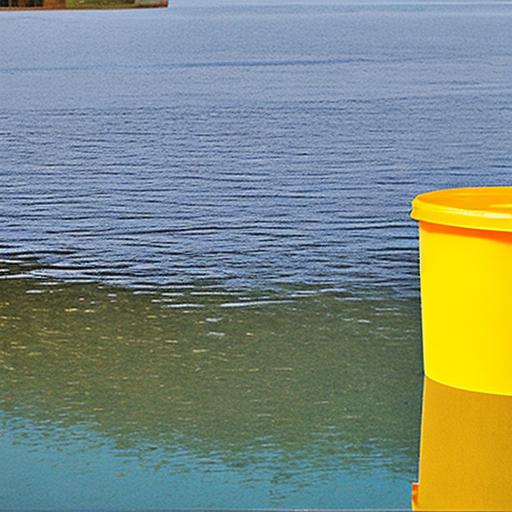} &
        \includegraphics[valign=c, width=\ww,frame]{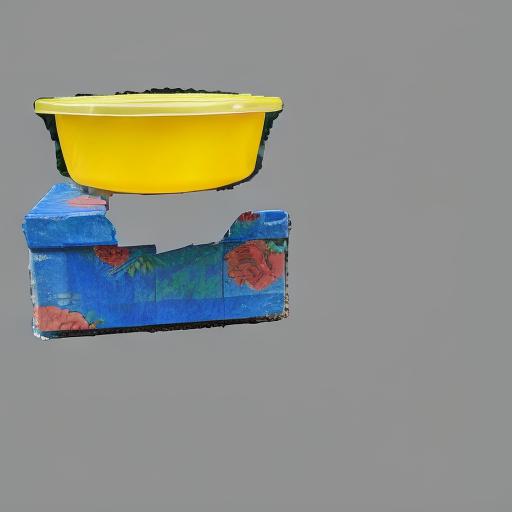} &
        \includegraphics[valign=c, width=\ww,frame]{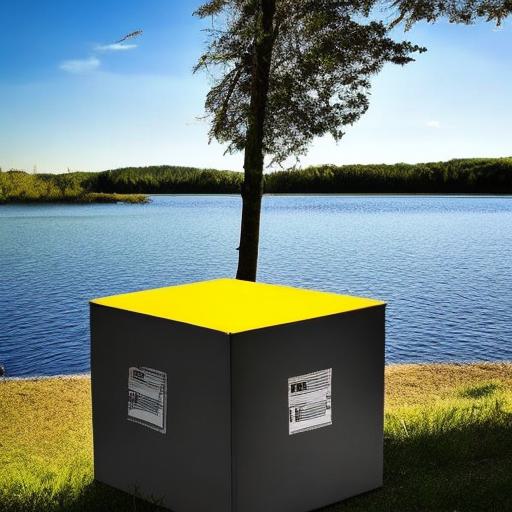} &
        \includegraphics[valign=c, width=\ww,frame]{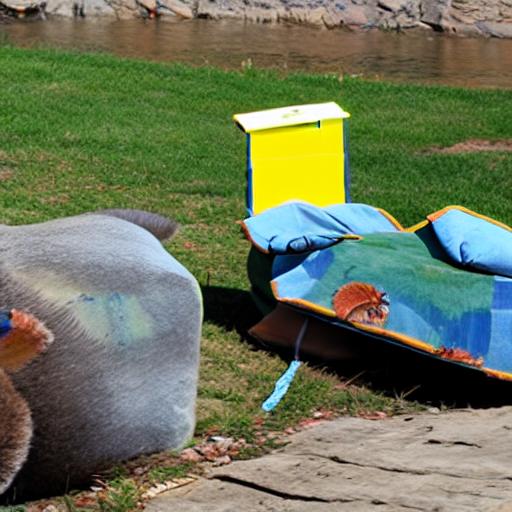} &
        \includegraphics[valign=c, width=\ww,frame]{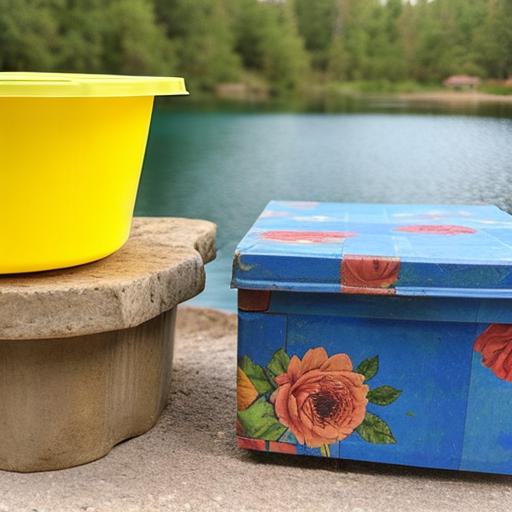}
        \\
        \\

        &
        \multicolumn{5}{c}{``a photo of \tokena and \tokenb in the desert''}
        \\
        \\

        &
        \includegraphics[valign=c, width=\ww,frame]{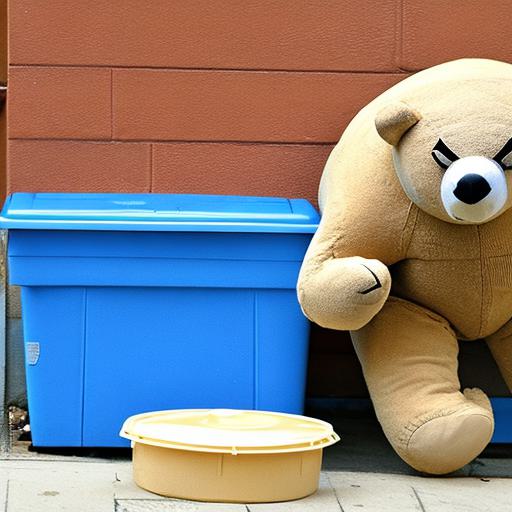} &
        \includegraphics[valign=c, width=\ww,frame]{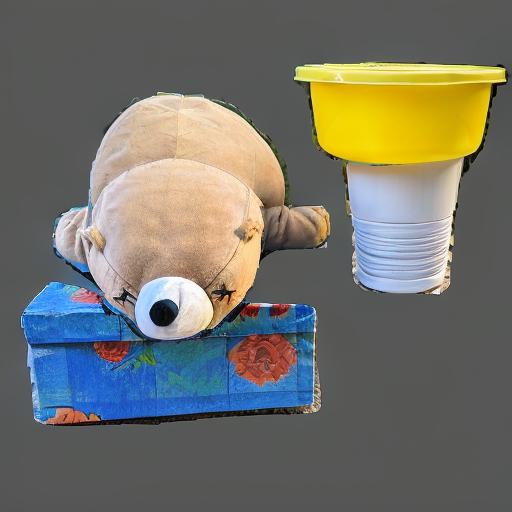} &
        \includegraphics[valign=c, width=\ww,frame]{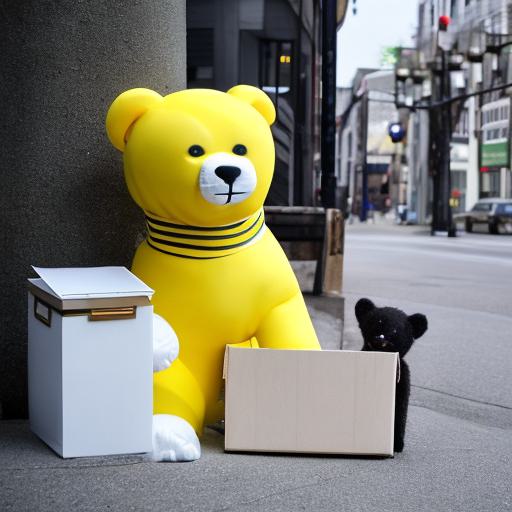} &
        \includegraphics[valign=c, width=\ww,frame]{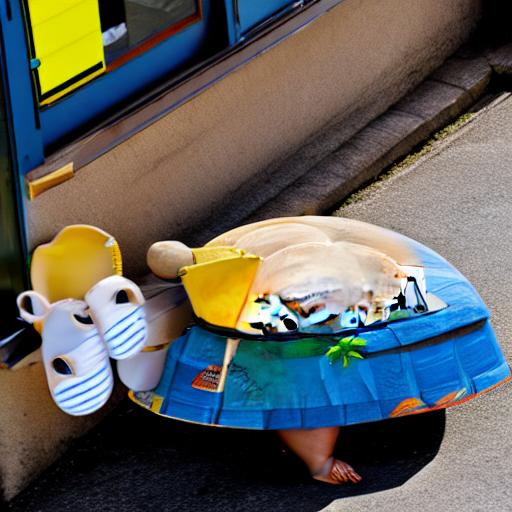} &
        \includegraphics[valign=c, width=\ww,frame]{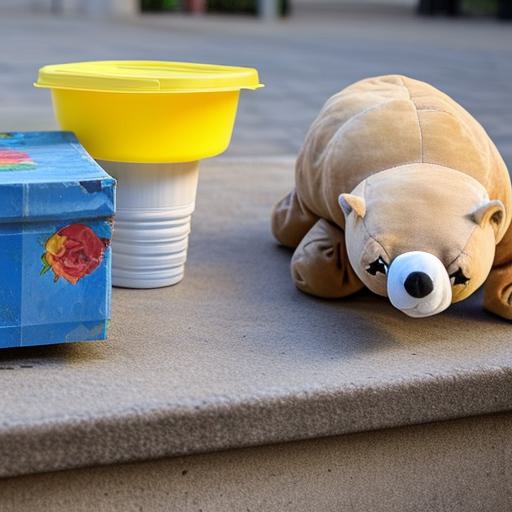}
        \\
        \\

        &
        \multicolumn{5}{c}{``a photo of \tokena and \tokenb and \tokenc and \tokend in the street''}
        \\
        \\

    \end{tabular}
    
    \caption{\textbf{A qualitative comparison} between several baselines and our method. As can be seen, \maskedTI and \maskedCD struggle with preserving the concept identities, while the images generated by \maskedDB effectively ignore the text prompt. ELITE preserves the identities better than \maskedTI/\maskedCD, but the concepts are still not recognizable enough, especially when more than one concept is generated. Finally, our method is able to preserve the identities as well as to follow the text prompt, even when learning four different concepts (bottom row).}
    \label{fig:additional_qualitative_comparison}
\end{figure*}

%% file: figures/automatic_dataset_comparison/fig.tex
\begin{figure*}[t]
    \centering
    \setlength{\tabcolsep}{0.5pt}
    \renewcommand{\arraystretch}{0.5}
    \setlength{\ww}{0.285\columnwidth}
    \begin{tabular}{c @{\hspace{10\tabcolsep}} @{\hspace{10\tabcolsep}} ccccc}

        \textbf{Inputs} &
        \textbf{\maskedTI} &
        \textbf{\maskedDB} &
        \textbf{\maskedCD} &
        \textbf{ELITE} &
        \textbf{Ours}
        \\

        &
        \cite{Gal2022AnII} &
        \cite{Ruiz2022DreamBoothFT} &
        \cite{Kumari2022MultiConceptCO} &
        \cite{Wei2023ELITEEV}
        \\

        \\
        \raisebox{-1.3\height}[0pt][0pt]{\includegraphics[valign=c, width=\ww,frame]{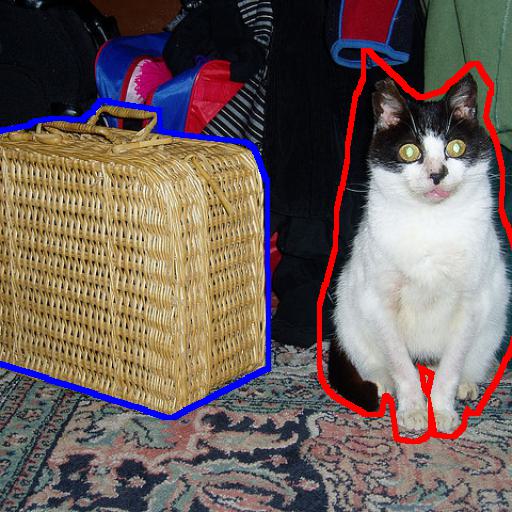}} &
        \includegraphics[valign=c, width=\ww,frame]{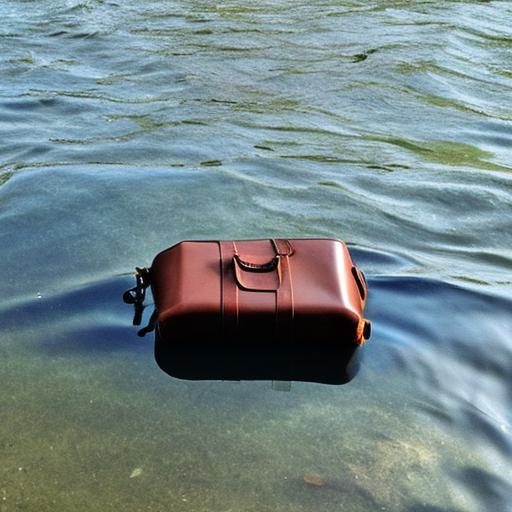} &
        \includegraphics[valign=c, width=\ww,frame]{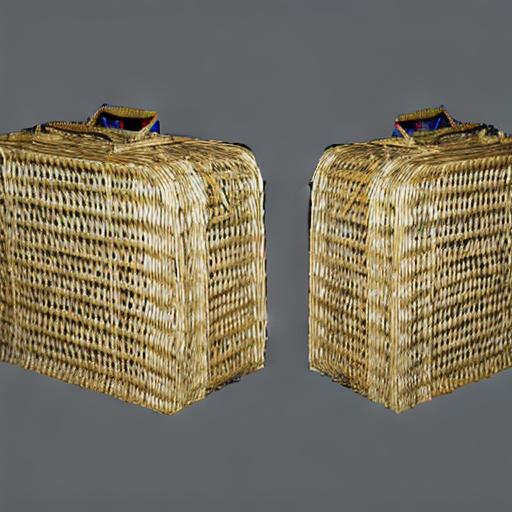} &
        \includegraphics[valign=c, width=\ww,frame]{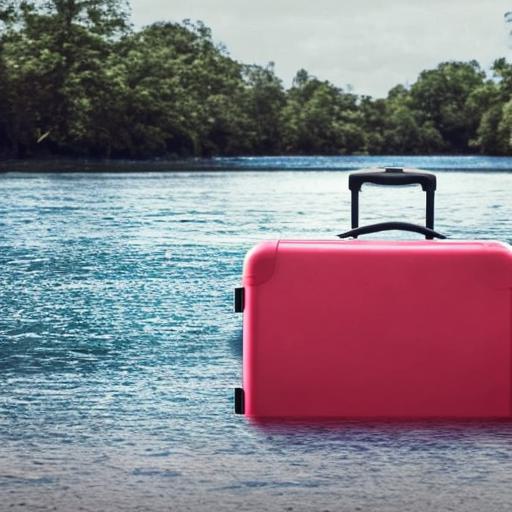} &
        \includegraphics[valign=c, width=\ww,frame]{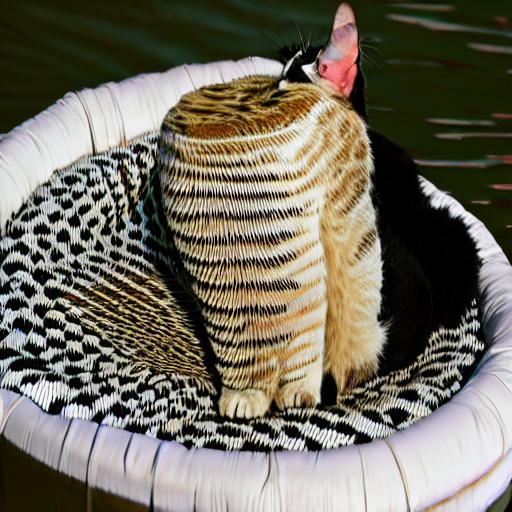} &
        \includegraphics[valign=c, width=\ww,frame]{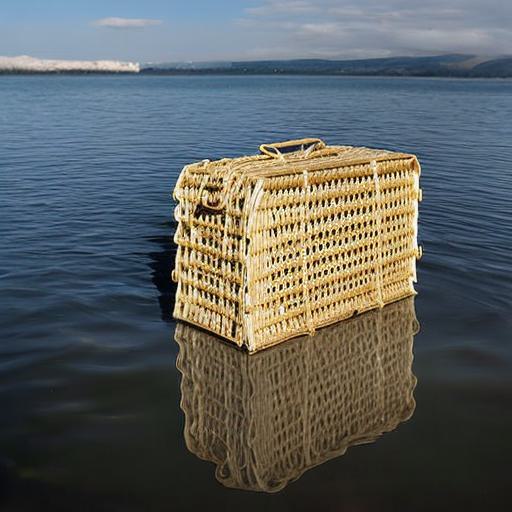}
        \\
        \\

        &
        \multicolumn{5}{c}{``a photo of \tokenb floating on top of water''}
        \\
        \\

        &
        \includegraphics[valign=c, width=\ww,frame]{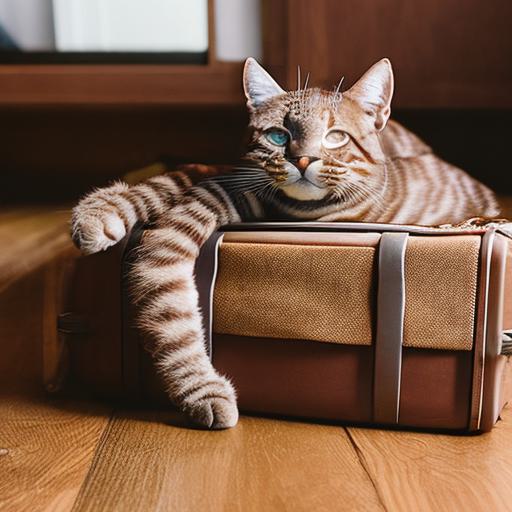} &
        \includegraphics[valign=c, width=\ww,frame]{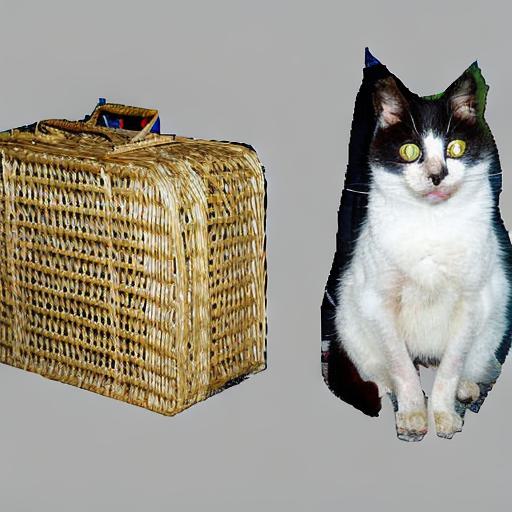} &
        \includegraphics[valign=c, width=\ww,frame]{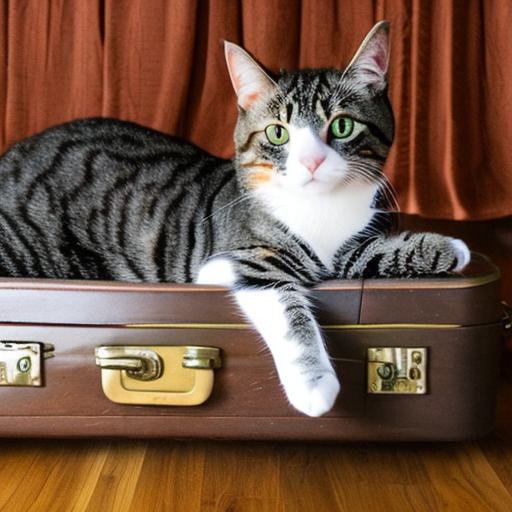} &
        \includegraphics[valign=c, width=\ww,frame]{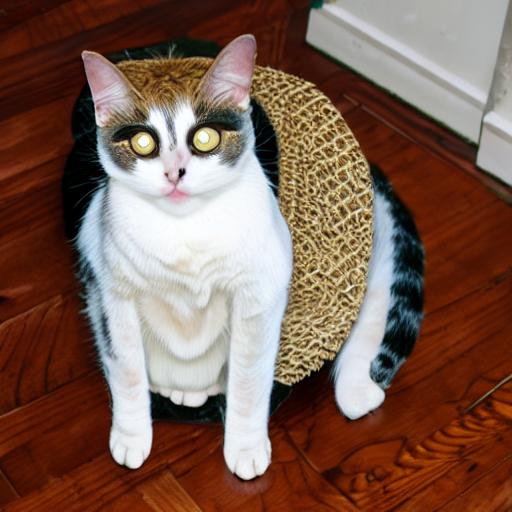} &
        \includegraphics[valign=c, width=\ww,frame]{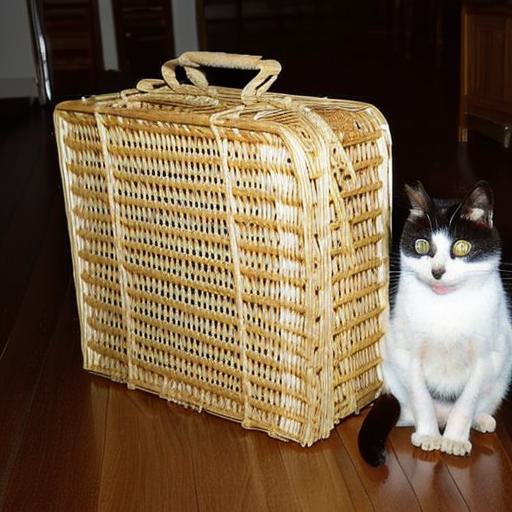}
        \\
        \\

        &
        \multicolumn{5}{c}{``a photo of \tokena and \tokenb on top of a wooden floor''}
        \\
        \\
        \midrule

        \\
        \raisebox{-1.3\height}[0pt][0pt]{\includegraphics[valign=c, width=\ww,frame]{figures/automatic_dataset_comparison/assets/bear_glass/mask_overlay.jpg}} &
        \includegraphics[valign=c, width=\ww,frame]{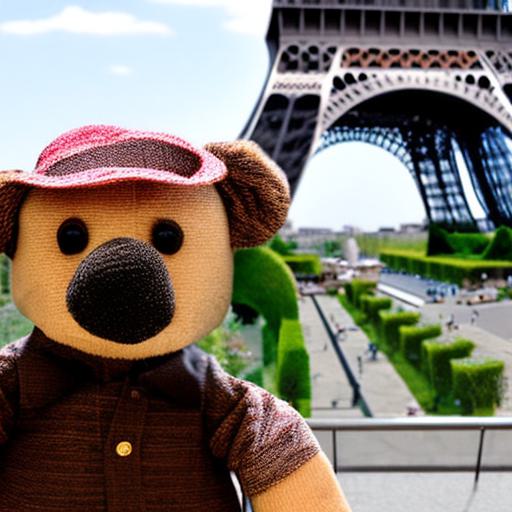} &
        \includegraphics[valign=c, width=\ww,frame]{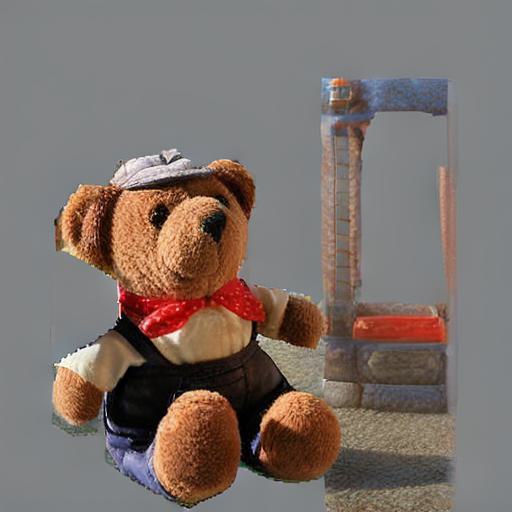} &
        \includegraphics[valign=c, width=\ww,frame]{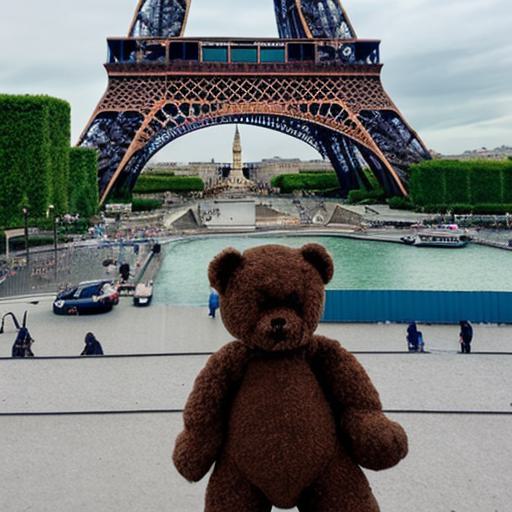} &
        \includegraphics[valign=c, width=\ww,frame]{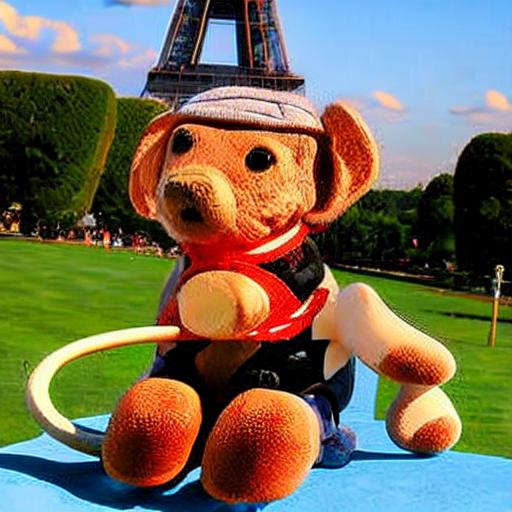} &
        \includegraphics[valign=c, width=\ww,frame]{figures/automatic_dataset_comparison/assets/bear_glass/eiffel/ours.jpg}
        \\
        \\

        &
        \multicolumn{5}{c}{``a photo of \tokenb with the Eiffel Tower in the background''}
        \\
        \\

        &
        \includegraphics[valign=c, width=\ww,frame]{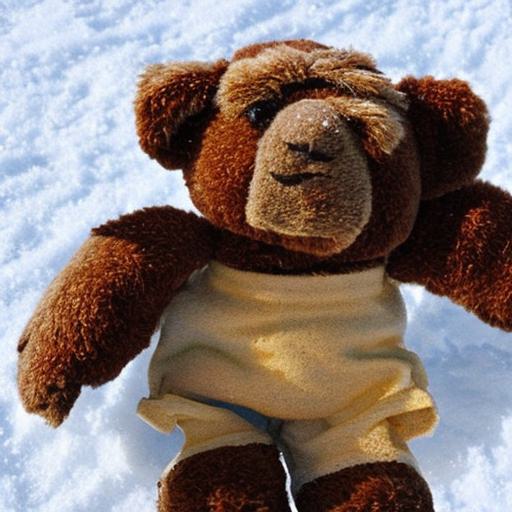} &
        \includegraphics[valign=c, width=\ww,frame]{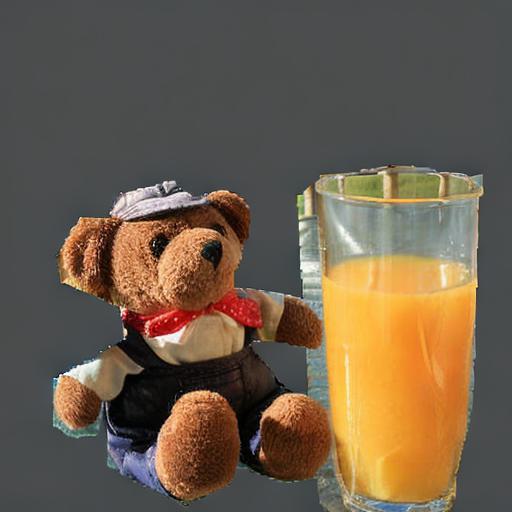} &
        \includegraphics[valign=c, width=\ww,frame]{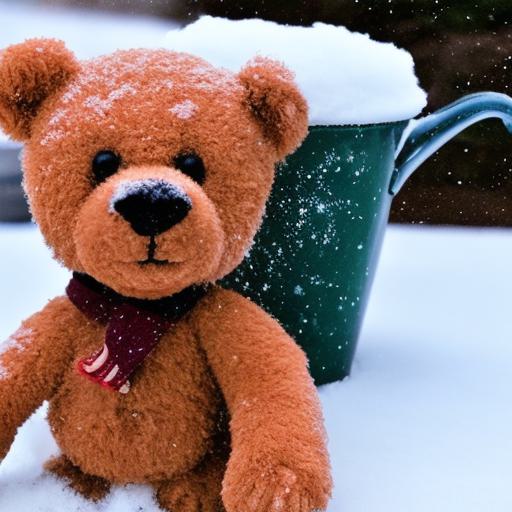} &
        \includegraphics[valign=c, width=\ww,frame]{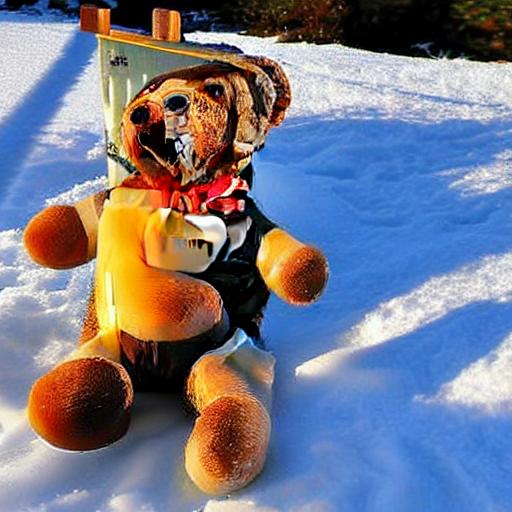} &
        \includegraphics[valign=c, width=\ww,frame]{figures/automatic_dataset_comparison/assets/bear_glass/snow/ours.jpg}
        \\
        \\

        &
        \multicolumn{5}{c}{``a photo of \tokena and \tokenb in the snow''}
        \\
        \\
        \midrule

        \\
        \raisebox{-1.3\height}[0pt][0pt]{\includegraphics[valign=c, width=\ww,frame]{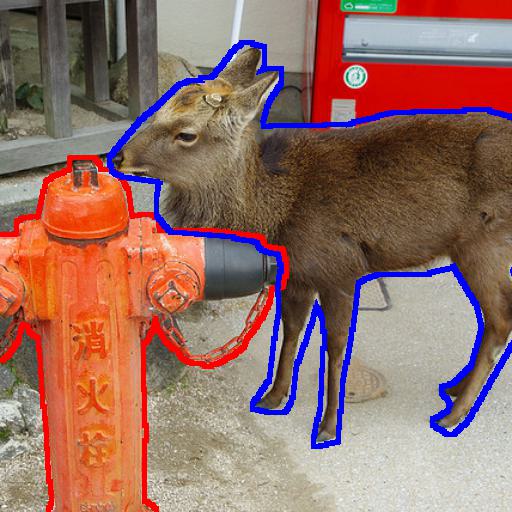}} &
        \includegraphics[valign=c, width=\ww,frame]{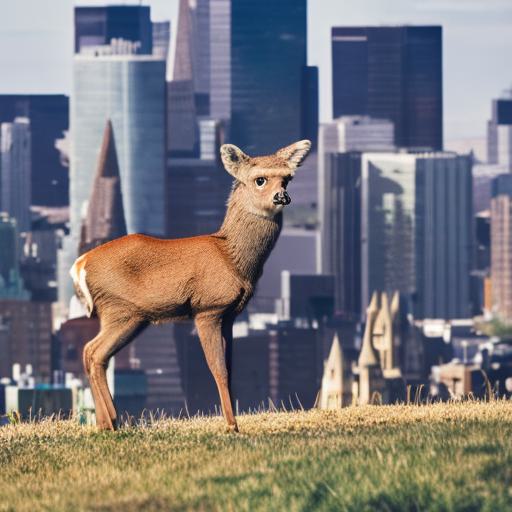} &
        \includegraphics[valign=c, width=\ww,frame]{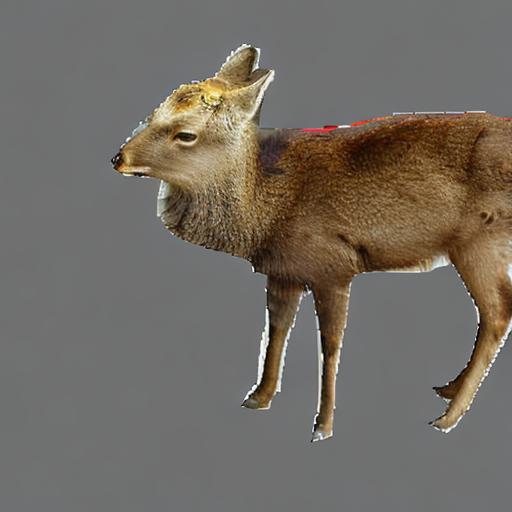} &
        \includegraphics[valign=c, width=\ww,frame]{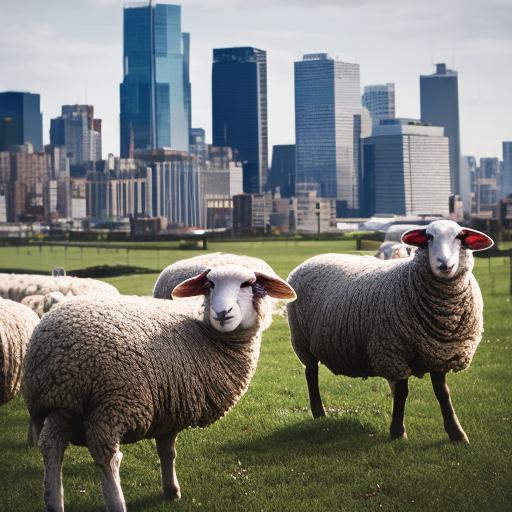} &
        \includegraphics[valign=c, width=\ww,frame]{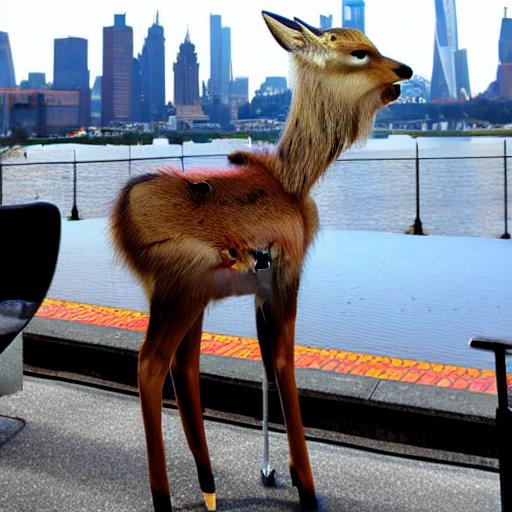} &
        \includegraphics[valign=c, width=\ww,frame]{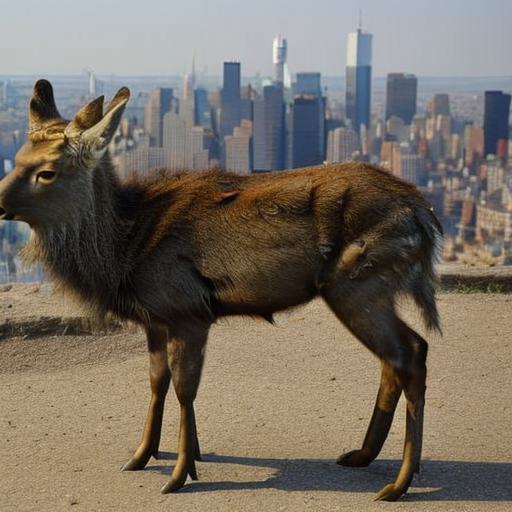}
        \\
        \\

        &
        \multicolumn{5}{c}{``a photo of \tokenb with a city in the background''}
        \\
        \\

        &
        \includegraphics[valign=c, width=\ww,frame]{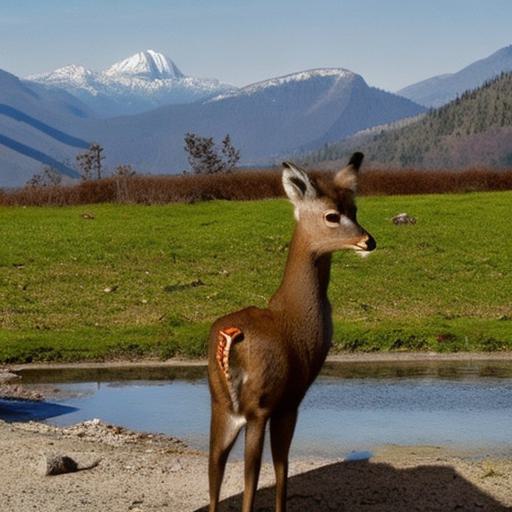} &
        \includegraphics[valign=c, width=\ww,frame]{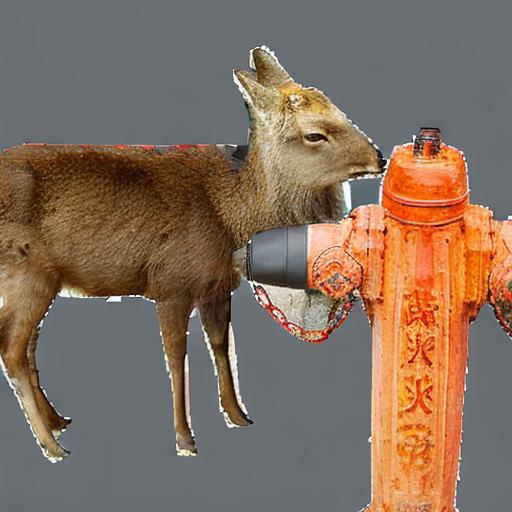} &
        \includegraphics[valign=c, width=\ww,frame]{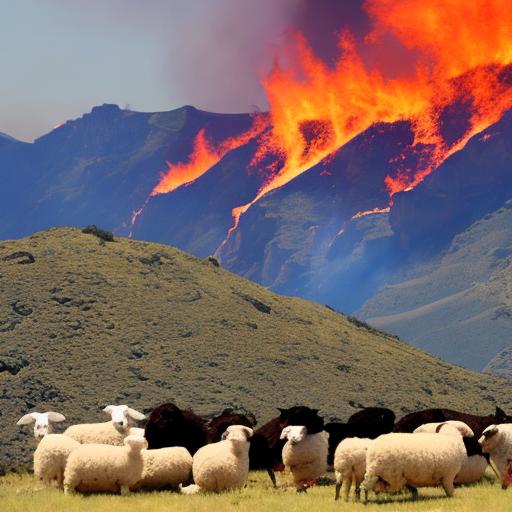} &
        \includegraphics[valign=c, width=\ww,frame]{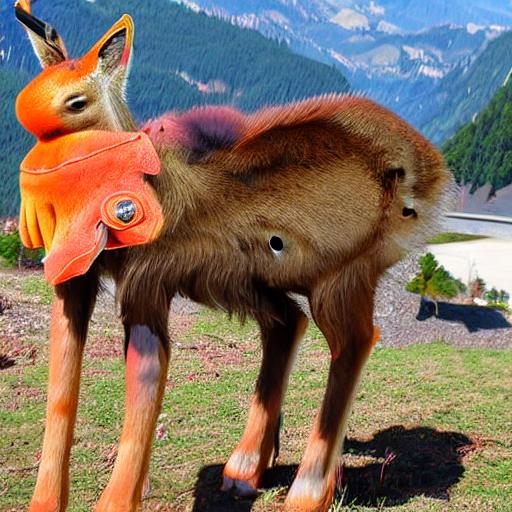} &
        \includegraphics[valign=c, width=\ww,frame]{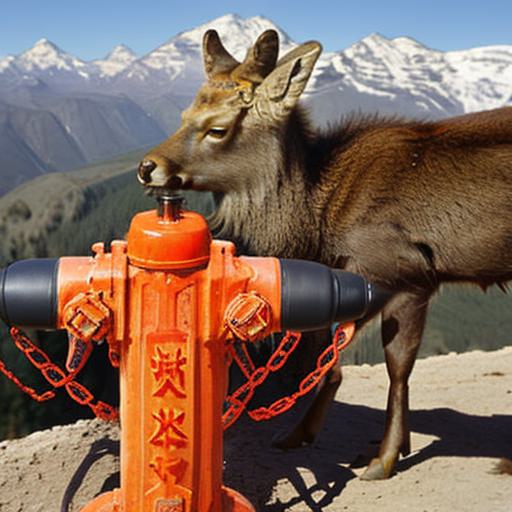}
        \\
        \\

        &
        \multicolumn{5}{c}{``a photo of \tokena and \tokenb with a mountain in the background''}
        \\
        \\

    \end{tabular}
    
    \caption{\textbf{Automatic dataset qualitative comparison:} we compare our method qualitatively against the baselines on the dataset that was generated automatically, as explained in \Cref{sec:automatic_dataset_creation}. As we can see, \maskedTI and \maskedCD struggle with preserving the concept identities, while \maskedDB struggle with generating an image the corresponds to the text prompt. ELITE is better preserving the identities than \maskedTI/\maskedCD, but they are still not recognizable enough, especially when trying to generate more than one concept. Finally, our method is able to preserve the identities as well as correspond to the text prompt, and even support generating up to four different concepts.}
    \label{fig:autoamtic_dataset_qualitative_comparison}
\end{figure*}

%% file: figures/automatic_dataset_comparison/fig_ablation.tex
\begin{figure*}[t]
    \centering
    \setlength{\tabcolsep}{0.5pt}
    \renewcommand{\arraystretch}{0.5}
    \setlength{\ww}{0.285\columnwidth}
    \begin{tabular}{c @{\hspace{10\tabcolsep}} @{\hspace{10\tabcolsep}} ccccc}

        \textbf{Inputs} &
        \textbf{Ours w/o} &
        \textbf{Ours w/o} &
        \textbf{Ours w/o} &
        \textbf{Ours w/o} &
        \textbf{Ours}
        \\

        &
        \textbf{two phases} &
        \textbf{masked loss} &
        \textbf{attention loss} &
        \textbf{union sampling} &
        \\

        \\
        \raisebox{-1.3\height}[0pt][0pt]{\includegraphics[valign=c, width=\ww,frame]{figures/automatic_dataset_comparison/assets/cat_bag/mask_overlay.jpg}} &
        \includegraphics[valign=c, width=\ww,frame]{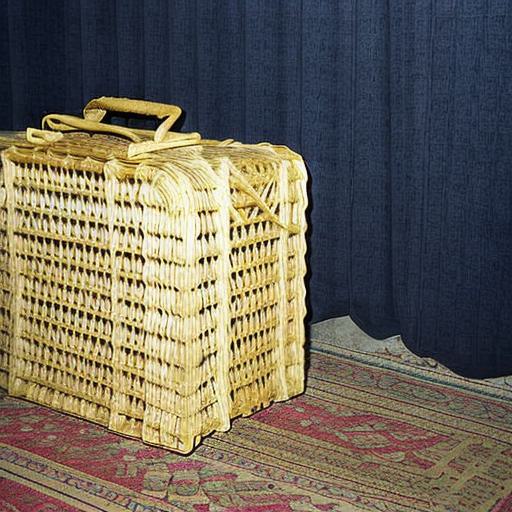} &
        \includegraphics[valign=c, width=\ww,frame]{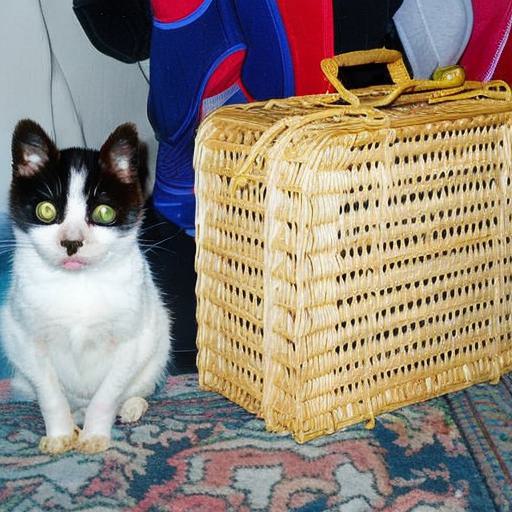} &
        \includegraphics[valign=c, width=\ww,frame]{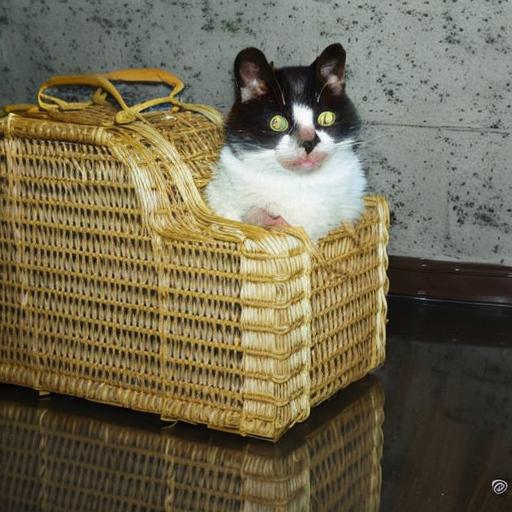} &
        \includegraphics[valign=c, width=\ww,frame]{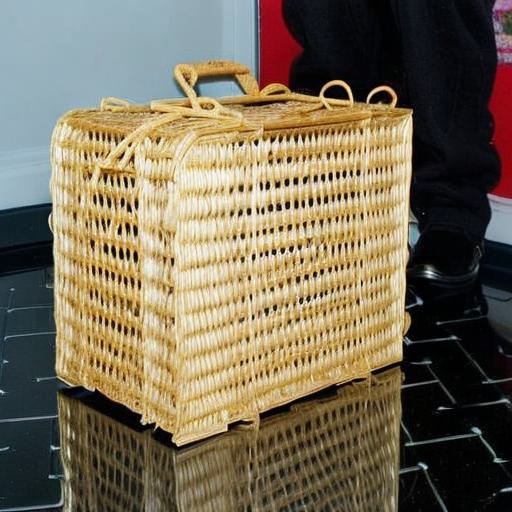} &
        \includegraphics[valign=c, width=\ww,frame]{figures/automatic_dataset_comparison/assets/cat_bag/water/ours.jpg}
        \\
        \\

        &
        \multicolumn{5}{c}{``a photo of \tokenb floating on top of water''}
        \\
        \\

        &
        \includegraphics[valign=c, width=\ww,frame]{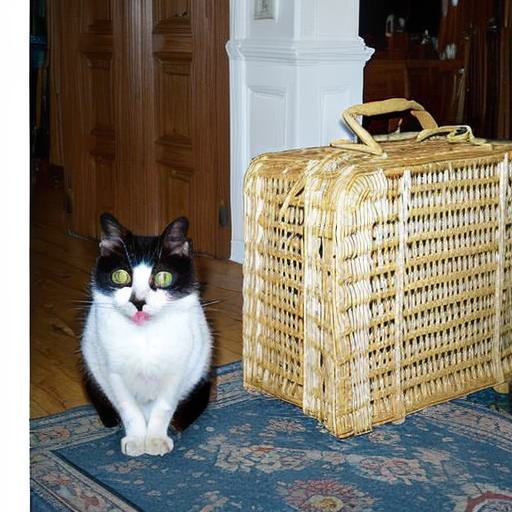} &
        \includegraphics[valign=c, width=\ww,frame]{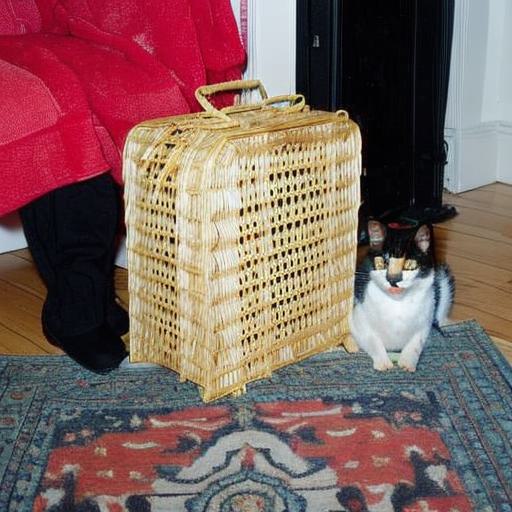} &
        \includegraphics[valign=c, width=\ww,frame]{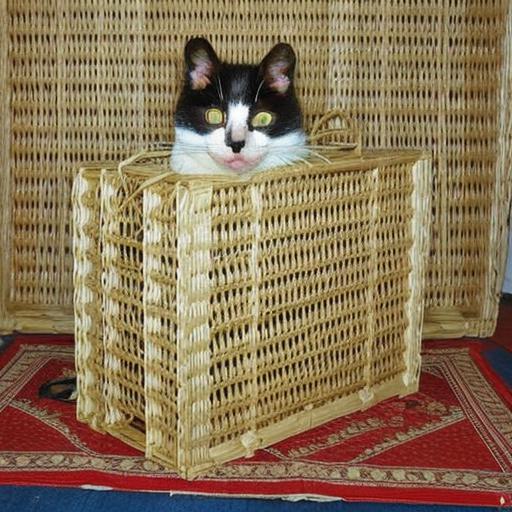} &
        \includegraphics[valign=c, width=\ww,frame]{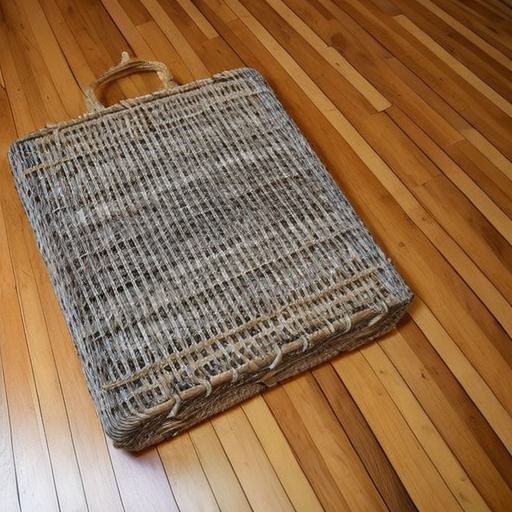} &
        \includegraphics[valign=c, width=\ww,frame]{figures/automatic_dataset_comparison/assets/cat_bag/wooden/ours.jpg}
        \\
        \\

        &
        \multicolumn{5}{c}{``a photo of \tokena and \tokenb on top of a wooden floor''}
        \\
        \\
        \midrule

        \\
        \raisebox{-1.3\height}[0pt][0pt]{\includegraphics[valign=c, width=\ww,frame]{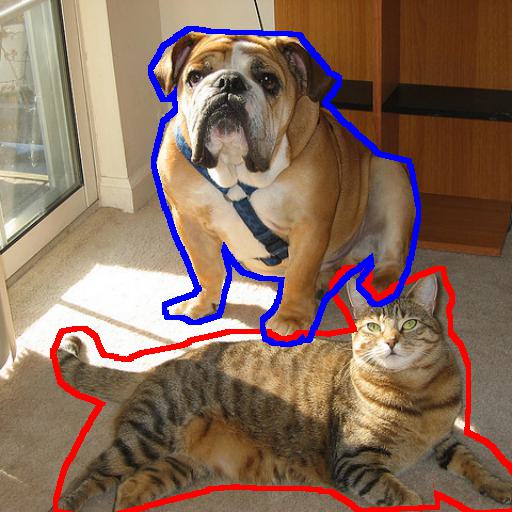}} &
        \includegraphics[valign=c, width=\ww,frame]{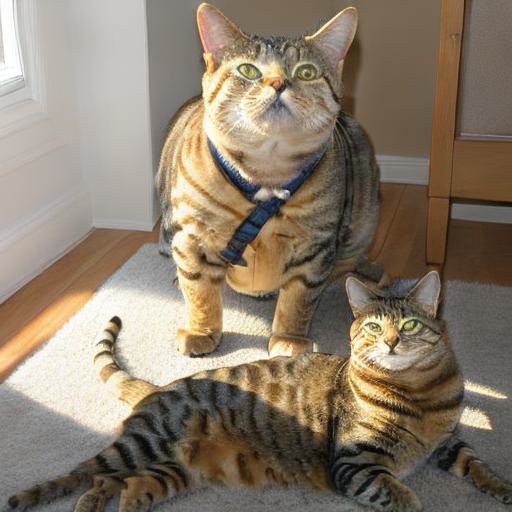} &
        \includegraphics[valign=c, width=\ww,frame]{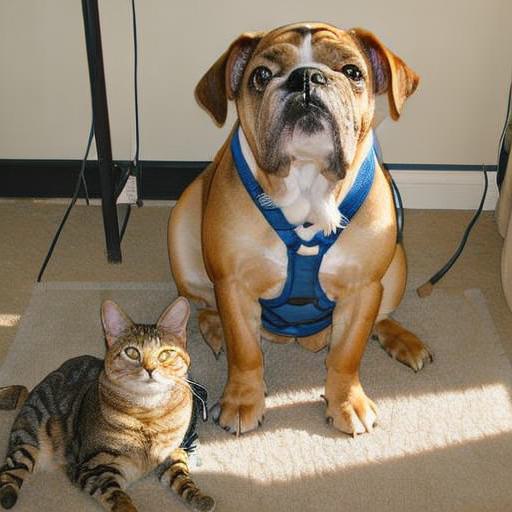} &
        \includegraphics[valign=c, width=\ww,frame]{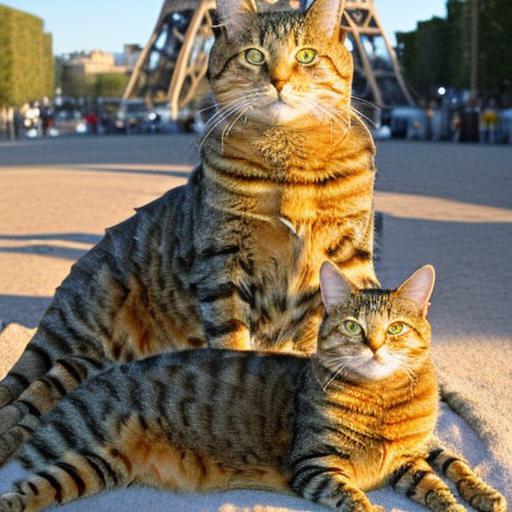} &
        \includegraphics[valign=c, width=\ww,frame]{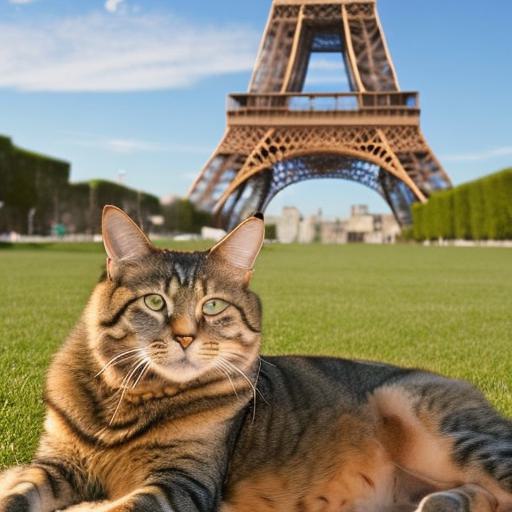} &
        \includegraphics[valign=c, width=\ww,frame]{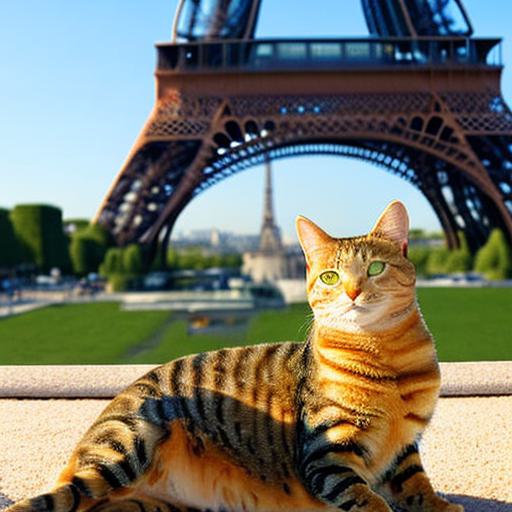}
        \\
        \\

        &
        \multicolumn{5}{c}{``a photo of \tokena with the Eiffel Tower in the background''}
        \\
        \\

        &
        \includegraphics[valign=c, width=\ww,frame]{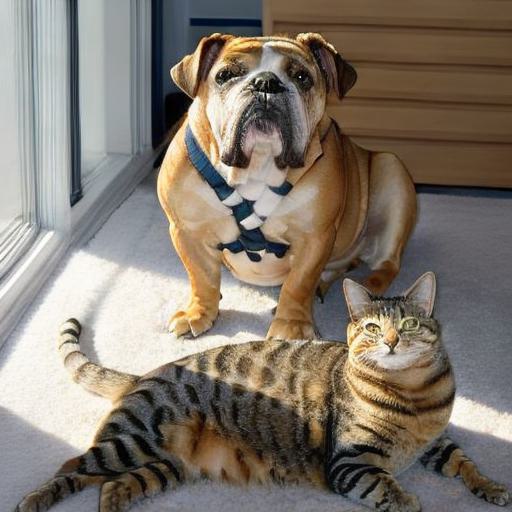} &
        \includegraphics[valign=c, width=\ww,frame]{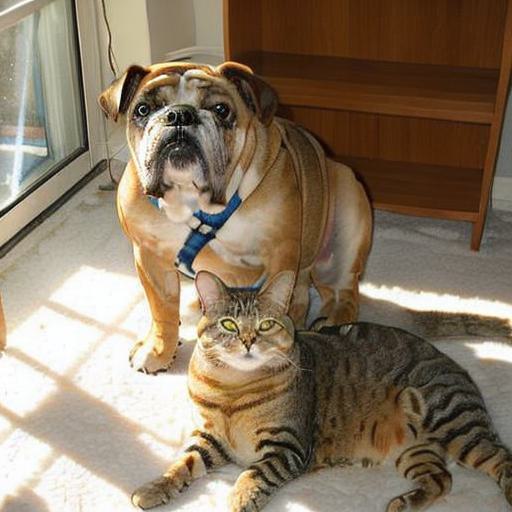} &
        \includegraphics[valign=c, width=\ww,frame]{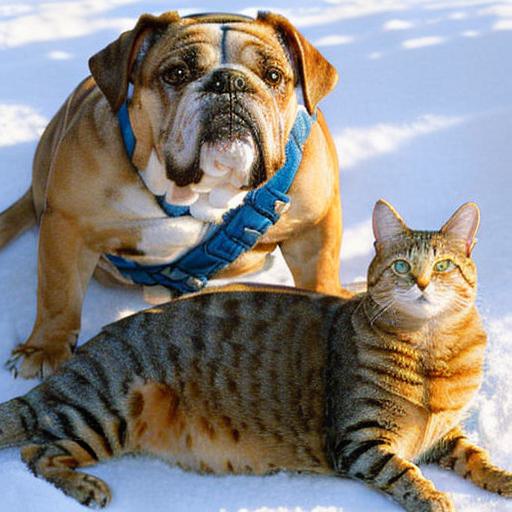} &
        \includegraphics[valign=c, width=\ww,frame]{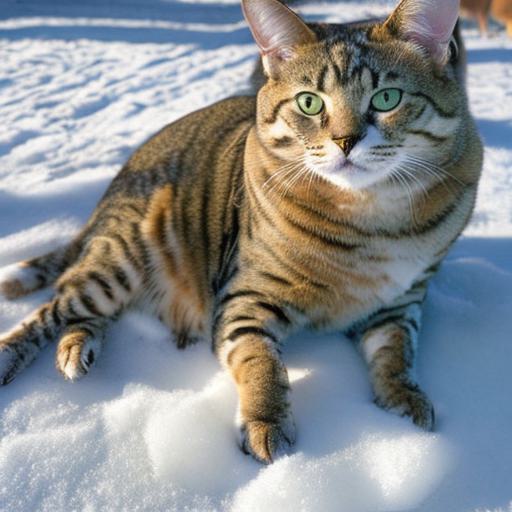} &
        \includegraphics[valign=c, width=\ww,frame]{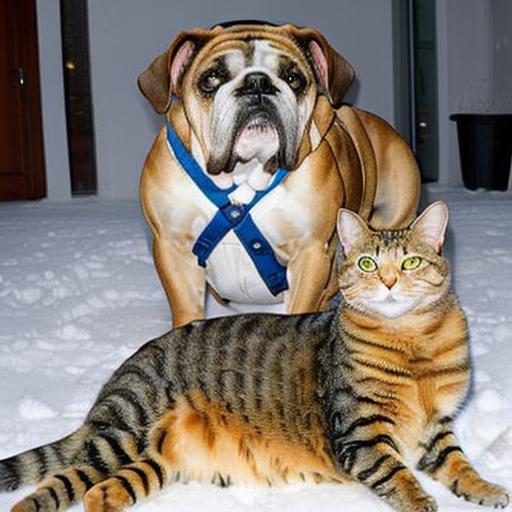}
        \\
        \\

        &
        \multicolumn{5}{c}{``a photo of \tokena and \tokenb in the snow''}
        \\
        \\
        \midrule

        \\
        \raisebox{-1.3\height}[0pt][0pt]{\includegraphics[valign=c, width=\ww,frame]{figures/automatic_dataset_comparison/assets/fire_goat/mask_overlay.jpg}} &
        \includegraphics[valign=c, width=\ww,frame]{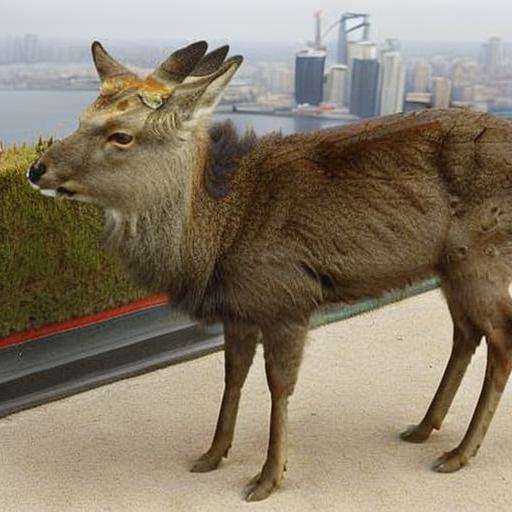} &
        \includegraphics[valign=c, width=\ww,frame]{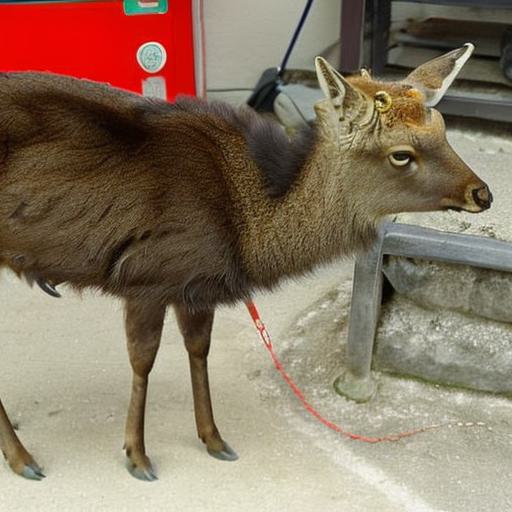} &
        \includegraphics[valign=c, width=\ww,frame]{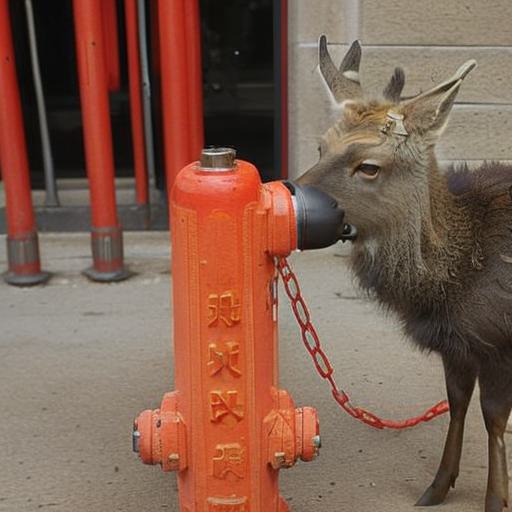} &
        \includegraphics[valign=c, width=\ww,frame]{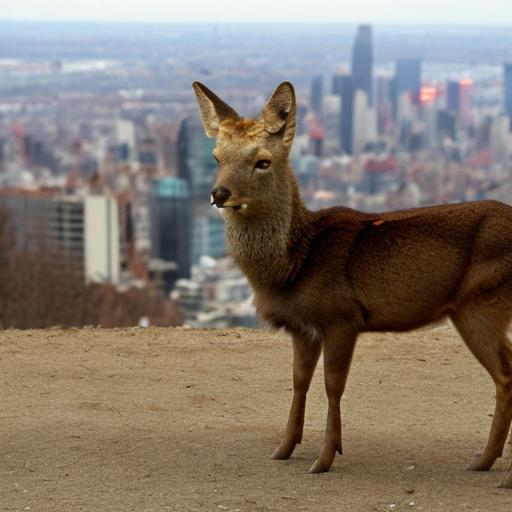} &
        \includegraphics[valign=c, width=\ww,frame]{figures/automatic_dataset_comparison/assets/fire_goat/city/ours.jpg}
        \\
        \\

        &
        \multicolumn{5}{c}{``a photo of \tokenb with a city in the background''}
        \\
        \\

        &
        \includegraphics[valign=c, width=\ww,frame]{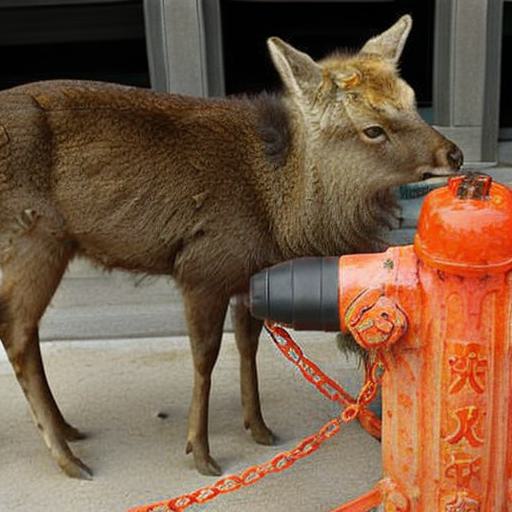} &
        \includegraphics[valign=c, width=\ww,frame]{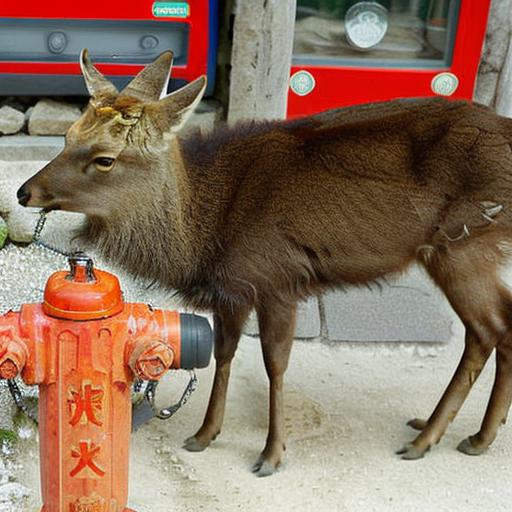} &
        \includegraphics[valign=c, width=\ww,frame]{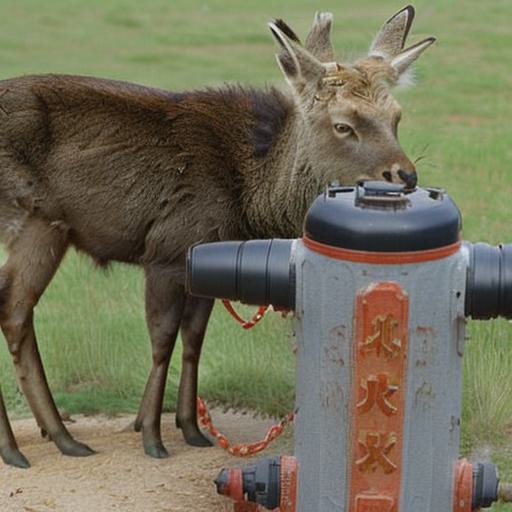} &
        \includegraphics[valign=c, width=\ww,frame]{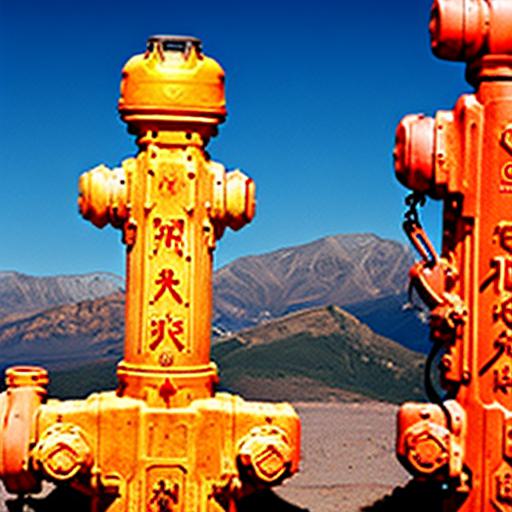} &
        \includegraphics[valign=c, width=\ww,frame]{figures/automatic_dataset_comparison/assets/fire_goat/mountain/ours.jpg}
        \\
        \\

        &
        \multicolumn{5}{c}{``a photo of \tokena and \tokenb with a mountain in the background''}
        \\
        \\

    \end{tabular}
    
    \caption{\textbf{Qualitative ablation study:} we conduct an ablation study by removing the first phase in our two-phase training regime, removing the masked diffusion loss, removing the cross-attention loss, and removing the union-sampling. As can be seen, when removing the first training phase, the model tends to correspond less to the input text prompt. In addition, when removing the masked loss, the model tends to learn also the background, which diminishes the target prompt correspondence. Furthermore, when removing the cross-attention loss, the model tends to mix between the concepts or replicate one of them. Finally, when removing the union-sampling, the model struggles with generating images with multiple concepts.}
    \label{fig:autoamtic_dataset_qualitative_comparison_ablation}
\end{figure*}

%% file: sections/appendix/implementation_details.tex
\section{Implementation Details}
\label{sec:implemetation_details}

In the following section, we start by providing some implementation details of our method. Next, we provide more details about the automatic comparison dataset creation, as well as the automatic metrics. Finally, we provide the full details of the user study we conducted.

\subsection{Method Implementation Details}

\input{tables/methods_overview_full.tex}

We based our method, as well as the baselines (except ELITE \cite{Wei2023ELITEEV}) on Stable Diffusion V2.1 \cite{Rombach2021HighResolutionIS} implementations of the HuggingFace diffusers library \cite{von-platen-etal-2022-diffusers}. For ELITE, we used the official implementation by the authors \cite{ELITE-implmentation} that used Stable Diffusion V1.4, which we had to use because their encoders were trained on this model embeddings. In addition to these four baselines, many concurrent works were proposed recently, as detailed in \Cref{tab:previous_methods_overview_full}, none of which tackles the problem of extracting \emph{multiple concepts} from a \emph{single image}.

As explained in
Section 3 of the main paper,
our method is divided into two stages: in the first stage we optimize only the text embeddings with a high learning rate of $5e^{-4}$, while in the second stage, we train both the UNet weights, and the text encoder weights with a small learning rate of $2e^{-6}$. For both stages, we used Adam optimizer \cite{Kingma2014AdamAM} with $\beta_1 = 0.9, \beta_2 = 0.99$ and a weight decay of $1e^{-8}$. For all of our experiments, we used $400$ training steps for each one of the stages, which we found to work well empirically. When applying the masked version of the baselines, we used the corresponding learning rate and optimized parameters as our method. We performed the union-sampling in both of the training stages.

The UNet of the Stable Diffusion models consisted of a series of self-attention layers followed by cross-attention layers that inject the textual information into the image formation process. This is done in various resolutions of ${8, 16, 32, 64}$.  As was shown in \cite{Hertz2022PrompttoPromptIE}, these cross-attention layers also control the layout of the generated scene, and can be utilized for generating images with the same structure but with different semantics, or edit generated images. As explained in
Section 3 of the main paper,
we utilize these cross-attention maps for disentangling between the learned concepts. To this end, we average all the cross-attention maps corresponding to each one of the newly-added personalized tokens at resolution $16 \times 16$, which was shown by \cite{Hertz2022PrompttoPromptIE} to contains most of the semantics, and normalized them to range $[0, 1]$. For brevity, we refer to this normalized averages cross-attention map as $CA_{\theta}(v_i, z_t)$, the cross-attention map between the token $v_i$ and the noisy latent $z_t$.

\subsection{Automatic Dataset Creation}
\label{sec:automatic_dataset_creation}

As explained in
Section 4.1 in the main paper,
we created an automated pipeline for creating a comparisons dataset and use it to compare our method (quantitatively and via a user study). To this end, we use COCO \cite{Lin2014MicrosoftCC} dataset, which contains images along with their instance segmentation masks. We crop COCO images into a square shape, and filter only those that contain at least two segments of distinct ``things'' type, with each segment occupying at least 15\% of the image. We also filter out concepts from COCO classes that are hard to distinguish from each other (orange, banana, broccoli, carrot, zebra, giraffe). Using this method, we extracted 50 scenes of different types. Next, we paired each of these inputs with a text prompt from a fixed list, \eg, ``a photo of \{tokens\} in the snow'', where \{tokens\} was iterated on all the combinations of the powerset of the input tokens, yielding a total number of 5400 generations per baseline. \Cref{fig:autoamtic_dataset_qualitative_comparison} presents a qualitative comparison of the baselines against our method on this automatically generated dataset. 

The fixed formats that we used are:
\begin{itemize}
    \item "a photo of \{tokens\} at the beach"
    \item "a photo of \{tokens\} in the jungle"
    \item "a photo of \{tokens\} in the snow"
    \item "a photo of \{tokens\} in the street"
    \item "a photo of \{tokens\} on top of a pink fabric"
    \item "a photo of \{tokens\} on top of a wooden floor"
    \item "a photo of \{tokens\} with a city in the background"
    \item "a photo of \{tokens\} with a mountain in the background"
    \item "a photo of \{tokens\} with the Eiffel tower in the background"
    \item "a photo of \{tokens\} floating on top of water"
\end{itemize}

As explained in
Section 4.1 in the main paper,
we focused on two evaluation metrics: prompt similarity and identity similarity. For calculating the prompt similarity we used CLIP \cite{Radford2021LearningTV} model ViT-L/14 \cite{Dosovitskiy2020AnII} implementation by HuggingFace and calculated normalized cosine similarity of the CLIP text embeddings of the input prompt (the tokens were replaced with the ground-truth classes) and CLIP image embedding of the generated image. For calculating the identity similarity, we offered a metric that supports the multi-subject case: for each generation we compare the masked version of the input image (by the input mask from the COCO dataset) with a masked version of the generated image, which we acquire by utilizing a pre-trained image segmentation model MaskFormer \cite{Cheng2021PerPixelCI} that was trained on COCO panoptic segmentation (large-sized version, SWIN \cite{Liu2021SwinTH} backbone) implemented by HuggingFace. For the image embeddings comparison, we used DINO model \cite{Caron2021EmergingPI} (base-sized model, patch size 16) that was shown \cite{Ruiz2022DreamBoothFT} to better encompass the object's identity.

\subsection{User Study Details}
\label{sec:user_study_details}

As described in
Section 4.1 in the main paper,
we conducted a user study employing the Amazon Mechanical Turk (AMT) in order to assess the human perception of the metrics of interest: prompt similarity and identity similarity. For assessing the prompt correspondence, we instructed the workers ``For each of the following images, please rank on a scale of 1 to 5 its correspondence to this text description: \{PROMPT\}'' where \{PROMPT\} is the modified text prompt resulted by replacing the special token with the class textual token (e.g., ``a photo of a cat at the beach'' instead of ``a photo of \tokena at the beach'' which was used to create the image). All the baselines, as well as our method, were presented in the same page, and the evaluators rated each result by a slider from 1 (``Do not match at all'') to 5 (``Match perfectly
''). For assessing identity similarity, we showed a masked version of the input image that contains only the object being generated, put it next to each one of the baseline results, and instructed the workers ``For each of the following image pairs, please rank on a scale of 1 to 5 if they contain the same object (1 means that they contain totally different objects and 5 means that they contain exactly the same object). The images can have different backgrounds''. The questions were presented to the raters in a random order, and we collected three ratings per question, resulting in 1215 ratings per task (prompt similarity/identity similarity). The time allotted per image-pair task was one hour, to allow the raters to properly evaluate the results without time pressure.

\input{tables/statistical_analysis.tex}
\input{tables/user_study_variences.tex}

We conducted a statistical analysis of our user study by validating that the difference between all the conditions is statistically significant using Kruskal-Wallis \cite{Kruskal1952UseOR} test ($p < 10^{-213}$). In addition, we used Tukey's honestly significant difference procedure \cite{Tukey1949ComparingIM} to show that the comparison of our method against all the baselines is statistically significant, as detailed in \Cref{tab:statistical_analysis}. The means and variances of the user study are reported in \Cref{tab:user_study_variances}.

\subsection{Blended Latent Diffusion Integration}
As explained in
Section 4.2 in the main paper,
in order to edit an image using the extracted concepts from another image, we utilized Blended Latent Diffusion \cite{avrahami2022blended,avrahami2022blendedlatent} off-the-shelf text-driven image editing method. As shown in \Cref{fig:bld_additional}, we can perform it in an iterative manner, editing the image region-by-region. That way, we can edit the image in an elaborated manner by giving a different text prompt per region.

%% file: tables/methods_overview_full.tex
\begin{table}
    \centering
    \caption{\textbf{Personalization baselines comparison.} Our method is the first to suggest a solution for the problem of \emph{single image} with \emph{multiple concepts} personalization. This is an extended version of Table 1 in the main paper that includes concurrent works. Only the first four methods have an open-source implementation.}
    \begin{adjustbox}{width=1\columnwidth}
        \begin{tabular}{>{\columncolor[gray]{0.95}}lcc}
            \toprule
            
            \textbf{Method} & 
            Single input &
            Multi-concept
            \\

            &
            image &
            output
            \\
            
            \midrule

            Textual Inversion \cite{Gal2022AnII} &
            \xmark &
            \xmark
            \\

            Dreambooth \cite{Ruiz2022DreamBoothFT} &
            \xmark &
            \xmark
            \\

            Custom Diffusion \cite{Kumari2022MultiConceptCO} &
            \xmark & 
            \cmark
            \\

            ELITE \cite{Wei2023ELITEEV} &
            \cmark &
            \xmark
            \\

            E4T \cite{Gal2023EncoderbasedDT} &
            \cmark & 
            \xmark
            \\

            SVDiff \cite{Han2023SVDiffCP} &
            \xmark & 
            \cmark
            \\

            SuTI \cite{Chen2023SubjectdrivenTG} &
            \xmark & 
            \xmark
            \\

            Taming \cite{Jia2023TamingEF} &
            \cmark & 
            \xmark
            \\

            InstantBooth \cite{Shi2023InstantBoothPT} &
            \cmark & 
            \xmark
            \\

            XTI \cite{Voynov2023PET} &
            \cmark & 
            \xmark
            \\

            Perfusion \cite{Tewel2023KeyLockedRO} &
            \xmark & 
            \cmark
            \\

            \midrule
            Ours &
            \cmark &
            \cmark
            \\
            
            \bottomrule
        \end{tabular}
    \end{adjustbox}
    \label{tab:previous_methods_overview_full}
\end{table}

%% file: tables/statistical_analysis.tex
\begin{table}
    \centering
    \caption{\textbf{Statistical analysis.} We use Tukey's  honestly significant difference procedure \cite{Tukey1949ComparingIM} to test whether the differences
    between mean scores in our user study are statistically significant.}
    \begin{adjustbox}{width=1\columnwidth}
        \begin{tabular}{>{\columncolor[gray]{0.95}}c>{\columncolor[gray]{0.95}}ccc}
            \toprule
            
            \textbf{Method 1} &
            \textbf{Method 2} &
            Prompt similarity &
            Identities similarity
            \\

            &
            &
            p-value &
            p-value
            \\
            \midrule

            \maskedTI &
            Ours &
            $p < 10^{-10}$ &
            $p < 10^{-10}$
            \\

            \maskedDB &
            Ours &
            $p < 0.05$ &
            $p < 10^{-10}$
            \\

            \maskedCD &
            Ours &
            $p < 10^{-7}$ &
            $p < 10^{-10}$
            \\

            ELITE &
            Ours &
            $p < 10^{-10}$ &
            $p < 10^{-10}$
            \\
            
            \bottomrule
        \end{tabular}
    \end{adjustbox}
    \label{tab:statistical_analysis}
\end{table}
    

%% file: tables/user_study_variences.tex
\begin{table}
    \centering
    \caption{\textbf{Users' rankings means and variances.} the means and variances of the rankings that are reported in the user study.}
        \begin{tabular}{>{\columncolor[gray]{0.95}}lcc}
            \toprule
            
            \textbf{Method} & 
            Identity similarity &
            Prompt similarity
            \\
            
            \midrule

            \maskedTI &
            $2.69 \pm 1.3$ &
            $3.88 \pm 1.21$
            \\

            \maskedDB &
            $3.97 \pm 0.95$ &
            $2.37 \pm 1.11$
            \\

            \maskedCD &
            $2.47 \pm 1.3$ &
            $4.08 \pm 1.12$
            \\

            ELITE &
            $3.05 \pm 1.31$ &
            $3.53 \pm 1.31$
            \\

            Ours &
            $3.56 \pm 1.27$ &
            $3.85 \pm 1.21$
            \\
            
            \bottomrule
        \end{tabular}
    \label{tab:user_study_variances}
\end{table}

%% file: sections/appendix/societal_impact.tex
\section{Societal Impact}
\label{sec:societal_impact}

Our method may help democratizing content creation, empowering individuals with limited artistic skills or resources to produce visually engaging content. This may not only open up opportunities for individuals who were previously excluded, but also foster a more diverse and inclusive creative landscape.

In addition, it can help generate visuals that align with specific rare cultural contexts, where the input may be scarce and contain a single image. This may enhance cultural appreciation, foster a sense of belonging, and promote intercultural understanding.

On the other hand, our method, may cause intellectual property and copyright issues when being used on an existing copyrighted content as reference. In addition, malicious users can exploit this model to create realistic but fabricated images, potentially deceiving other individuals.

%% file: main.bbl

\begin{thebibliography}{91}


\ifx \showCODEN    \undefined \def \showCODEN     #1{\unskip}     \fi
\ifx \showDOI      \undefined \def \showDOI       #1{#1}\fi
\ifx \showISBNx    \undefined \def \showISBNx     #1{\unskip}     \fi
\ifx \showISBNxiii \undefined \def \showISBNxiii  #1{\unskip}     \fi
\ifx \showISSN     \undefined \def \showISSN      #1{\unskip}     \fi
\ifx \showLCCN     \undefined \def \showLCCN      #1{\unskip}     \fi
\ifx \shownote     \undefined \def \shownote      #1{#1}          \fi
\ifx \showarticletitle \undefined \def \showarticletitle #1{#1}   \fi
\ifx \showURL      \undefined \def \showURL       {\relax}        \fi
\providecommand\bibfield[2]{#2}
\providecommand\bibinfo[2]{#2}
\providecommand\natexlab[1]{#1}
\providecommand\showeprint[2][]{arXiv:#2}

\bibitem[Abdal et~al\mbox{.}(2019)]%
        {abdal2019image2stylegan}
\bibfield{author}{\bibinfo{person}{Rameen Abdal}, \bibinfo{person}{Yipeng Qin},
  {and} \bibinfo{person}{Peter Wonka}.} \bibinfo{year}{2019}\natexlab{}.
\newblock \showarticletitle{Image2stylegan: How to embed images into the
  stylegan latent space?}. In \bibinfo{booktitle}{\emph{Proceedings of the
  IEEE/CVF International Conference on Computer Vision}}.
  \bibinfo{pages}{4432--4441}.
\newblock


\bibitem[Abdal et~al\mbox{.}(2020)]%
        {abdal2020image2stylegan++}
\bibfield{author}{\bibinfo{person}{Rameen Abdal}, \bibinfo{person}{Yipeng Qin},
  {and} \bibinfo{person}{Peter Wonka}.} \bibinfo{year}{2020}\natexlab{}.
\newblock \showarticletitle{Image2stylegan++: How to edit the embedded
  images?}. In \bibinfo{booktitle}{\emph{Proceedings of the IEEE/CVF conference
  on computer vision and pattern recognition}}. \bibinfo{pages}{8296--8305}.
\newblock


\bibitem[Alaluf et~al\mbox{.}(2021)]%
        {alaluf2021hyperstyle}
\bibfield{author}{\bibinfo{person}{Yuval Alaluf}, \bibinfo{person}{Omer Tov},
  \bibinfo{person}{Ron Mokady}, \bibinfo{person}{Rinon Gal}, {and}
  \bibinfo{person}{Amit~Haim Bermano}.} \bibinfo{year}{2021}\natexlab{}.
\newblock \showarticletitle{HyperStyle: StyleGAN Inversion with HyperNetworks
  for Real Image Editing}.
\newblock \bibinfo{journal}{\emph{2022 IEEE/CVF Conference on Computer Vision
  and Pattern Recognition (CVPR)}} (\bibinfo{year}{2021}),
  \bibinfo{pages}{18490--18500}.
\newblock
\urldef\tempurl%
\url{https://api.semanticscholar.org/CorpusID:244729249}
\showURL{%
\tempurl}


\bibitem[Avrahami et~al\mbox{.}(2023a)]%
        {avrahami2022blendedlatent}
\bibfield{author}{\bibinfo{person}{Omri Avrahami}, \bibinfo{person}{Ohad
  Fried}, {and} \bibinfo{person}{Dani Lischinski}.}
  \bibinfo{year}{2023}\natexlab{a}.
\newblock \showarticletitle{Blended Latent Diffusion}.
\newblock \bibinfo{journal}{\emph{ACM Trans. Graph.}} \bibinfo{volume}{42},
  \bibinfo{number}{4}, Article \bibinfo{articleno}{149} (\bibinfo{date}{jul}
  \bibinfo{year}{2023}), \bibinfo{numpages}{11}~pages.
\newblock
\showISSN{0730-0301}
\urldef\tempurl%
\url{https://doi.org/10.1145/3592450}
\showDOI{\tempurl}


\bibitem[Avrahami et~al\mbox{.}(2023b)]%
        {Avrahami2022SpaTextSR}
\bibfield{author}{\bibinfo{person}{Omri Avrahami}, \bibinfo{person}{Thomas
  Hayes}, \bibinfo{person}{Oran Gafni}, \bibinfo{person}{Sonal Gupta},
  \bibinfo{person}{Yaniv Taigman}, \bibinfo{person}{Devi Parikh},
  \bibinfo{person}{Dani Lischinski}, \bibinfo{person}{Ohad Fried}, {and}
  \bibinfo{person}{Xi Yin}.} \bibinfo{year}{2023}\natexlab{b}.
\newblock \showarticletitle{SpaText: Spatio-Textual Representation for
  Controllable Image Generation}. In \bibinfo{booktitle}{\emph{Proceedings of
  the IEEE/CVF Conference on Computer Vision and Pattern Recognition (CVPR)}}.
  \bibinfo{pages}{18370--18380}.
\newblock


\bibitem[Avrahami et~al\mbox{.}(2022)]%
        {avrahami2022blended}
\bibfield{author}{\bibinfo{person}{Omri Avrahami}, \bibinfo{person}{Dani
  Lischinski}, {and} \bibinfo{person}{Ohad Fried}.}
  \bibinfo{year}{2022}\natexlab{}.
\newblock \showarticletitle{Blended diffusion for text-driven editing of
  natural images}. In \bibinfo{booktitle}{\emph{Proceedings of the IEEE/CVF
  Conference on Computer Vision and Pattern Recognition}}.
  \bibinfo{pages}{18208--18218}.
\newblock


\bibitem[Bar-Tal et~al\mbox{.}(2022)]%
        {bar2022text2live}
\bibfield{author}{\bibinfo{person}{Omer Bar-Tal}, \bibinfo{person}{Dolev
  Ofri-Amar}, \bibinfo{person}{Rafail Fridman}, \bibinfo{person}{Yoni Kasten},
  {and} \bibinfo{person}{Tali Dekel}.} \bibinfo{year}{2022}\natexlab{}.
\newblock \showarticletitle{Text2live: Text-driven layered image and video
  editing}. In \bibinfo{booktitle}{\emph{European conference on computer
  vision}}. Springer, \bibinfo{pages}{707--723}.
\newblock


\bibitem[Bar-Tal et~al\mbox{.}(2023)]%
        {BarTal2023MultiDiffusionFD}
\bibfield{author}{\bibinfo{person}{Omer Bar-Tal}, \bibinfo{person}{Lior Yariv},
  \bibinfo{person}{Yaron Lipman}, {and} \bibinfo{person}{Tali Dekel}.}
  \bibinfo{year}{2023}\natexlab{}.
\newblock \showarticletitle{Multidiffusion: Fusing diffusion paths for
  controlled image generation}.
\newblock  (\bibinfo{year}{2023}).
\newblock


\bibitem[Bau et~al\mbox{.}(2021)]%
        {bau2021paint}
\bibfield{author}{\bibinfo{person}{David Bau}, \bibinfo{person}{Alex Andonian},
  \bibinfo{person}{Audrey Cui}, \bibinfo{person}{YeonHwan Park},
  \bibinfo{person}{Ali Jahanian}, \bibinfo{person}{Aude Oliva}, {and}
  \bibinfo{person}{Antonio Torralba}.} \bibinfo{year}{2021}\natexlab{}.
\newblock \bibinfo{title}{Paint by Word}.
\newblock
\newblock
\showeprint[arxiv]{2103.10951}~[cs.CV]


\bibitem[Bau et~al\mbox{.}(2019)]%
        {Bau2019SemanticPM}
\bibfield{author}{\bibinfo{person}{David Bau}, \bibinfo{person}{Hendrik
  Strobelt}, \bibinfo{person}{William~S. Peebles}, \bibinfo{person}{Jonas
  Wulff}, \bibinfo{person}{Bolei Zhou}, \bibinfo{person}{Jun-Yan Zhu}, {and}
  \bibinfo{person}{Antonio Torralba}.} \bibinfo{year}{2019}\natexlab{}.
\newblock \showarticletitle{Semantic photo manipulation with a generative image
  prior}.
\newblock \bibinfo{journal}{\emph{ACM Transactions on Graphics (TOG)}}
  \bibinfo{volume}{38} (\bibinfo{year}{2019}), \bibinfo{pages}{1 -- 11}.
\newblock


\bibitem[Brooks et~al\mbox{.}(2023)]%
        {brooks2022instructpix2pix}
\bibfield{author}{\bibinfo{person}{Tim Brooks}, \bibinfo{person}{Aleksander
  Holynski}, {and} \bibinfo{person}{Alexei~A. Efros}.}
  \bibinfo{year}{2023}\natexlab{}.
\newblock \showarticletitle{InstructPix2Pix: Learning to Follow Image Editing
  Instructions}. In \bibinfo{booktitle}{\emph{CVPR}}.
\newblock


\bibitem[Caron et~al\mbox{.}(2021)]%
        {Caron2021EmergingPI}
\bibfield{author}{\bibinfo{person}{Mathilde Caron}, \bibinfo{person}{Hugo
  Touvron}, \bibinfo{person}{Ishan Misra}, \bibinfo{person}{Herv'e J'egou},
  \bibinfo{person}{Julien Mairal}, \bibinfo{person}{Piotr Bojanowski}, {and}
  \bibinfo{person}{Armand Joulin}.} \bibinfo{year}{2021}\natexlab{}.
\newblock \showarticletitle{Emerging Properties in Self-Supervised Vision
  Transformers}.
\newblock \bibinfo{journal}{\emph{2021 IEEE/CVF International Conference on
  Computer Vision (ICCV)}} (\bibinfo{year}{2021}), \bibinfo{pages}{9630--9640}.
\newblock


\bibitem[Chang et~al\mbox{.}(2023)]%
        {chang2023muse}
\bibfield{author}{\bibinfo{person}{Huiwen Chang}, \bibinfo{person}{Han Zhang},
  \bibinfo{person}{Jarred Barber}, \bibinfo{person}{AJ Maschinot},
  \bibinfo{person}{Jos{\'e} Lezama}, \bibinfo{person}{Lu Jiang},
  \bibinfo{person}{Ming Yang}, \bibinfo{person}{Kevin~P. Murphy},
  \bibinfo{person}{William~T. Freeman}, \bibinfo{person}{Michael Rubinstein},
  \bibinfo{person}{Yuanzhen Li}, {and} \bibinfo{person}{Dilip Krishnan}.}
  \bibinfo{year}{2023}\natexlab{}.
\newblock \showarticletitle{Muse: Text-To-Image Generation via Masked
  Generative Transformers}. In \bibinfo{booktitle}{\emph{International
  Conference on Machine Learning}}.
\newblock
\urldef\tempurl%
\url{https://api.semanticscholar.org/CorpusID:255372955}
\showURL{%
\tempurl}


\bibitem[Chefer et~al\mbox{.}(2023)]%
        {Chefer2023AttendandExciteAS}
\bibfield{author}{\bibinfo{person}{Hila Chefer}, \bibinfo{person}{Yuval
  Alaluf}, \bibinfo{person}{Yael Vinker}, \bibinfo{person}{Lior Wolf}, {and}
  \bibinfo{person}{Daniel Cohen-Or}.} \bibinfo{year}{2023}\natexlab{}.
\newblock \showarticletitle{Attend-and-Excite: Attention-Based Semantic
  Guidance for Text-to-Image Diffusion Models}.
\newblock \bibinfo{journal}{\emph{ACM Transactions on Graphics (TOG)}}
  \bibinfo{volume}{42} (\bibinfo{year}{2023}), \bibinfo{pages}{1 -- 10}.
\newblock
\urldef\tempurl%
\url{https://api.semanticscholar.org/CorpusID:256416326}
\showURL{%
\tempurl}


\bibitem[Chefer et~al\mbox{.}(2020)]%
        {Chefer2020TransformerIB}
\bibfield{author}{\bibinfo{person}{Hila Chefer}, \bibinfo{person}{Shir Gur},
  {and} \bibinfo{person}{Lior Wolf}.} \bibinfo{year}{2020}\natexlab{}.
\newblock \showarticletitle{Transformer Interpretability Beyond Attention
  Visualization}.
\newblock \bibinfo{journal}{\emph{2021 IEEE/CVF Conference on Computer Vision
  and Pattern Recognition (CVPR)}} (\bibinfo{year}{2020}),
  \bibinfo{pages}{782--791}.
\newblock


\bibitem[Chefer et~al\mbox{.}(2021)]%
        {Chefer2021GenericAE}
\bibfield{author}{\bibinfo{person}{Hila Chefer}, \bibinfo{person}{Shir Gur},
  {and} \bibinfo{person}{Lior Wolf}.} \bibinfo{year}{2021}\natexlab{}.
\newblock \showarticletitle{Generic Attention-model Explainability for
  Interpreting Bi-Modal and Encoder-Decoder Transformers}.
\newblock \bibinfo{journal}{\emph{2021 IEEE/CVF International Conference on
  Computer Vision (ICCV)}} (\bibinfo{year}{2021}), \bibinfo{pages}{387--396}.
\newblock


\bibitem[Chen et~al\mbox{.}(2023)]%
        {Chen2023SubjectdrivenTG}
\bibfield{author}{\bibinfo{person}{Wenhu Chen}, \bibinfo{person}{Hexiang Hu},
  \bibinfo{person}{Yandong Li}, \bibinfo{person}{Nataniel Rui},
  \bibinfo{person}{Xuhui Jia}, \bibinfo{person}{Ming-Wei Chang}, {and}
  \bibinfo{person}{William~W. Cohen}.} \bibinfo{year}{2023}\natexlab{}.
\newblock \showarticletitle{Subject-driven Text-to-Image Generation via
  Apprenticeship Learning}.
\newblock \bibinfo{journal}{\emph{ArXiv}}  \bibinfo{volume}{abs/2304.00186}
  (\bibinfo{year}{2023}).
\newblock


\bibitem[Cheng et~al\mbox{.}(2021)]%
        {Cheng2021PerPixelCI}
\bibfield{author}{\bibinfo{person}{Bowen Cheng}, \bibinfo{person}{Alexander~G.
  Schwing}, {and} \bibinfo{person}{Alexander Kirillov}.}
  \bibinfo{year}{2021}\natexlab{}.
\newblock \showarticletitle{Per-Pixel Classification is Not All You Need for
  Semantic Segmentation}. In \bibinfo{booktitle}{\emph{Neural Information
  Processing Systems}}.
\newblock


\bibitem[Cohen et~al\mbox{.}(2022)]%
        {Cohen2022ThisIM}
\bibfield{author}{\bibinfo{person}{Niv Cohen}, \bibinfo{person}{Rinon Gal},
  \bibinfo{person}{Eli~A Meirom}, \bibinfo{person}{Gal Chechik}, {and}
  \bibinfo{person}{Yuval Atzmon}.} \bibinfo{year}{2022}\natexlab{}.
\newblock \showarticletitle{“This is my unicorn, Fluffy”: Personalizing
  frozen vision-language representations}. In
  \bibinfo{booktitle}{\emph{European Conference on Computer Vision}}. Springer,
  \bibinfo{pages}{558--577}.
\newblock


\bibitem[Couairon et~al\mbox{.}(2022)]%
        {couairon2022diffedit}
\bibfield{author}{\bibinfo{person}{Guillaume Couairon}, \bibinfo{person}{Jakob
  Verbeek}, \bibinfo{person}{Holger Schwenk}, {and} \bibinfo{person}{Matthieu
  Cord}.} \bibinfo{year}{2022}\natexlab{}.
\newblock \showarticletitle{DiffEdit: Diffusion-based semantic image editing
  with mask guidance}. In \bibinfo{booktitle}{\emph{The Eleventh International
  Conference on Learning Representations}}.
\newblock


\bibitem[Crowson et~al\mbox{.}(2022)]%
        {crowson2022vqgan}
\bibfield{author}{\bibinfo{person}{Katherine Crowson}, \bibinfo{person}{Stella
  Biderman}, \bibinfo{person}{Daniel Kornis}, \bibinfo{person}{Dashiell
  Stander}, \bibinfo{person}{Eric Hallahan}, \bibinfo{person}{Louis
  Castricato}, {and} \bibinfo{person}{Edward Raff}.}
  \bibinfo{year}{2022}\natexlab{}.
\newblock \showarticletitle{Vqgan-clip: Open domain image generation and
  editing with natural language guidance}. In
  \bibinfo{booktitle}{\emph{European Conference on Computer Vision}}. Springer,
  \bibinfo{pages}{88--105}.
\newblock


\bibitem[Dosovitskiy et~al\mbox{.}(2020)]%
        {Dosovitskiy2020AnII}
\bibfield{author}{\bibinfo{person}{Alexey Dosovitskiy}, \bibinfo{person}{Lucas
  Beyer}, \bibinfo{person}{Alexander Kolesnikov}, \bibinfo{person}{Dirk
  Weissenborn}, \bibinfo{person}{Xiaohua Zhai}, \bibinfo{person}{Thomas
  Unterthiner}, \bibinfo{person}{Mostafa Dehghani}, \bibinfo{person}{Matthias
  Minderer}, \bibinfo{person}{Georg Heigold}, \bibinfo{person}{Sylvain Gelly},
  {et~al\mbox{.}}} \bibinfo{year}{2020}\natexlab{}.
\newblock \showarticletitle{An Image is Worth 16x16 Words: Transformers for
  Image Recognition at Scale}. In \bibinfo{booktitle}{\emph{International
  Conference on Learning Representations}}.
\newblock


\bibitem[Gafni et~al\mbox{.}(2022)]%
        {gafni2022make}
\bibfield{author}{\bibinfo{person}{Oran Gafni}, \bibinfo{person}{Adam Polyak},
  \bibinfo{person}{Oron Ashual}, \bibinfo{person}{Shelly Sheynin},
  \bibinfo{person}{Devi Parikh}, {and} \bibinfo{person}{Yaniv Taigman}.}
  \bibinfo{year}{2022}\natexlab{}.
\newblock \showarticletitle{Make-a-scene: Scene-based text-to-image generation
  with human priors}. In \bibinfo{booktitle}{\emph{European Conference on
  Computer Vision}}. Springer, \bibinfo{pages}{89--106}.
\newblock


\bibitem[Gal et~al\mbox{.}(2022)]%
        {Gal2022AnII}
\bibfield{author}{\bibinfo{person}{Rinon Gal}, \bibinfo{person}{Yuval Alaluf},
  \bibinfo{person}{Yuval Atzmon}, \bibinfo{person}{Or Patashnik},
  \bibinfo{person}{Amit~Haim Bermano}, \bibinfo{person}{Gal Chechik}, {and}
  \bibinfo{person}{Daniel Cohen-or}.} \bibinfo{year}{2022}\natexlab{}.
\newblock \showarticletitle{An Image is Worth One Word: Personalizing
  Text-to-Image Generation using Textual Inversion}. In
  \bibinfo{booktitle}{\emph{The Eleventh International Conference on Learning
  Representations}}.
\newblock


\bibitem[Gal et~al\mbox{.}(2023)]%
        {Gal2023EncoderbasedDT}
\bibfield{author}{\bibinfo{person}{Rinon Gal}, \bibinfo{person}{Moab Arar},
  \bibinfo{person}{Yuval Atzmon}, \bibinfo{person}{Amit~Haim Bermano},
  \bibinfo{person}{Gal Chechik}, {and} \bibinfo{person}{Daniel Cohen-Or}.}
  \bibinfo{year}{2023}\natexlab{}.
\newblock \showarticletitle{Encoder-based Domain Tuning for Fast
  Personalization of Text-to-Image Models}.
\newblock \bibinfo{journal}{\emph{ACM Transactions on Graphics (TOG)}}
  \bibinfo{volume}{42} (\bibinfo{year}{2023}), \bibinfo{pages}{1 -- 13}.
\newblock
\urldef\tempurl%
\url{https://api.semanticscholar.org/CorpusID:257364757}
\showURL{%
\tempurl}


\bibitem[Goodfellow et~al\mbox{.}(2014)]%
        {goodfellow2014generative}
\bibfield{author}{\bibinfo{person}{Ian Goodfellow}, \bibinfo{person}{Jean
  Pouget-Abadie}, \bibinfo{person}{Mehdi Mirza}, \bibinfo{person}{Bing Xu},
  \bibinfo{person}{David Warde-Farley}, \bibinfo{person}{Sherjil Ozair},
  \bibinfo{person}{Aaron Courville}, {and} \bibinfo{person}{Yoshua Bengio}.}
  \bibinfo{year}{2014}\natexlab{}.
\newblock \showarticletitle{Generative adversarial nets}.
\newblock \bibinfo{journal}{\emph{Advances in neural information processing
  systems}}  \bibinfo{volume}{27} (\bibinfo{year}{2014}).
\newblock


\bibitem[Han et~al\mbox{.}(2023)]%
        {Han2023SVDiffCP}
\bibfield{author}{\bibinfo{person}{Ligong Han}, \bibinfo{person}{Yinxiao Li},
  \bibinfo{person}{Han Zhang}, \bibinfo{person}{Peyman Milanfar},
  \bibinfo{person}{Dimitris~N. Metaxas}, {and} \bibinfo{person}{Feng Yang}.}
  \bibinfo{year}{2023}\natexlab{}.
\newblock \showarticletitle{SVDiff: Compact Parameter Space for Diffusion
  Fine-Tuning}.
\newblock \bibinfo{journal}{\emph{ArXiv}}  \bibinfo{volume}{abs/2303.11305}
  (\bibinfo{year}{2023}).
\newblock


\bibitem[Hertz et~al\mbox{.}(2022)]%
        {Hertz2022PrompttoPromptIE}
\bibfield{author}{\bibinfo{person}{Amir Hertz}, \bibinfo{person}{Ron Mokady},
  \bibinfo{person}{Jay Tenenbaum}, \bibinfo{person}{Kfir Aberman},
  \bibinfo{person}{Yael Pritch}, {and} \bibinfo{person}{Daniel Cohen-or}.}
  \bibinfo{year}{2022}\natexlab{}.
\newblock \showarticletitle{Prompt-to-Prompt Image Editing with Cross-Attention
  Control}. In \bibinfo{booktitle}{\emph{The Eleventh International Conference
  on Learning Representations}}.
\newblock


\bibitem[Ho et~al\mbox{.}(2022)]%
        {Ho2022ImagenVH}
\bibfield{author}{\bibinfo{person}{Jonathan Ho}, \bibinfo{person}{William
  Chan}, \bibinfo{person}{Chitwan Saharia}, \bibinfo{person}{Jay Whang},
  \bibinfo{person}{Ruiqi Gao}, \bibinfo{person}{Alexey~A. Gritsenko},
  \bibinfo{person}{Diederik~P. Kingma}, \bibinfo{person}{Ben Poole},
  \bibinfo{person}{Mohammad Norouzi}, \bibinfo{person}{David~J. Fleet}, {and}
  \bibinfo{person}{Tim Salimans}.} \bibinfo{year}{2022}\natexlab{}.
\newblock \showarticletitle{Imagen Video: High Definition Video Generation with
  Diffusion Models}.
\newblock \bibinfo{journal}{\emph{ArXiv}}  \bibinfo{volume}{abs/2210.02303}
  (\bibinfo{year}{2022}).
\newblock


\bibitem[Ho et~al\mbox{.}(2020)]%
        {ho2020denoising}
\bibfield{author}{\bibinfo{person}{Jonathan Ho}, \bibinfo{person}{Ajay Jain},
  {and} \bibinfo{person}{Pieter Abbeel}.} \bibinfo{year}{2020}\natexlab{}.
\newblock \showarticletitle{Denoising Diffusion Probabilistic Models}. In
  \bibinfo{booktitle}{\emph{Proc.~NeurIPS}}.
\newblock


\bibitem[Horwitz and Hoshen(2022)]%
        {Horwitz2022ConffusionCI}
\bibfield{author}{\bibinfo{person}{Eliahu Horwitz} {and} \bibinfo{person}{Yedid
  Hoshen}.} \bibinfo{year}{2022}\natexlab{}.
\newblock \showarticletitle{Conffusion: Confidence Intervals for Diffusion
  Models}.
\newblock \bibinfo{journal}{\emph{ArXiv}}  \bibinfo{volume}{abs/2211.09795}
  (\bibinfo{year}{2022}).
\newblock


\bibitem[Hu et~al\mbox{.}(2021)]%
        {lora}
\bibfield{author}{\bibinfo{person}{Edward~J Hu}, \bibinfo{person}{Phillip
  Wallis}, \bibinfo{person}{Zeyuan Allen-Zhu}, \bibinfo{person}{Yuanzhi Li},
  \bibinfo{person}{Shean Wang}, \bibinfo{person}{Lu Wang},
  \bibinfo{person}{Weizhu Chen}, {et~al\mbox{.}}}
  \bibinfo{year}{2021}\natexlab{}.
\newblock \showarticletitle{LoRA: Low-Rank Adaptation of Large Language
  Models}. In \bibinfo{booktitle}{\emph{International Conference on Learning
  Representations}}.
\newblock


\bibitem[Iluz et~al\mbox{.}(2023)]%
        {Iluz2023WordAsImageFS}
\bibfield{author}{\bibinfo{person}{Shira Iluz}, \bibinfo{person}{Yael Vinker},
  \bibinfo{person}{Amir Hertz}, \bibinfo{person}{Daniel Berio},
  \bibinfo{person}{Daniel Cohen-Or}, {and} \bibinfo{person}{Ariel Shamir}.}
  \bibinfo{year}{2023}\natexlab{}.
\newblock \showarticletitle{Word-As-Image for Semantic Typography}.
\newblock \bibinfo{journal}{\emph{ACM Transactions on Graphics (TOG)}}
  \bibinfo{volume}{42} (\bibinfo{year}{2023}), \bibinfo{pages}{1 -- 11}.
\newblock
\urldef\tempurl%
\url{https://api.semanticscholar.org/CorpusID:257353586}
\showURL{%
\tempurl}


\bibitem[Jia et~al\mbox{.}(2023)]%
        {Jia2023TamingEF}
\bibfield{author}{\bibinfo{person}{Xuhui Jia}, \bibinfo{person}{Yang Zhao},
  \bibinfo{person}{Kelvin C.~K. Chan}, \bibinfo{person}{Yandong Li},
  \bibinfo{person}{Han-Ying Zhang}, \bibinfo{person}{Boqing Gong},
  \bibinfo{person}{Tingbo Hou}, \bibinfo{person}{H. Wang}, {and}
  \bibinfo{person}{Yu-Chuan Su}.} \bibinfo{year}{2023}\natexlab{}.
\newblock \showarticletitle{Taming Encoder for Zero Fine-tuning Image
  Customization with Text-to-Image Diffusion Models}.
\newblock \bibinfo{journal}{\emph{ArXiv}}  \bibinfo{volume}{abs/2304.02642}
  (\bibinfo{year}{2023}).
\newblock


\bibitem[Karras et~al\mbox{.}(2019)]%
        {karras2019style}
\bibfield{author}{\bibinfo{person}{Tero Karras}, \bibinfo{person}{Samuli
  Laine}, {and} \bibinfo{person}{Timo Aila}.} \bibinfo{year}{2019}\natexlab{}.
\newblock \showarticletitle{A style-based generator architecture for generative
  adversarial networks}. In \bibinfo{booktitle}{\emph{Proceedings of the IEEE
  conference on computer vision and pattern recognition}}.
  \bibinfo{pages}{4401--4410}.
\newblock


\bibitem[Karras et~al\mbox{.}(2020)]%
        {karras2020analyzing}
\bibfield{author}{\bibinfo{person}{Tero Karras}, \bibinfo{person}{Samuli
  Laine}, \bibinfo{person}{Miika Aittala}, \bibinfo{person}{Janne Hellsten},
  \bibinfo{person}{Jaakko Lehtinen}, {and} \bibinfo{person}{Timo Aila}.}
  \bibinfo{year}{2020}\natexlab{}.
\newblock \showarticletitle{Analyzing and improving the image quality of
  stylegan}. In \bibinfo{booktitle}{\emph{Proceedings of the IEEE/CVF
  Conference on Computer Vision and Pattern Recognition}}.
  \bibinfo{pages}{8110--8119}.
\newblock


\bibitem[Kawar et~al\mbox{.}(2023)]%
        {Kawar2022ImagicTR}
\bibfield{author}{\bibinfo{person}{Bahjat Kawar}, \bibinfo{person}{Shiran
  Zada}, \bibinfo{person}{Oran Lang}, \bibinfo{person}{Omer Tov},
  \bibinfo{person}{Huiwen Chang}, \bibinfo{person}{Tali Dekel},
  \bibinfo{person}{Inbar Mosseri}, {and} \bibinfo{person}{Michal Irani}.}
  \bibinfo{year}{2023}\natexlab{}.
\newblock \showarticletitle{Imagic: Text-based real image editing with
  diffusion models}. In \bibinfo{booktitle}{\emph{Proceedings of the IEEE/CVF
  Conference on Computer Vision and Pattern Recognition}}.
  \bibinfo{pages}{6007--6017}.
\newblock


\bibitem[Kingma and Ba(2014)]%
        {Kingma2014AdamAM}
\bibfield{author}{\bibinfo{person}{Diederik~P. Kingma} {and}
  \bibinfo{person}{Jimmy Ba}.} \bibinfo{year}{2014}\natexlab{}.
\newblock \showarticletitle{Adam: A Method for Stochastic Optimization}.
\newblock \bibinfo{journal}{\emph{CoRR}}  \bibinfo{volume}{abs/1412.6980}
  (\bibinfo{year}{2014}).
\newblock


\bibitem[Kirillov et~al\mbox{.}(2023)]%
        {kirillov2023segment}
\bibfield{author}{\bibinfo{person}{Alexander Kirillov}, \bibinfo{person}{Eric
  Mintun}, \bibinfo{person}{Nikhila Ravi}, \bibinfo{person}{Hanzi Mao},
  \bibinfo{person}{Chloe Rolland}, \bibinfo{person}{Laura Gustafson},
  \bibinfo{person}{Tete Xiao}, \bibinfo{person}{Spencer Whitehead},
  \bibinfo{person}{Alexander~C. Berg}, \bibinfo{person}{Wan-Yen Lo},
  \bibinfo{person}{Piotr Doll\'{a}r}, {and} \bibinfo{person}{Ross Girshick}.}
  \bibinfo{year}{2023}\natexlab{}.
\newblock \bibinfo{title}{Segment Anything}.
\newblock
\newblock
\showeprint[arxiv]{2304.02643}~[cs.CV]


\bibitem[Kruskal and Wallis(1952)]%
        {Kruskal1952UseOR}
\bibfield{author}{\bibinfo{person}{William~H. Kruskal} {and}
  \bibinfo{person}{Wilson~Allen Wallis}.} \bibinfo{year}{1952}\natexlab{}.
\newblock \showarticletitle{Use of Ranks in One-Criterion Variance Analysis}.
\newblock \bibinfo{journal}{\emph{J. Amer. Statist. Assoc.}}
  \bibinfo{volume}{47} (\bibinfo{year}{1952}), \bibinfo{pages}{583--621}.
\newblock


\bibitem[Kumari et~al\mbox{.}(2023)]%
        {Kumari2022MultiConceptCO}
\bibfield{author}{\bibinfo{person}{Nupur Kumari}, \bibinfo{person}{Bingliang
  Zhang}, \bibinfo{person}{Richard Zhang}, \bibinfo{person}{Eli Shechtman},
  {and} \bibinfo{person}{Jun-Yan Zhu}.} \bibinfo{year}{2023}\natexlab{}.
\newblock \showarticletitle{Multi-concept customization of text-to-image
  diffusion}. In \bibinfo{booktitle}{\emph{Proceedings of the IEEE/CVF
  Conference on Computer Vision and Pattern Recognition}}.
  \bibinfo{pages}{1931--1941}.
\newblock


\bibitem[Kwon and Ye(2022)]%
        {kwon2021clipstyler}
\bibfield{author}{\bibinfo{person}{Gihyun Kwon} {and}
  \bibinfo{person}{Jong~Chul Ye}.} \bibinfo{year}{2022}\natexlab{}.
\newblock \showarticletitle{Clipstyler: Image style transfer with a single text
  condition}. In \bibinfo{booktitle}{\emph{Proceedings of the IEEE/CVF
  Conference on Computer Vision and Pattern Recognition}}.
  \bibinfo{pages}{18062--18071}.
\newblock


\bibitem[Lin et~al\mbox{.}(2014)]%
        {Lin2014MicrosoftCC}
\bibfield{author}{\bibinfo{person}{Tsung-Yi Lin}, \bibinfo{person}{Michael
  Maire}, \bibinfo{person}{Serge~J. Belongie}, \bibinfo{person}{James Hays},
  \bibinfo{person}{Pietro Perona}, \bibinfo{person}{Deva Ramanan},
  \bibinfo{person}{Piotr Doll{\'a}r}, {and} \bibinfo{person}{C.~Lawrence
  Zitnick}.} \bibinfo{year}{2014}\natexlab{}.
\newblock \showarticletitle{Microsoft COCO: Common Objects in Context}. In
  \bibinfo{booktitle}{\emph{European Conference on Computer Vision}}.
\newblock


\bibitem[Liu et~al\mbox{.}(2021)]%
        {Liu2021SwinTH}
\bibfield{author}{\bibinfo{person}{Ze Liu}, \bibinfo{person}{Yutong Lin},
  \bibinfo{person}{Yue Cao}, \bibinfo{person}{Han Hu}, \bibinfo{person}{Yixuan
  Wei}, \bibinfo{person}{Zheng Zhang}, \bibinfo{person}{Stephen Lin}, {and}
  \bibinfo{person}{Baining Guo}.} \bibinfo{year}{2021}\natexlab{}.
\newblock \showarticletitle{Swin Transformer: Hierarchical Vision Transformer
  using Shifted Windows}.
\newblock \bibinfo{journal}{\emph{2021 IEEE/CVF International Conference on
  Computer Vision (ICCV)}} (\bibinfo{year}{2021}),
  \bibinfo{pages}{9992--10002}.
\newblock


\bibitem[Mansimov et~al\mbox{.}(2016)]%
        {Mansimov2016GeneratingIF}
\bibfield{author}{\bibinfo{person}{Elman Mansimov}, \bibinfo{person}{Emilio
  Parisotto}, \bibinfo{person}{Jimmy Ba}, {and} \bibinfo{person}{Ruslan
  Salakhutdinov}.} \bibinfo{year}{2016}\natexlab{}.
\newblock \showarticletitle{Generating Images from Captions with Attention}.
\newblock \bibinfo{journal}{\emph{CoRR}}  \bibinfo{volume}{abs/1511.02793}
  (\bibinfo{year}{2016}).
\newblock


\bibitem[Meng et~al\mbox{.}(2021)]%
        {meng2021sdedit}
\bibfield{author}{\bibinfo{person}{Chenlin Meng}, \bibinfo{person}{Yutong He},
  \bibinfo{person}{Yang Song}, \bibinfo{person}{Jiaming Song},
  \bibinfo{person}{Jiajun Wu}, \bibinfo{person}{Jun-Yan Zhu}, {and}
  \bibinfo{person}{Stefano Ermon}.} \bibinfo{year}{2021}\natexlab{}.
\newblock \showarticletitle{SDEdit: Guided Image Synthesis and Editing with
  Stochastic Differential Equations}. In
  \bibinfo{booktitle}{\emph{International Conference on Learning
  Representations}}.
\newblock


\bibitem[Mokady et~al\mbox{.}(2023)]%
        {mokady2022null}
\bibfield{author}{\bibinfo{person}{Ron Mokady}, \bibinfo{person}{Amir Hertz},
  \bibinfo{person}{Kfir Aberman}, \bibinfo{person}{Yael Pritch}, {and}
  \bibinfo{person}{Daniel Cohen-Or}.} \bibinfo{year}{2023}\natexlab{}.
\newblock \showarticletitle{Null-text inversion for editing real images using
  guided diffusion models}. In \bibinfo{booktitle}{\emph{Proceedings of the
  IEEE/CVF Conference on Computer Vision and Pattern Recognition}}.
  \bibinfo{pages}{6038--6047}.
\newblock


\bibitem[Molad et~al\mbox{.}(2023)]%
        {Molad2023DreamixVD}
\bibfield{author}{\bibinfo{person}{Eyal Molad}, \bibinfo{person}{Eliahu
  Horwitz}, \bibinfo{person}{Dani Valevski}, \bibinfo{person}{Alex~Rav Acha},
  \bibinfo{person}{Y. Matias}, \bibinfo{person}{Yael Pritch},
  \bibinfo{person}{Yaniv Leviathan}, {and} \bibinfo{person}{Yedid Hoshen}.}
  \bibinfo{year}{2023}\natexlab{}.
\newblock \showarticletitle{Dreamix: Video Diffusion Models are General Video
  Editors}.
\newblock \bibinfo{journal}{\emph{ArXiv}}  \bibinfo{volume}{abs/2302.01329}
  (\bibinfo{year}{2023}).
\newblock


\bibitem[Nichol et~al\mbox{.}(2021)]%
        {nichol2021glide}
\bibfield{author}{\bibinfo{person}{Alex Nichol}, \bibinfo{person}{Prafulla
  Dhariwal}, \bibinfo{person}{Aditya Ramesh}, \bibinfo{person}{Pranav Shyam},
  \bibinfo{person}{Pamela Mishkin}, \bibinfo{person}{Bob McGrew},
  \bibinfo{person}{Ilya Sutskever}, {and} \bibinfo{person}{Mark Chen}.}
  \bibinfo{year}{2021}\natexlab{}.
\newblock \showarticletitle{GLIDE: Towards Photorealistic Image Generation and
  Editing with Text-Guided Diffusion Models}. In
  \bibinfo{booktitle}{\emph{International Conference on Machine Learning}}.
\newblock
\urldef\tempurl%
\url{https://api.semanticscholar.org/CorpusID:245335086}
\showURL{%
\tempurl}


\bibitem[Nitzan et~al\mbox{.}(2022)]%
        {nitzan2022mystyle}
\bibfield{author}{\bibinfo{person}{Yotam Nitzan}, \bibinfo{person}{Kfir
  Aberman}, \bibinfo{person}{Qiurui He}, \bibinfo{person}{Orly Liba},
  \bibinfo{person}{Michal Yarom}, \bibinfo{person}{Yossi Gandelsman},
  \bibinfo{person}{Inbar Mosseri}, \bibinfo{person}{Yael Pritch}, {and}
  \bibinfo{person}{Daniel Cohen-Or}.} \bibinfo{year}{2022}\natexlab{}.
\newblock \showarticletitle{Mystyle: A personalized generative prior}.
\newblock \bibinfo{journal}{\emph{ACM Transactions on Graphics (TOG)}}
  \bibinfo{volume}{41}, \bibinfo{number}{6} (\bibinfo{year}{2022}),
  \bibinfo{pages}{1--10}.
\newblock


\bibitem[Patashnik et~al\mbox{.}(2023)]%
        {Patashnik2023LocalizingOS}
\bibfield{author}{\bibinfo{person}{Or Patashnik}, \bibinfo{person}{Daniel
  Garibi}, \bibinfo{person}{Idan Azuri}, \bibinfo{person}{Hadar Averbuch-Elor},
  {and} \bibinfo{person}{Daniel Cohen-Or}.} \bibinfo{year}{2023}\natexlab{}.
\newblock \showarticletitle{Localizing Object-level Shape Variations with
  Text-to-Image Diffusion Models}.
\newblock \bibinfo{journal}{\emph{ArXiv}}  \bibinfo{volume}{abs/2303.11306}
  (\bibinfo{year}{2023}).
\newblock


\bibitem[Patashnik et~al\mbox{.}(2021)]%
        {patashnik2021styleclip}
\bibfield{author}{\bibinfo{person}{Or Patashnik}, \bibinfo{person}{Zongze Wu},
  \bibinfo{person}{Eli Shechtman}, \bibinfo{person}{Daniel Cohen-Or}, {and}
  \bibinfo{person}{Dani Lischinski}.} \bibinfo{year}{2021}\natexlab{}.
\newblock \showarticletitle{StyleCLIP: Text-Driven Manipulation of StyleGAN
  Imagery}.
\newblock \bibinfo{journal}{\emph{2021 IEEE/CVF International Conference on
  Computer Vision (ICCV)}} (\bibinfo{year}{2021}), \bibinfo{pages}{2065--2074}.
\newblock
\urldef\tempurl%
\url{https://api.semanticscholar.org/CorpusID:232428282}
\showURL{%
\tempurl}


\bibitem[Pidhorskyi et~al\mbox{.}(2020)]%
        {Pidhorskyi2020AdversarialLA}
\bibfield{author}{\bibinfo{person}{Stanislav Pidhorskyi},
  \bibinfo{person}{Donald~A. Adjeroh}, {and} \bibinfo{person}{Gianfranco
  Doretto}.} \bibinfo{year}{2020}\natexlab{}.
\newblock \showarticletitle{Adversarial Latent Autoencoders}.
\newblock \bibinfo{journal}{\emph{2020 IEEE/CVF Conference on Computer Vision
  and Pattern Recognition (CVPR)}} (\bibinfo{year}{2020}),
  \bibinfo{pages}{14092--14101}.
\newblock


\bibitem[Radford et~al\mbox{.}(2021)]%
        {Radford2021LearningTV}
\bibfield{author}{\bibinfo{person}{Alec Radford}, \bibinfo{person}{Jong~Wook
  Kim}, \bibinfo{person}{Chris Hallacy}, \bibinfo{person}{Aditya Ramesh},
  \bibinfo{person}{Gabriel Goh}, \bibinfo{person}{Sandhini Agarwal},
  \bibinfo{person}{Girish Sastry}, \bibinfo{person}{Amanda Askell},
  \bibinfo{person}{Pamela Mishkin}, \bibinfo{person}{Jack Clark},
  \bibinfo{person}{Gretchen Krueger}, {and} \bibinfo{person}{Ilya Sutskever}.}
  \bibinfo{year}{2021}\natexlab{}.
\newblock \showarticletitle{Learning Transferable Visual Models From Natural
  Language Supervision}. In \bibinfo{booktitle}{\emph{International Conference
  on Machine Learning}}.
\newblock


\bibitem[Ramesh et~al\mbox{.}(2022)]%
        {ramesh2022hierarchical}
\bibfield{author}{\bibinfo{person}{Aditya Ramesh}, \bibinfo{person}{Prafulla
  Dhariwal}, \bibinfo{person}{Alex Nichol}, \bibinfo{person}{Casey Chu}, {and}
  \bibinfo{person}{Mark Chen}.} \bibinfo{year}{2022}\natexlab{}.
\newblock \showarticletitle{Hierarchical text-conditional image generation with
  clip latents}.
\newblock \bibinfo{journal}{\emph{arXiv preprint arXiv:2204.06125}}
  (\bibinfo{year}{2022}).
\newblock


\bibitem[Ramesh et~al\mbox{.}(2021)]%
        {Ramesh2021ZeroShotTG}
\bibfield{author}{\bibinfo{person}{Aditya Ramesh}, \bibinfo{person}{Mikhail
  Pavlov}, \bibinfo{person}{Gabriel Goh}, \bibinfo{person}{Scott Gray},
  \bibinfo{person}{Chelsea Voss}, \bibinfo{person}{Alec Radford},
  \bibinfo{person}{Mark Chen}, {and} \bibinfo{person}{Ilya Sutskever}.}
  \bibinfo{year}{2021}\natexlab{}.
\newblock \showarticletitle{Zero-shot text-to-image generation}. In
  \bibinfo{booktitle}{\emph{International Conference on Machine Learning}}.
  PMLR, \bibinfo{pages}{8821--8831}.
\newblock


\bibitem[Reed et~al\mbox{.}(2016)]%
        {reed2016generative}
\bibfield{author}{\bibinfo{person}{Scott Reed}, \bibinfo{person}{Zeynep Akata},
  \bibinfo{person}{Xinchen Yan}, \bibinfo{person}{Lajanugen Logeswaran},
  \bibinfo{person}{Bernt Schiele}, {and} \bibinfo{person}{Honglak Lee}.}
  \bibinfo{year}{2016}\natexlab{}.
\newblock \showarticletitle{Generative adversarial text to image synthesis}. In
  \bibinfo{booktitle}{\emph{Proc.~ICLR}}. \bibinfo{pages}{1060--1069}.
\newblock


\bibitem[Richardson et~al\mbox{.}(2020)]%
        {Richardson2020EncodingIS}
\bibfield{author}{\bibinfo{person}{Elad Richardson}, \bibinfo{person}{Yuval
  Alaluf}, \bibinfo{person}{Or Patashnik}, \bibinfo{person}{Yotam Nitzan},
  \bibinfo{person}{Yaniv Azar}, \bibinfo{person}{Stav Shapiro}, {and}
  \bibinfo{person}{Daniel Cohen-Or}.} \bibinfo{year}{2020}\natexlab{}.
\newblock \showarticletitle{Encoding in Style: a StyleGAN Encoder for
  Image-to-Image Translation}.
\newblock \bibinfo{journal}{\emph{2021 IEEE/CVF Conference on Computer Vision
  and Pattern Recognition (CVPR)}} (\bibinfo{year}{2020}),
  \bibinfo{pages}{2287--2296}.
\newblock


\bibitem[Richardson et~al\mbox{.}(2023)]%
        {Richardson2023TEXTureTT}
\bibfield{author}{\bibinfo{person}{Elad Richardson}, \bibinfo{person}{Gal
  Metzer}, \bibinfo{person}{Yuval Alaluf}, \bibinfo{person}{Raja Giryes}, {and}
  \bibinfo{person}{Daniel Cohen-Or}.} \bibinfo{year}{2023}\natexlab{}.
\newblock \showarticletitle{TEXTure: Text-Guided Texturing of 3D Shapes}.
\newblock \bibinfo{journal}{\emph{ACM SIGGRAPH 2023 Conference Proceedings}}
  (\bibinfo{year}{2023}).
\newblock
\urldef\tempurl%
\url{https://api.semanticscholar.org/CorpusID:256597953}
\showURL{%
\tempurl}


\bibitem[Roich et~al\mbox{.}(2021)]%
        {Roich2021PivotalTF}
\bibfield{author}{\bibinfo{person}{Daniel Roich}, \bibinfo{person}{Ron Mokady},
  \bibinfo{person}{Amit~H. Bermano}, {and} \bibinfo{person}{Daniel Cohen-Or}.}
  \bibinfo{year}{2021}\natexlab{}.
\newblock \showarticletitle{Pivotal Tuning for Latent-based Editing of Real
  Images}.
\newblock \bibinfo{journal}{\emph{ACM Transactions on Graphics (TOG)}}
  \bibinfo{volume}{42} (\bibinfo{year}{2021}), \bibinfo{pages}{1 -- 13}.
\newblock


\bibitem[Rombach et~al\mbox{.}(2021)]%
        {Rombach2021HighResolutionIS}
\bibfield{author}{\bibinfo{person}{Robin Rombach}, \bibinfo{person}{A.
  Blattmann}, \bibinfo{person}{Dominik Lorenz}, \bibinfo{person}{Patrick
  Esser}, {and} \bibinfo{person}{Bj{\"o}rn Ommer}.}
  \bibinfo{year}{2021}\natexlab{}.
\newblock \showarticletitle{High-Resolution Image Synthesis with Latent
  Diffusion Models}.
\newblock \bibinfo{journal}{\emph{2022 IEEE/CVF Conference on Computer Vision
  and Pattern Recognition (CVPR)}} (\bibinfo{year}{2021}),
  \bibinfo{pages}{10674--10685}.
\newblock


\bibitem[Ruiz et~al\mbox{.}(2023)]%
        {Ruiz2022DreamBoothFT}
\bibfield{author}{\bibinfo{person}{Nataniel Ruiz}, \bibinfo{person}{Yuanzhen
  Li}, \bibinfo{person}{Varun Jampani}, \bibinfo{person}{Yael Pritch},
  \bibinfo{person}{Michael Rubinstein}, {and} \bibinfo{person}{Kfir Aberman}.}
  \bibinfo{year}{2023}\natexlab{}.
\newblock \showarticletitle{Dreambooth: Fine tuning text-to-image diffusion
  models for subject-driven generation}. In
  \bibinfo{booktitle}{\emph{Proceedings of the IEEE/CVF Conference on Computer
  Vision and Pattern Recognition}}. \bibinfo{pages}{22500--22510}.
\newblock


\bibitem[Ryu(2022)]%
        {lora_diffusion}
\bibfield{author}{\bibinfo{person}{Simo Ryu}.} \bibinfo{year}{2022}\natexlab{}.
\newblock \bibinfo{title}{Low-rank Adaptation for Fast Text-to-Image Diffusion
  Fine-tuning}.
\newblock \bibinfo{howpublished}{\url{https://github.com/cloneofsimo/lora}}.
\newblock


\bibitem[Saharia et~al\mbox{.}(2021a)]%
        {Saharia2021PaletteID}
\bibfield{author}{\bibinfo{person}{Chitwan Saharia}, \bibinfo{person}{William
  Chan}, \bibinfo{person}{Huiwen Chang}, \bibinfo{person}{Chris~A. Lee},
  \bibinfo{person}{Jonathan Ho}, \bibinfo{person}{Tim Salimans},
  \bibinfo{person}{David~J. Fleet}, {and} \bibinfo{person}{Mohammad Norouzi}.}
  \bibinfo{year}{2021}\natexlab{a}.
\newblock \showarticletitle{Palette: Image-to-Image Diffusion Models}.
\newblock \bibinfo{journal}{\emph{ACM SIGGRAPH 2022 Conference Proceedings}}
  (\bibinfo{year}{2021}).
\newblock


\bibitem[Saharia et~al\mbox{.}(2022)]%
        {Saharia2022PhotorealisticTD}
\bibfield{author}{\bibinfo{person}{Chitwan Saharia}, \bibinfo{person}{William
  Chan}, \bibinfo{person}{Saurabh Saxena}, \bibinfo{person}{Lala Li},
  \bibinfo{person}{Jay Whang}, \bibinfo{person}{Emily~L Denton},
  \bibinfo{person}{Kamyar Ghasemipour}, \bibinfo{person}{Raphael
  Gontijo~Lopes}, \bibinfo{person}{Burcu Karagol~Ayan}, \bibinfo{person}{Tim
  Salimans}, {et~al\mbox{.}}} \bibinfo{year}{2022}\natexlab{}.
\newblock \showarticletitle{Photorealistic text-to-image diffusion models with
  deep language understanding}.
\newblock \bibinfo{journal}{\emph{Advances in Neural Information Processing
  Systems}}  \bibinfo{volume}{35} (\bibinfo{year}{2022}),
  \bibinfo{pages}{36479--36494}.
\newblock


\bibitem[Saharia et~al\mbox{.}(2021b)]%
        {Saharia2021ImageSV}
\bibfield{author}{\bibinfo{person}{Chitwan Saharia}, \bibinfo{person}{Jonathan
  Ho}, \bibinfo{person}{William Chan}, \bibinfo{person}{Tim Salimans},
  \bibinfo{person}{David~J. Fleet}, {and} \bibinfo{person}{Mohammad Norouzi}.}
  \bibinfo{year}{2021}\natexlab{b}.
\newblock \showarticletitle{Image Super-Resolution via Iterative Refinement}.
\newblock \bibinfo{journal}{\emph{IEEE Transactions on Pattern Analysis and
  Machine Intelligence}}  \bibinfo{volume}{45} (\bibinfo{year}{2021}),
  \bibinfo{pages}{4713--4726}.
\newblock


\bibitem[Sheynin et~al\mbox{.}(2022)]%
        {sheynin2022knn}
\bibfield{author}{\bibinfo{person}{Shelly Sheynin}, \bibinfo{person}{Oron
  Ashual}, \bibinfo{person}{Adam Polyak}, \bibinfo{person}{Uriel Singer},
  \bibinfo{person}{Oran Gafni}, \bibinfo{person}{Eliya Nachmani}, {and}
  \bibinfo{person}{Yaniv Taigman}.} \bibinfo{year}{2022}\natexlab{}.
\newblock \showarticletitle{kNN-Diffusion: Image Generation via Large-Scale
  Retrieval}. In \bibinfo{booktitle}{\emph{The Eleventh International
  Conference on Learning Representations}}.
\newblock


\bibitem[Shi et~al\mbox{.}(2023)]%
        {Shi2023InstantBoothPT}
\bibfield{author}{\bibinfo{person}{Jing Shi}, \bibinfo{person}{Wei Xiong},
  \bibinfo{person}{Zhe~L. Lin}, {and} \bibinfo{person}{Hyun~Joon Jung}.}
  \bibinfo{year}{2023}\natexlab{}.
\newblock \showarticletitle{InstantBooth: Personalized Text-to-Image Generation
  without Test-Time Finetuning}.
\newblock \bibinfo{journal}{\emph{ArXiv}}  \bibinfo{volume}{abs/2304.03411}
  (\bibinfo{year}{2023}).
\newblock


\bibitem[Singer et~al\mbox{.}(2022)]%
        {Singer2022MakeAVideoTG}
\bibfield{author}{\bibinfo{person}{Uriel Singer}, \bibinfo{person}{Adam
  Polyak}, \bibinfo{person}{Thomas Hayes}, \bibinfo{person}{Xi Yin},
  \bibinfo{person}{Jie An}, \bibinfo{person}{Songyang Zhang},
  \bibinfo{person}{Qiyuan Hu}, \bibinfo{person}{Harry Yang},
  \bibinfo{person}{Oron Ashual}, \bibinfo{person}{Oran Gafni}, {et~al\mbox{.}}}
  \bibinfo{year}{2022}\natexlab{}.
\newblock \showarticletitle{Make-A-Video: Text-to-Video Generation without
  Text-Video Data}. In \bibinfo{booktitle}{\emph{The Eleventh International
  Conference on Learning Representations}}.
\newblock


\bibitem[Sohl-Dickstein et~al\mbox{.}(2015)]%
        {sohl2015deep}
\bibfield{author}{\bibinfo{person}{Jascha Sohl-Dickstein},
  \bibinfo{person}{Eric Weiss}, \bibinfo{person}{Niru Maheswaranathan}, {and}
  \bibinfo{person}{Surya Ganguli}.} \bibinfo{year}{2015}\natexlab{}.
\newblock \showarticletitle{Deep unsupervised learning using nonequilibrium
  thermodynamics}. In \bibinfo{booktitle}{\emph{International Conference on
  Machine Learning}}. PMLR, \bibinfo{pages}{2256--2265}.
\newblock


\bibitem[Song et~al\mbox{.}(2020)]%
        {song2020denoising}
\bibfield{author}{\bibinfo{person}{Jiaming Song}, \bibinfo{person}{Chenlin
  Meng}, {and} \bibinfo{person}{Stefano Ermon}.}
  \bibinfo{year}{2020}\natexlab{}.
\newblock \showarticletitle{Denoising Diffusion Implicit Models}. In
  \bibinfo{booktitle}{\emph{International Conference on Learning
  Representations}}.
\newblock


\bibitem[Song and Ermon(2019)]%
        {song2019generative}
\bibfield{author}{\bibinfo{person}{Yang Song} {and} \bibinfo{person}{Stefano
  Ermon}.} \bibinfo{year}{2019}\natexlab{}.
\newblock \showarticletitle{Generative modeling by estimating gradients of the
  data distribution}.
\newblock \bibinfo{journal}{\emph{Advances in Neural Information Processing
  Systems}}  \bibinfo{volume}{32} (\bibinfo{year}{2019}).
\newblock


\bibitem[Song et~al\mbox{.}(2022)]%
        {Song2022ObjectStitchGO}
\bibfield{author}{\bibinfo{person}{Yi-Zhe Song}, \bibinfo{person}{Zhifei
  Zhang}, \bibinfo{person}{Zhe~L. Lin}, \bibinfo{person}{Scott~D. Cohen},
  \bibinfo{person}{Brian~L. Price}, \bibinfo{person}{Jianming Zhang},
  \bibinfo{person}{Soo~Ye Kim}, {and} \bibinfo{person}{Daniel~G. Aliaga}.}
  \bibinfo{year}{2022}\natexlab{}.
\newblock \showarticletitle{ObjectStitch: Generative Object Compositing}.
\newblock \bibinfo{journal}{\emph{ArXiv}}  \bibinfo{volume}{abs/2212.00932}
  (\bibinfo{year}{2022}).
\newblock


\bibitem[Tewel et~al\mbox{.}(2023)]%
        {Tewel2023KeyLockedRO}
\bibfield{author}{\bibinfo{person}{Yoad Tewel}, \bibinfo{person}{Rinon Gal},
  \bibinfo{person}{Gal Chechik}, {and} \bibinfo{person}{Yuval Atzmon}.}
  \bibinfo{year}{2023}\natexlab{}.
\newblock \showarticletitle{Key-Locked Rank One Editing for Text-to-Image
  Personalization}.
\newblock \bibinfo{journal}{\emph{ACM SIGGRAPH 2023 Conference Proceedings}}
  (\bibinfo{year}{2023}).
\newblock
\urldef\tempurl%
\url{https://api.semanticscholar.org/CorpusID:258436985}
\showURL{%
\tempurl}


\bibitem[Tov et~al\mbox{.}(2021)]%
        {Tov2021DesigningAE}
\bibfield{author}{\bibinfo{person}{Omer Tov}, \bibinfo{person}{Yuval Alaluf},
  \bibinfo{person}{Yotam Nitzan}, \bibinfo{person}{Or Patashnik}, {and}
  \bibinfo{person}{Daniel Cohen-Or}.} \bibinfo{year}{2021}\natexlab{}.
\newblock \showarticletitle{Designing an encoder for StyleGAN image
  manipulation}.
\newblock \bibinfo{journal}{\emph{ACM Transactions on Graphics (TOG)}}
  \bibinfo{volume}{40} (\bibinfo{year}{2021}), \bibinfo{pages}{1 -- 14}.
\newblock


\bibitem[Tukey(1949)]%
        {Tukey1949ComparingIM}
\bibfield{author}{\bibinfo{person}{John~W. Tukey}.}
  \bibinfo{year}{1949}\natexlab{}.
\newblock \showarticletitle{Comparing individual means in the analysis of
  variance.}
\newblock \bibinfo{journal}{\emph{Biometrics}}  \bibinfo{volume}{5 2}
  (\bibinfo{year}{1949}), \bibinfo{pages}{99--114}.
\newblock


\bibitem[Tumanyan et~al\mbox{.}(2023)]%
        {pnpDiffusion2022}
\bibfield{author}{\bibinfo{person}{Narek Tumanyan}, \bibinfo{person}{Michal
  Geyer}, \bibinfo{person}{Shai Bagon}, {and} \bibinfo{person}{Tali Dekel}.}
  \bibinfo{year}{2023}\natexlab{}.
\newblock \showarticletitle{Plug-and-play diffusion features for text-driven
  image-to-image translation}. In \bibinfo{booktitle}{\emph{Proceedings of the
  IEEE/CVF Conference on Computer Vision and Pattern Recognition}}.
  \bibinfo{pages}{1921--1930}.
\newblock


\bibitem[Valevski et~al\mbox{.}(2022)]%
        {valevski2022unitune}
\bibfield{author}{\bibinfo{person}{Dani Valevski}, \bibinfo{person}{Matan
  Kalman}, \bibinfo{person}{Yossi Matias}, {and} \bibinfo{person}{Yaniv
  Leviathan}.} \bibinfo{year}{2022}\natexlab{}.
\newblock \showarticletitle{UniTune: Text-Driven Image Editing by Fine Tuning
  an Image Generation Model on a Single Image}.
\newblock \bibinfo{journal}{\emph{arXiv preprint arXiv:2210.09477}}
  (\bibinfo{year}{2022}).
\newblock


\bibitem[von Platen et~al\mbox{.}(2022)]%
        {von-platen-etal-2022-diffusers}
\bibfield{author}{\bibinfo{person}{Patrick von Platen}, \bibinfo{person}{Suraj
  Patil}, \bibinfo{person}{Anton Lozhkov}, \bibinfo{person}{Pedro Cuenca},
  \bibinfo{person}{Nathan Lambert}, \bibinfo{person}{Kashif Rasul},
  \bibinfo{person}{Mishig Davaadorj}, {and} \bibinfo{person}{Thomas Wolf}.}
  \bibinfo{year}{2022}\natexlab{}.
\newblock \bibinfo{title}{Diffusers: State-of-the-art diffusion models}.
\newblock
  \bibinfo{howpublished}{\url{https://github.com/huggingface/diffusers}}.
\newblock


\bibitem[Voynov et~al\mbox{.}(2023)]%
        {Voynov2023PET}
\bibfield{author}{\bibinfo{person}{Andrey Voynov}, \bibinfo{person}{Q. Chu},
  \bibinfo{person}{Daniel Cohen-Or}, {and} \bibinfo{person}{Kfir Aberman}.}
  \bibinfo{year}{2023}\natexlab{}.
\newblock \showarticletitle{P+: Extended Textual Conditioning in Text-to-Image
  Generation}.
\newblock \bibinfo{journal}{\emph{ArXiv}}  \bibinfo{volume}{abs/2303.09522}
  (\bibinfo{year}{2023}).
\newblock


\bibitem[Wang et~al\mbox{.}(2023)]%
        {wang2022imagen}
\bibfield{author}{\bibinfo{person}{Su Wang}, \bibinfo{person}{Chitwan Saharia},
  \bibinfo{person}{Ceslee Montgomery}, \bibinfo{person}{Jordi Pont-Tuset},
  \bibinfo{person}{Shai Noy}, \bibinfo{person}{Stefano Pellegrini},
  \bibinfo{person}{Yasumasa Onoe}, \bibinfo{person}{Sarah Laszlo},
  \bibinfo{person}{David~J Fleet}, \bibinfo{person}{Radu Soricut},
  {et~al\mbox{.}}} \bibinfo{year}{2023}\natexlab{}.
\newblock \showarticletitle{Imagen editor and editbench: Advancing and
  evaluating text-guided image inpainting}. In
  \bibinfo{booktitle}{\emph{Proceedings of the IEEE/CVF Conference on Computer
  Vision and Pattern Recognition}}. \bibinfo{pages}{18359--18369}.
\newblock


\bibitem[Wei(2023)]%
        {ELITE-implmentation}
\bibfield{author}{\bibinfo{person}{Yuxiang Wei}.}
  \bibinfo{year}{2023}\natexlab{}.
\newblock \bibinfo{title}{Official Implementation of ELITE}.
\newblock \bibinfo{howpublished}{\url{https://github.com/csyxwei/ELITE}}.
\newblock
\newblock
\shownote{Accessed: 2023-05-01}.


\bibitem[Wei et~al\mbox{.}(2023)]%
        {Wei2023ELITEEV}
\bibfield{author}{\bibinfo{person}{Yuxiang Wei}, \bibinfo{person}{Yabo Zhang},
  \bibinfo{person}{Zhilong Ji}, \bibinfo{person}{Jinfeng Bai},
  \bibinfo{person}{Lei Zhang}, {and} \bibinfo{person}{Wangmeng Zuo}.}
  \bibinfo{year}{2023}\natexlab{}.
\newblock \showarticletitle{ELITE: Encoding Visual Concepts into Textual
  Embeddings for Customized Text-to-Image Generation}.
\newblock \bibinfo{journal}{\emph{ArXiv}}  \bibinfo{volume}{abs/2302.13848}
  (\bibinfo{year}{2023}).
\newblock


\bibitem[Xia et~al\mbox{.}(2021)]%
        {Xia2021GANIA}
\bibfield{author}{\bibinfo{person}{Weihao Xia}, \bibinfo{person}{Yulun Zhang},
  \bibinfo{person}{Yujiu Yang}, \bibinfo{person}{Jing-Hao Xue},
  \bibinfo{person}{Bolei Zhou}, {and} \bibinfo{person}{Ming-Hsuan Yang}.}
  \bibinfo{year}{2021}\natexlab{}.
\newblock \showarticletitle{GAN Inversion: A Survey}.
\newblock \bibinfo{journal}{\emph{IEEE Transactions on Pattern Analysis and
  Machine Intelligence}}  \bibinfo{volume}{45} (\bibinfo{year}{2021}),
  \bibinfo{pages}{3121--3138}.
\newblock


\bibitem[Xu et~al\mbox{.}(2018)]%
        {xu2018attngan}
\bibfield{author}{\bibinfo{person}{Tao Xu}, \bibinfo{person}{Pengchuan Zhang},
  \bibinfo{person}{Qiuyuan Huang}, \bibinfo{person}{Han Zhang},
  \bibinfo{person}{Zhe Gan}, \bibinfo{person}{Xiaolei Huang}, {and}
  \bibinfo{person}{Xiaodong He}.} \bibinfo{year}{2018}\natexlab{}.
\newblock \showarticletitle{{AttnGAN}: Fine-grained text to image generation
  with attentional generative adversarial networks}. In
  \bibinfo{booktitle}{\emph{Proceedings of the IEEE conference on computer
  vision and pattern recognition}}. \bibinfo{pages}{1316--1324}.
\newblock


\bibitem[Yang et~al\mbox{.}(2023)]%
        {Yang2022PaintBE}
\bibfield{author}{\bibinfo{person}{Binxin Yang}, \bibinfo{person}{Shuyang Gu},
  \bibinfo{person}{Bo Zhang}, \bibinfo{person}{Ting Zhang},
  \bibinfo{person}{Xuejin Chen}, \bibinfo{person}{Xiaoyan Sun},
  \bibinfo{person}{Dong Chen}, {and} \bibinfo{person}{Fang Wen}.}
  \bibinfo{year}{2023}\natexlab{}.
\newblock \showarticletitle{Paint by example: Exemplar-based image editing with
  diffusion models}. In \bibinfo{booktitle}{\emph{Proceedings of the IEEE/CVF
  Conference on Computer Vision and Pattern Recognition}}.
  \bibinfo{pages}{18381--18391}.
\newblock


\bibitem[Yu et~al\mbox{.}(2022)]%
        {yu2022scaling}
\bibfield{author}{\bibinfo{person}{Jiahui Yu}, \bibinfo{person}{Yuanzhong Xu},
  \bibinfo{person}{Jing~Yu Koh}, \bibinfo{person}{Thang Luong},
  \bibinfo{person}{Gunjan Baid}, \bibinfo{person}{Zirui Wang},
  \bibinfo{person}{Vijay Vasudevan}, \bibinfo{person}{Alexander Ku},
  \bibinfo{person}{Yinfei Yang}, \bibinfo{person}{Burcu~Karagol Ayan},
  {et~al\mbox{.}}} \bibinfo{year}{2022}\natexlab{}.
\newblock \showarticletitle{Scaling Autoregressive Models for Content-Rich
  Text-to-Image Generation}.
\newblock \bibinfo{journal}{\emph{arXiv preprint arXiv:2206.10789}}
  (\bibinfo{year}{2022}).
\newblock


\bibitem[Zhang et~al\mbox{.}(2017)]%
        {zhang2017stackgan}
\bibfield{author}{\bibinfo{person}{Han Zhang}, \bibinfo{person}{Tao Xu},
  \bibinfo{person}{Hongsheng Li}, \bibinfo{person}{Shaoting Zhang},
  \bibinfo{person}{Xiaogang Wang}, \bibinfo{person}{Xiaolei Huang}, {and}
  \bibinfo{person}{Dimitris~N Metaxas}.} \bibinfo{year}{2017}\natexlab{}.
\newblock \showarticletitle{{StackGAN}: Text to photo-realistic image synthesis
  with stacked generative adversarial networks}. In
  \bibinfo{booktitle}{\emph{Proc.~ICCV}}. \bibinfo{pages}{5907--5915}.
\newblock


\bibitem[Zhang et~al\mbox{.}(2018)]%
        {zhang2018stackgan++}
\bibfield{author}{\bibinfo{person}{Han Zhang}, \bibinfo{person}{Tao Xu},
  \bibinfo{person}{Hongsheng Li}, \bibinfo{person}{Shaoting Zhang},
  \bibinfo{person}{Xiaogang Wang}, \bibinfo{person}{Xiaolei Huang}, {and}
  \bibinfo{person}{Dimitris~N Metaxas}.} \bibinfo{year}{2018}\natexlab{}.
\newblock \showarticletitle{{StackGAN}++: Realistic image synthesis with
  stacked generative adversarial networks}.
\newblock \bibinfo{journal}{\emph{IEEE transactions on pattern analysis and
  machine intelligence}} \bibinfo{volume}{41}, \bibinfo{number}{8}
  (\bibinfo{year}{2018}), \bibinfo{pages}{1947--1962}.
\newblock


\bibitem[Zhu et~al\mbox{.}(2020b)]%
        {zhu2020domain}
\bibfield{author}{\bibinfo{person}{Jiapeng Zhu}, \bibinfo{person}{Yujun Shen},
  \bibinfo{person}{Deli Zhao}, {and} \bibinfo{person}{Bolei Zhou}.}
  \bibinfo{year}{2020}\natexlab{b}.
\newblock \showarticletitle{In-domain gan inversion for real image editing}. In
  \bibinfo{booktitle}{\emph{European conference on computer vision}}. Springer,
  \bibinfo{pages}{592--608}.
\newblock


\bibitem[Zhu et~al\mbox{.}(2020a)]%
        {Zhu2020ImprovedSE}
\bibfield{author}{\bibinfo{person}{Peihao Zhu}, \bibinfo{person}{Rameen Abdal},
  \bibinfo{person}{Yipeng Qin}, {and} \bibinfo{person}{Peter Wonka}.}
  \bibinfo{year}{2020}\natexlab{a}.
\newblock \showarticletitle{Improved StyleGAN Embedding: Where are the Good
  Latents?}
\newblock \bibinfo{journal}{\emph{ArXiv}}  \bibinfo{volume}{abs/2012.09036}
  (\bibinfo{year}{2020}).
\newblock


\end{thebibliography}
